\documentclass[10pt,twocolumn,letterpaper]{article}
\usepackage[pagenumbers]{cvpr}
\usepackage[T1]{fontenc}
\usepackage[utf8]{inputenc}
\usepackage{amsmath,amssymb}
\usepackage{graphicx}
\usepackage{booktabs,array,multirow}
\usepackage{pifont}
\newcommand{\realyes}{\ding{51}}
\newcommand{\realno}{\ding{55}}
\newcommand{\releasedmark}{\textsuperscript{\ensuremath{\ast}}}
\usepackage{longtable}
\usepackage{pdflscape}
\usepackage{placeins}
\usepackage{microtype}
\definecolor{cvprblue}{rgb}{0.21,0.49,0.74}

\newcommand{\HRvox}{\ensuremath{\mathrm{HR\,vox}}}
\newcommand{\siref}[1]{\ref{#1}}
\newcommand{\sg}{\operatorname{sg}}
\newcommand{\clip}{\operatorname{clip}}
\newsavebox{\tablebox}
\newcommand{\SITableCaption}{}

\newcommand{\maintablenotes}{\par\vspace{3pt}{\footnotesize Real: \realyes{} real-pair; \realno{} synthetic-only training. \releasedmark{} Directly reused VoDaSuRe checkpoints. SDice: Surface Dice; HRcov: HR-covered; Unsup.: Unsupported. PSNR: dB; ASSD, HD95 and Tb.Th: \HRvox{}. SDice, HRcov and Unsup. (\%) use a 1~\HRvox{} tolerance. BV/TV and Tb.Th: absolute errors. \textbf{Bold}: best.}}
\newcommand{\secondarytablenotes}[1][]{\par\vspace{3pt}{\footnotesize #1 Metric conventions follow Table~\ref{tab:1}. \textbf{Bold}: best.}}

\usepackage[resetlabels]{multibib}
\usepackage[pagebackref,breaklinks,colorlinks,allcolors=cvprblue]{hyperref}
\newcites{supp}{References}
\hypersetup{
  pdftitle={Real Pairs Are Not Enough: Physically Anchored Supervision for Volumetric CT Super-Resolution},
  pdfauthor={Haoran Liu, Zipei Yan, Mingzhe Liu, Jizhou Li, Guibin Zan},
  bookmarksnumbered=true,
  bookmarksdepth=3
}
\title{Real Pairs Are Not Enough: Physically Anchored Supervision for Volumetric CT Super-Resolution}
\author{%
  Haoran Liu\textsuperscript{1}\quad
  Zipei Yan\textsuperscript{2}\quad
  Mingzhe Liu\textsuperscript{1}\quad
  Jizhou Li\textsuperscript{2}\quad
  Guibin Zan\textsuperscript{3}\\[0.35em]
  {\normalsize\textsuperscript{1}Chengdu University of Technology}\quad
  {\normalsize\textsuperscript{2}The Chinese University of Hong Kong}\\
  {\normalsize\textsuperscript{3}Specreation Instruments Co., Ltd.}%
}
\begin{document}
\raggedbottom
\makeatletter
\let\PINsavedtextbottom\@textbottom
\def\@textbottom{%
  \ifnum\value{page}=1\relax
    \if@firstcolumn\PINsavedtextbottom\fi
  \else\PINsavedtextbottom\fi}
\makeatother
\maketitle
\addtocontents{toc}{\protect\setcounter{tocdepth}{-1}}
\begin{abstract}
Volumetric CT super-resolution (SR) models are commonly trained on synthetic or physically acquired pairs of low-resolution (LR) and high-resolution (HR) volumes. However, synthetic pairs miss real acquisition effects, whereas registered physical pairs retain inter-scan noise, artifact, and microstructure differences---causing direct voxelwise regression, and even naive synthetic--real mixing, to oversmooth fine porous structures. We introduce the Physically-anchored Implicit Network (PIN) with Real-Pair Dual-Anchor learning, which decouples physical-pair supervision into two complementary roles. The HR anchor teaches fine structure from registered HR through pair-conditioned spectral replay, while the LR anchor constrains the prediction from acquired LR via affine-calibrated low-pass projection. Together with exact synthetic supervision in one shared volumetric implicit model, PIN learns from both physical scans without forcing the acquired-LR reconstruction to match registered HR voxel by voxel. Inference requires only acquired LR. Across 30 external baseline configurations on real CT, PIN achieves the lowest full-volume LPIPS, reduces average symmetric surface distance by 42.3--49.5\% and bone-volume-fraction error by 52.2--53.4\% relative to the three strongest baselines for each metric, and lowers the erased reference-pore volume fraction from 37.3--72.9\% to 10.4\%. Code, pretrained models, and evaluation scripts will be publicly available.
\end{abstract}

\section{Introduction}
\label{sec:introduction}

In volumetric computed tomography (CT), a thin wall lost by reconstruction, a narrow separation closed into a false connection, or a displaced porous boundary can change specimen geometry and morphology. Such errors can bias thickness and volume-fraction measurements even when the volume looks plausible. Super-resolution (SR) therefore needs supervision that reflects physical acquisition and supports the recovery of fine structures on which these measurements depend.

\begin{figure}[t]
\centering
\includegraphics[width=\linewidth]{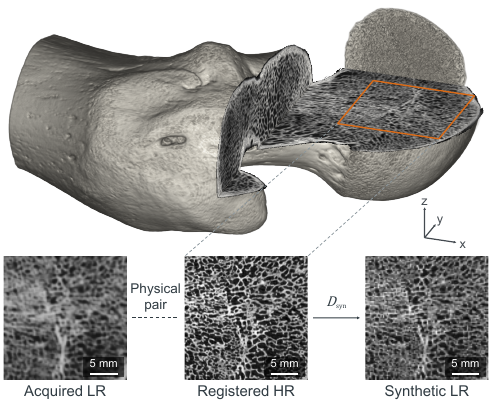}
\caption{\textbf{Why neither synthetic nor physical pairs alone suffice for voxelwise CT SR supervision.} The orange box in the femur cutaway marks the trabecular region shown below under a shared intensity window. Constructing Synthetic LR from Registered HR via $D_{\mathrm{syn}}$ preserves exact voxelwise alignment, but the resulting view differs visibly from Acquired LR, exposing the synthetic-to-real acquisition gap. Conversely, the registered physical pair retains pronounced differences in noise and resolved detail between Acquired LR and Registered HR, causing direct voxelwise regression to oversmooth fine porous structures. This raises a central challenge: how to learn from both physical scans without forcing the Acquired LR reconstruction to match Registered HR voxelwise.}
\label{fig:1}
\end{figure}

Volumetric CT SR models are commonly trained using synthetic or physically acquired pairs of low-resolution (LR) and high-resolution (HR) volumes. Downsampling HR creates synthetic LR inputs with exact input--target correspondence, but the resulting degradation may not reproduce physical acquisition. RPLHR-CT documented this gap between synthetic and physically acquired LR~\cite{ref2}. Registered real pairs are expected to bridge this gap by providing acquisition-specific information from both scans, yet registration does not make two independent acquisitions an exact voxelwise input--target pair. Both strategies therefore have limitations as supervision for real-world CT SR (Fig.~\ref{fig:1}).

Recent work on VoDaSuRe showed that direct voxelwise training on registered physical pairs produces spatially averaged fine structure, and found that imperfect registration alone could not explain this behavior~\cite{ref1}. Differences in resolution, artifacts, and noise between the two acquisitions may contribute. When acquired LR leaves acquisition-specific variation in registered HR unresolved, voxelwise $L_1$ regression favors conditional-median predictions~\cite{Lehtinen2018Noise2Noise}, which can smooth fine structure and close narrow pores---a limitation that persists even when synthetic and physical voxelwise targets are directly mixed. The resulting design problem is how to learn from both acquired LR and HR volumes while addressing the synthetic-to-real gap and the oversmoothing associated with direct voxelwise supervision of real pairs.

\begin{figure*}[t]
\centering
\includegraphics[width=\textwidth]{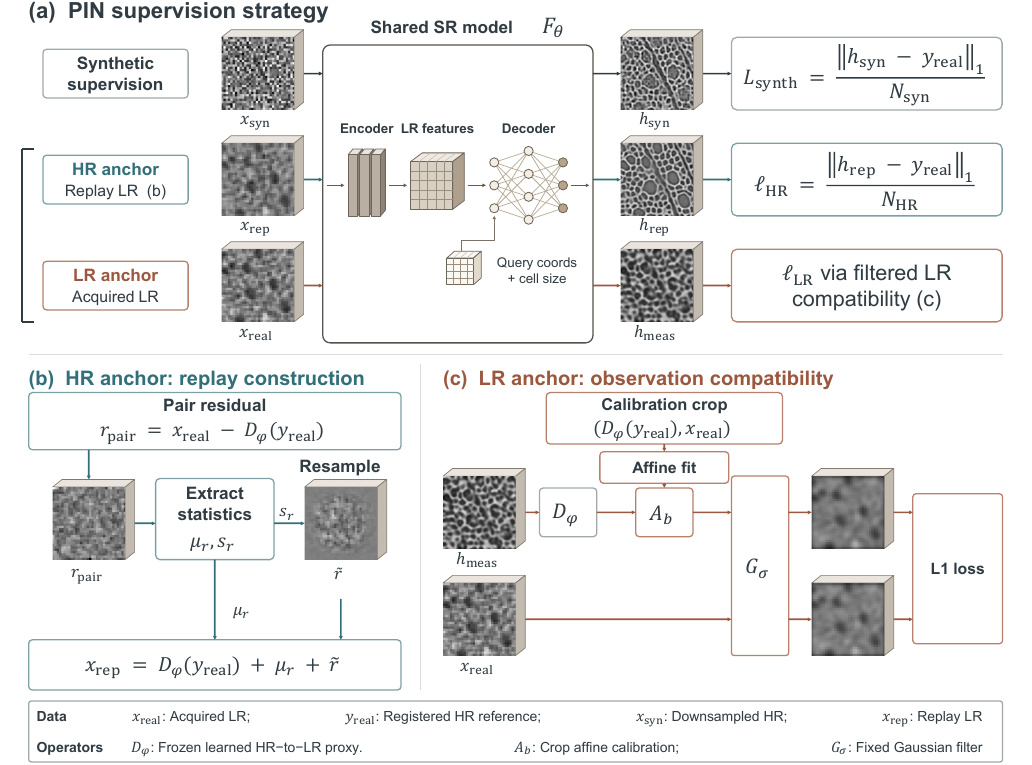}
\caption{\textbf{PIN supervision strategy.} (a) Synthetic supervision and Real-Pair Dual-Anchor learning jointly train one shared SR model comprising a 3D encoder and a coordinate decoder. The synthetic and HR-anchor paths use registered-HR point supervision, while the LR anchor enforces acquired-LR observation compatibility. (b) The paired residual supplies its mean and band strengths for replay construction. The mean and resampled residual are added to the frozen learned HR-to-LR projection to form replay LR. (c) The projected SR output is affine-calibrated and compared with acquired LR after identical Gaussian filtering, using the normalized $L_1$ loss. Here $N_{\mathrm{syn}}$, $N_{\mathrm{HR}}$, and $N_{\mathrm{LR}}$ count the scalar entries compared in the three paths. Matched Oak views share a CT intensity window; the two residual views share a separate window. All displayed SR outputs use the same frozen PIN checkpoint. Inference requires only acquired LR. Crop/query indices and replay clipping are omitted for clarity (Sec.~\ref{sec:method}).}
\label{fig:2}
\end{figure*}

We introduce the \textbf{Physically-anchored Implicit Network (PIN)} with \textbf{Real-Pair Dual-Anchor learning}. The \textbf{LR anchor} constrains the reconstruction by mapping it back to LR through a frozen proxy and comparing it with the acquired LR scan after affine calibration and low-pass filtering. Because low-pass observation consistency alone leaves high-frequency HR detail underconstrained, the \textbf{HR anchor} teaches fine structure from registered HR through pair-conditioned replay, which constructs HR-aligned LR training inputs matching paired residual statistics with randomized Fourier phase. Together, these anchors use both acquisitions for supervision without directly using registered HR as a voxelwise target for the acquired-LR reconstruction. A third path retains exact synthetic supervision using downsampled HR inputs and their corresponding HR targets. All three paths jointly train a single volumetric implicit model that requires only acquired LR at inference (Fig.~\ref{fig:2}).

We validate Real-Pair Dual-Anchor learning through controlled comparisons against synthetic-only, real-only, and directly mixed supervision on the same backbone, component ablations of each path, and benchmarking against 30 external configurations. Beyond PSNR, SSIM, and LPIPS, we evaluate surface geometry, material selectivity, 3D reference-pore retention, and bone morphology. PIN achieves the lowest full-volume LPIPS, reduces surface distance and bone-morphology errors, and substantially lowers reference-pore erasure. The contrasting rankings between PSNR/SSIM and structural metrics further show that pixelwise similarity alone is insufficient for evaluating scientific CT SR.

\section{Related work}
\label{sec:related}

\subsection{Volumetric super-resolution}
Earlier volumetric SR studies focused on network architecture and representation design to improve reconstruction quality. CT research has explored learned reconstruction~\cite{Frazer2024ClinicalCT,Klos2025LatticeCT,Zhou2024CTModels,Jackson2022MultiresolutionCT,Zhang2025AESR3D,Senck2025TarsalBoneSR} and introduced paired imaging resources~\cite{Wang2026LumbarSR,Vidal2024MultiresolutionCT,Green2025SynchrotronData}, while ArSSR, CuNeRF, and CycleINR investigate implicit representations and flexible volumetric reconstruction~\cite{ref4,ref5,ref3}. VoDaSuRe~\cite{ref1}, however, identified a different limitation: direct pointwise training on registered physical LR--HR pairs can oversmooth fine structure, and registration errors alone cannot explain this behavior. PIN takes this finding as its starting point and redesigns how registered physical pairs contribute to supervision.

Studies of bone CT SR have reported that conventional image-fidelity metrics do not always reflect the quality of recovered microstructure and morphology~\cite{Jhuboo2022BoneMetrics,Chan2023BoneDiffusion}. This motivates including structural measurements alongside PSNR and SSIM in CT SR evaluation.

\subsection{Real-world degradation and adaptation}
RPLHR-CT~\cite{ref2} documented a gap between synthetically downsampled and physically acquired LR CT scans. Real-world SR approaches seek to bridge such gaps through degradation estimation~\cite{BellKligler2019KernelGAN,Gu2019IKC}, degradation-aware modeling~\cite{Wang2021DegradationRepresentation,Yue2022BSRDM}, and richer synthesis, as in BSRGAN and Real-ESRGAN~\cite{ref6,Zhang2021PracticalDegradation}. DASR uses domain-distance-aware adaptation~\cite{ref7}, while coupled-dictionary learning~\cite{ref8} and SR-CycleGAN~\cite{ref9} address transfer between LR and HR domains. These approaches primarily aim to improve the realism or transferability of LR inputs.

\subsection{Alternative supervision and LR constraints}
Observation-domain constraints connect a prediction to its LR input in supervised and image-specific training~\cite{Zhang2020USRNet,Emad2021DualSR}. GAN-CIRCLE and CycleINR include LR reconstruction cycles~\cite{You2020GANCIRCLE,ref5}. Dual Regression Networks (DRN) combine paired HR teaching with learned LR regression and extend this supervision to unpaired acquired LR data~\cite{ref10}. LWay adapts a pretrained SR model to acquired LR through a frozen LR reconstructor~\cite{ref11}. Mimicked Alignment uses acquired LR to guide a learned transformation of downsampled HR into an aligned, degradation-matched HR-supervised input~\cite{Elezabi_2024_ACCV}.

Rather than adapting to unpaired LR scans~\cite{ref11,ref10} or training the SR network only on mimicked LR inputs~\cite{Elezabi_2024_ACCV}, PIN addresses how a registered physical CT pair supervises training. While Mimicked Alignment~\cite{Elezabi_2024_ACCV} does not explicitly enforce observation consistency on SR predictions from acquired LR and LWay~\cite{ref11} lacks paired-HR teaching during adaptation, PIN redesigns both mechanisms around a shared physical proxy: registered HR teaches fine structure via spectral replay with randomized Fourier phase, while acquired LR constrains the inference path through affine-calibrated low-pass projection.

\section{Method}
\label{sec:method}

\subsection{Problem formulation and PIN overview}
Let $(x_{\mathrm{real}},y_{\mathrm{real}})$ be a registered physical LR--HR pair and $F_\theta$ a fourfold volumetric SR model. To learn from both scans without forcing voxelwise alignment across distinct acquisitions, PIN jointly optimizes three input--supervision relations within a single model (Fig.~\ref{fig:2}): synthetic LR with its constructed HR point target, pair-conditioned replay with registered-HR point supervision (HR anchor), and actual acquired LR with observation-domain compatibility (LR anchor). At inference, reconstruction requires only $F_\theta(x_{\mathrm{real}})$.

A shared 3D residual-dense encoder extracts features from each LR input, and a LIIF-style decoder predicts HR intensities from these features, query coordinates, and cell sizes, using local ensemble and feature unfolding~\cite{ref12}. Both anchors use the same frozen learned 3D convolutional HR-to-LR proxy $D_\phi$, pretrained on the VoDaSuRe training pairs.

For LR input $z$ and HR query list or grid $\mathcal P$, $F_\theta(z;\mathcal P)$ predicts intensities with cell widths set by the HR grid. $F_\theta(z)$ denotes dense-grid reconstruction.

\subsection{Exact synthetic supervision}
Exact synthetic supervision uses voxelwise $L_1$ reconstruction with exact correspondence. Let $\mathcal Q_{\mathrm{syn}}$ be the sampled HR coordinate list and $y_{\mathcal Q_{\mathrm{syn}}}$ its voxel-lookup target array (Suppl. Sec.~\siref{sec:implementation}). Write $N_{\mathrm{syn}}=|\mathcal Q_{\mathrm{syn}}|$. Fourfold trilinear downsampling $D_{\mathrm{syn}}$ constructs LR from acquired HR:
\begin{equation}
\begin{aligned}
x_{\mathrm{syn}} &= D_{\mathrm{syn}}(y_{\mathrm{real}}),\\
h_{\mathrm{syn}} &= F_\theta(x_{\mathrm{syn}};\mathcal Q_{\mathrm{syn}}),\\
L_{\mathrm{synth}} &= \frac{1}{N_{\mathrm{syn}}}\|h_{\mathrm{syn}}-y_{\mathcal Q_{\mathrm{syn}}}\|_1.
\end{aligned}
\label{eq:synthetic}
\end{equation}

\subsection{HR anchor: Pair-conditioned replay}
The HR anchor applies registered-HR point supervision to a replay input constructed from the registered physical pair. Let $(x_b,y_b)$ be its paired anchor subcrop, indexed by $b$, with LR and HR grids $\mathcal G_{\mathrm{LR},b}$ and $\mathcal G_{\mathrm{HR},b}$. Their voxel counts are $N_{\mathrm{LR}}=|\mathcal G_{\mathrm{LR},b}|$ and $N_{\mathrm{HR}}=|\mathcal G_{\mathrm{HR},b}|$. The frozen proxy forms the base $L_0=D_\phi(y_b)$. From $r_{\mathrm{pair}}=x_b-L_0$, PIN retains the scalar mean $\mu_r=N_{\mathrm{LR}}^{-1}\sum_{i=1}^{N_{\mathrm{LR}}}(r_{\mathrm{pair}})_i$ and four frequency-band strengths $s_r=\operatorname{Stats}_4(r_{\mathrm{pair}}-\mu_r)\in\mathbb R_{\geq0}^{4}$. The seeded residual generator $\widetilde r=\mathcal R(s_r;\xi_b)$ produces a newly sampled zero-mean residual with randomized Fourier phase, where $\xi_b$ identifies its random draw:
\begin{equation}
\begin{aligned}
x_{\mathrm{rep}} &= \clip_{[0,1]}
 \bigl(L_0+\mu_r+\widetilde r\bigr),\\
h_{\mathrm{rep}} &= F_\theta(x_{\mathrm{rep}};\mathcal G_{\mathrm{HR},b}),\\
\ell_{\mathrm{HR}} &= \frac{1}{N_{\mathrm{HR}}}\|h_{\mathrm{rep}}-y_b\|_1,\\
\bar L_{\mathrm{HR}} &= c_{\mathrm{HR}}\ell_{\mathrm{HR}}.
\end{aligned}
\label{eq:replay}
\end{equation}
Using the complete measured residual would recover the acquired LR input $x_b$ before clipping and restore direct acquired-LR-to-registered-HR regression. PIN instead resamples the residual using its pair-conditioned mean and band strengths, changing the supervised input--target relation without reproducing the measured LR realization (Fig.~\ref{fig:2}b).

\subsection{LR anchor: Observation compatibility}
The LR anchor constrains the prediction made from the actual inference input, $h_{\mathrm{meas}}=F_\theta(x_b;\mathcal G_{\mathrm{HR},b})$. It projects that prediction through the same frozen proxy and compares it with the acquired observation on the LR grid. A robust affine map $A_b(z)=a_bz+b_b$ is estimated by Huber iteratively reweighted least squares between $D_\phi(y_b)$ and $x_b$ on each LR calibration crop, with validity flag $\chi_b\in\{0,1\}$. The fitted coefficients correct intensity scale and offset and are treated as constants during backpropagation, preserving gradient flow through $D_\phi(h_{\mathrm{meas}})$ to $F_\theta$. Let $\widehat L=A_b(D_\phi(h_{\mathrm{meas}}))$ be the calibrated projection. With identical Gaussian filters $G_\sigma$ ($\sigma=1.0$ LR voxel), the loss averages absolute differences over $\mathcal G_{\mathrm{LR},b}$:
\begin{equation}
\begin{aligned}
\bar L_{\mathrm{LR}}&=c_{\mathrm{LR}}\ell_{\mathrm{LR}},\\
\ell_{\mathrm{LR}}&=
\begin{cases}
\frac{1}{N_{\mathrm{LR}}}\|G_\sigma\ast\widehat L-G_\sigma\ast x_b\|_1,&\chi_b=1,\\
0,&\chi_b=0.
\end{cases}
\end{aligned}
\label{eq:lr-anchor}
\end{equation}
The Gaussian filter then focuses the comparison on lower-frequency LR structure, reducing sensitivity to acquisition-specific high-frequency noise and artifacts (Fig.~\ref{fig:2}c). If the affine fit is invalid, the LR anchor is disabled for that crop while the HR anchor remains active.

Because the LR anchor uses a fixed proxy and filtered LR comparison, it constrains observation consistency within the comparison band without selecting a unique HR reconstruction.

\subsection{Objective and cotangent correction}
The core objective combines the three losses:
\begin{equation}
L_{\mathrm{core}}=L_{\mathrm{synth}}
  +\alpha\bar L_{\mathrm{HR}}+\beta\bar L_{\mathrm{LR}}.
\label{eq:core}
\end{equation}
Here $\ell_{\mathrm{HR}}$ and $\ell_{\mathrm{LR}}$ are the raw normalized $L_1$ anchor losses; the fixed factors $c_{\mathrm{HR}}$ and $c_{\mathrm{LR}}$ normalize anchor loss scales. All three core terms update the same parameters $\theta$.

During training, an optional bounded cotangent safeguard provides auxiliary output-space guidance and monitors for high-frequency copying anomalies (Suppl. Sec.~\siref{sec:correction}).

\textbf{Implementation details.} PIN is trained on 13 VoDaSuRe volumes for 60 epochs using Adam with an initial learning rate of $10^{-4}$ and cosine decay to $10^{-6}$ on two NVIDIA RTX 5090 GPUs. The anchor weights are $\alpha=0.7$ and $\beta=0.3$. We evaluate the final checkpoint. Additional implementation details are provided in Suppl. Sec.~\siref{sec:implementation}.

\section{Experimental protocol}
\label{sec:protocol}

\subsection{Data and comparison protocol}
\label{sec:datasets-comparators}
We evaluate fourfold CT SR on registered physical LR--HR pairs from VoDaSuRe~\cite{ref1}, using its released training and test partitions. The benchmark uses VoDaSuRe's released registered and intensity-matched LR volumes (REG) as model inputs. The released test partition contains whole held-out femur and vertebrae scans and axial portions of other specimens that are disjoint from their training portions. Ten test evaluation units contribute four HR-grid ROIs each. All predictions are evaluated on the same registered-HR grids.

The external benchmark includes 30 configurations from 15 method families under synthetic-only and real-pair training. We evaluate 16 author-released VoDaSuRe checkpoints from RCAN~\cite{ref21}, HAT~\cite{ref22}, EDDSR~\cite{ref23}, MFER~\cite{ref24}, mDCSRN~\cite{ref25}, RRDBNet3D, SuperFormer~\cite{ref27}, and MTVNet~\cite{ref28}. The remaining 14 configurations from MetaSR~\cite{ref15}, LIIF~\cite{ref12}, LTE~\cite{ref16}, SRNO~\cite{ref17}, CiaoSR~\cite{ref18}, HIIF~\cite{ref19}, and LINF~\cite{ref20} are independently trained with PIN's 13-volume split and direct $4\times$ training, holding out Femur\_74 for validation. The released checkpoints use 14 training volumes and a staged $2\times$-to-$4\times$ curriculum~\cite{ref1}. Retraining the three strongest released baselines on PIN's split and scale protocol preserves PIN's geometric advantage (Suppl. Sec.~\siref{sec:training-protocol-sensitivity}). Table~\ref{tab:1} displays the eight lowest-ASSD configurations per training domain; complete results and model counts appear in Suppl. Sec.~\siref{sec:external}.

All supervision strategies and component ablations use the same 15.50M-parameter SR network $F_\theta(x_{\mathrm{real}})$ at inference. The learned proxy, both anchor losses, and the optional safeguard are used only during training and introduce no additional inference computation.

\begin{table*}[t]
\caption{\textbf{Representative external benchmark comparison.}}
\label{tab:1}
\centering
\fontsize{9}{10.8}\selectfont
\setlength{\tabcolsep}{2.5pt}
\renewcommand{\arraystretch}{1.13}
\sbox{\tablebox}{\begin{tabular}{lcrrrrrrrrrrrr}
\toprule
 & & \multicolumn{3}{c}{Image} & \multicolumn{4}{c}{Surface geometry} & \multicolumn{3}{c}{Material fidelity} & \multicolumn{2}{c}{Morphology} \\
\cmidrule(lr){3-5}\cmidrule(lr){6-9}\cmidrule(lr){10-12}\cmidrule(lr){13-14}
Method & Real & PSNR$\uparrow$ & SSIM$\uparrow$ & LPIPS$\downarrow$ & ASSD$\downarrow$ & HD95$\downarrow$ & SDice$\uparrow$ & HRcov$\uparrow$ & Precision$\uparrow$ & Unsup.$\downarrow$ & \shortstack{VF error$\downarrow$} & BV/TV$\downarrow$ & Tb.Th$\downarrow$ \\
\midrule
mDCSRN\releasedmark{} & \realno{} & 15.201 & 0.3840 & 0.4666 & 1.724 & 5.855 & 0.5669 & 0.4568 & 0.8330 & 2.41 & \textbf{0.0341} & 0.0394 & 5.326 \\
MTVNet\releasedmark{} & \realno{} & 15.042 & 0.3478 & 0.4890 & 1.839 & 6.183 & 0.5537 & 0.4453 & 0.8329 & 2.42 & 0.0358 & 0.0402 & 5.804 \\
RRDBNet3D\releasedmark{} & \realno{} & 15.230 & 0.3806 & 0.4772 & 1.968 & 6.538 & 0.5393 & 0.4308 & 0.8321 & 2.40 & 0.0378 & 0.0388 & 5.713 \\
MFER\releasedmark{} & \realno{} & 14.933 & 0.3504 & 0.4770 & 2.105 & 6.742 & 0.5109 & 0.4046 & 0.8258 & 2.84 & 0.0369 & 0.0411 & 5.460 \\
EDDSR\releasedmark{} & \realno{} & 15.320 & 0.3702 & 0.4781 & 2.175 & 6.889 & 0.5175 & 0.4106 & 0.8289 & 2.58 & 0.0377 & 0.0409 & 5.445 \\
SuperFormer\releasedmark{} & \realno{} & 15.348 & 0.3741 & 0.4631 & 2.195 & 7.119 & 0.5240 & 0.4157 & 0.8293 & 2.53 & 0.0386 & 0.0414 & 5.971 \\
RCAN\releasedmark{} & \realno{} & 14.915 & 0.3336 & 0.5082 & 2.759 & 10.984 & 0.4933 & 0.3979 & 0.8163 & 3.75 & 0.0386 & 0.0429 & 4.968 \\
HAT\releasedmark{} & \realno{} & 14.942 & 0.3075 & 0.5331 & 2.853 & 8.936 & 0.4750 & 0.3744 & 0.8215 & 2.89 & 0.0437 & 0.0418 & 5.172 \\
RRDBNet3D\releasedmark{} & \realyes{} & 16.353 & 0.4191 & 0.4788 & 3.998 & 12.094 & 0.4648 & 0.3810 & 0.8267 & 2.35 & 0.0744 & 0.0436 & 1.590 \\
HAT\releasedmark{} & \realyes{} & 15.686 & 0.3466 & 0.5342 & 4.194 & 11.688 & 0.4053 & 0.3280 & 0.8232 & 2.68 & 0.0712 & 0.0431 & 2.054 \\
SuperFormer\releasedmark{} & \realyes{} & \textbf{16.412} & \textbf{0.4229} & 0.4751 & 4.467 & 12.076 & 0.4412 & 0.3632 & 0.8281 & 2.31 & 0.0763 & 0.0397 & 2.456 \\
mDCSRN\releasedmark{} & \realyes{} & 16.273 & 0.4071 & 0.4789 & 4.694 & 13.651 & 0.4164 & 0.3410 & 0.8248 & 2.43 & 0.0824 & 0.0433 & 1.778 \\
MFER\releasedmark{} & \realyes{} & 16.165 & 0.4021 & 0.4793 & 4.915 & 14.357 & 0.4277 & 0.3500 & 0.8245 & 2.49 & 0.0793 & 0.0451 & 1.724 \\
HIIF & \realyes{} & 15.933 & 0.3850 & 0.4727 & 4.990 & 13.765 & 0.4249 & 0.3502 & 0.8218 & 2.65 & 0.0748 & 0.0404 & 1.465 \\
LTE & \realyes{} & 16.028 & 0.3977 & 0.4856 & 5.107 & 14.102 & 0.4006 & 0.3293 & 0.8194 & 2.73 & 0.0803 & 0.0399 & 2.243 \\
MetaSR & \realyes{} & 16.038 & 0.3976 & 0.4996 & 5.114 & 13.721 & 0.4213 & 0.3486 & 0.8208 & 2.68 & 0.0781 & 0.0403 & 2.402 \\
\textbf{PIN} & \textendash{} & 15.090 & 0.3878 & \textbf{0.4514} & \textbf{0.995} & \textbf{3.473} & \textbf{0.7046} & \textbf{0.6137} & \textbf{0.8574} & \textbf{1.60} & 0.0411 & \textbf{0.0185} & \textbf{1.425} \\
\bottomrule
\end{tabular}
}
\typeout{PIN_TABLE_1_WIDTH=\the\wd\tablebox; LIMIT=\the\textwidth}
\usebox{\tablebox}
\maintablenotes
\end{table*}

\subsection{Metrics and aggregation}
\label{sec:metrics-aggregation}
\textbf{Image quality.} We use slice-based PSNR, SSIM, and LPIPS~\cite{ref13}. \textbf{Surface geometry.} Area-weighted ASSD and HD95 summarize bidirectional surface distance and the larger directional 95th-percentile discrepancy, respectively. Surface Dice~\cite{ref14} and HR-covered measure symmetric agreement and reference-surface coverage within an HR-grid tolerance. Geometric distances, thickness errors, and distance tolerances are reported in native HR-voxel units (HR vox), using unit spacing on each evaluated HR grid; physical voxel pitch varies across VoDaSuRe specimens (Suppl. Sec.~\siref{sec:metrics}). \textbf{Material fidelity.} We use voxel Precision, Unsupported solid volume at 1~\HRvox{}, and absolute solid-volume-fraction (VF) error \(\bigl||S|/N-|H|/N\bigr|\), for predicted/reference solid masks \(S,H\) over \(N\) ROI voxels. \textbf{Morphology.} We report absolute errors in bone-volume fraction (BV/TV) and trabecular thickness (Tb.Th) on three and two predefined bone-profile units, respectively~\cite{Bouxsein2010MicroCTGuidelines,Rytky2024ClinicalBone}. We also evaluate local solid-diameter error. Additional material and pore diagnostics appear in Suppl. Sec.~\siref{sec:material-topology}.

Image scores average valid slices within each full volume, then volumes equally. Surface and material metrics average HR-grid ROI values within evaluation units, then units equally. Surface distances retain finite ROI support: Table~\ref{tab:1} ranks methods by native finite ASSD; paired comparisons use common finite ROIs.

\begin{figure*}[t]
\centering
\includegraphics[width=\textwidth]{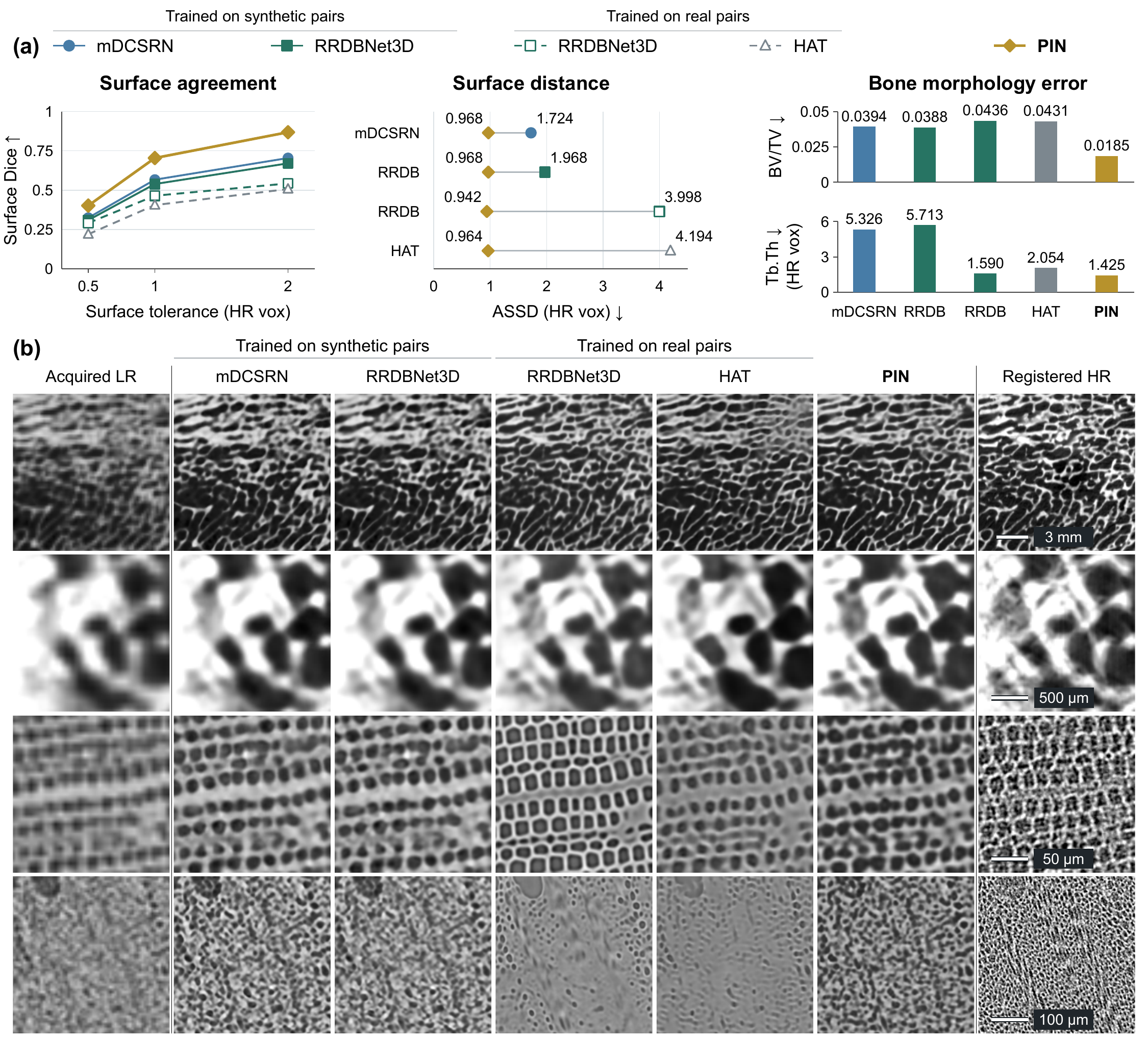}
\caption{\textbf{Geometric and morphological recovery in the external benchmark.} PIN achieves stronger surface agreement and lower morphology errors than the displayed baselines, while better preserving thin separations and porous interfaces visible in registered HR. (a) Surface Dice at 0.5, 1, and 2~\HRvox{} tolerances (left), paired ASSD (center), and BV/TV and Tb.Th absolute errors over three and two predefined bone-profile evaluation units, respectively (right). ASSD uses the finite ROIs shared by each baseline and PIN; PIN values vary with the paired support. (b) Matched Femur, Vertebra, Cypress, and Elm slices share orientation and window within rows. The LR input and registered HR reference come from separate scans of the same specimen. RRDB abbreviates RRDBNet3D.}
\label{fig:3}
\end{figure*}

\subsection{Ablation studies}
We examine PIN's supervision design through a supervision-strategy ablation and a component ablation. All runs use the same SR backbone, dataset split, preprocessing, optimizer, training schedule, accumulation, and evaluator.

\textbf{Supervision-strategy ablation.} We compare three alternative supervision strategies against PIN. Synth-only uses exact synthetic supervision alone. Real-only uses direct acquired-LR-to-registered-HR point supervision. Direct Mix combines the synthetic and acquired-pair point-target relations. Its acquired-pair query list $\mathcal Q_{\mathrm{real}}$ and voxel-lookup target array $y_{\mathcal Q_{\mathrm{real}}}$ are sampled separately from the synthetic queries, with the same query count $|\mathcal Q_{\mathrm{real}}|=N_{\mathrm{syn}}$:
\begin{equation}
\begin{aligned}
L_{\mathrm{DirectMix}}&=L_{\mathrm{synth}}\\
&\quad+\frac{1}{N_{\mathrm{syn}}}\|F_\theta(x_{\mathrm{real}};\mathcal Q_{\mathrm{real}})
             -y_{\mathcal Q_{\mathrm{real}}}\|_1.
\end{aligned}
\label{eq:direct-mix}
\end{equation}

\textbf{Component ablation.} We use A0 as the correction-off reference, retaining all three PIN supervision paths while disabling the optional cotangent correction. A1, A2, and A3 remove synthetic supervision, the HR anchor, and the LR anchor from A0, respectively, with the remaining coefficients unchanged. For A0--A3, the copy alarm does not stop training, so all ablations are evaluated at the final epoch.

To assess training variability in the key PIN--Direct Mix comparison, we repeat both methods with two additional seeds.

\section{Results}
\label{sec:results}

\subsection{External benchmark results}
\label{sec:external-results}

\begin{table*}[t]
\caption{\textbf{Supervision-strategy comparison.}}
\label{tab:2}
\centering
\fontsize{9}{10.8}\selectfont
\setlength{\tabcolsep}{2.5pt}
\renewcommand{\arraystretch}{1.13}
\sbox{\tablebox}{\begin{tabular}{lrrrrrrrrrrrr}
\toprule
 & \multicolumn{3}{c}{Image} & \multicolumn{4}{c}{Surface geometry} & \multicolumn{3}{c}{Material fidelity} & \multicolumn{2}{c}{Morphology} \\
\cmidrule(lr){2-4}\cmidrule(lr){5-8}\cmidrule(lr){9-11}\cmidrule(lr){12-13}
Method & PSNR$\uparrow$ & SSIM$\uparrow$ & LPIPS$\downarrow$ & ASSD$\downarrow$ & HD95$\downarrow$ & SDice$\uparrow$ & HRcov$\uparrow$ & Precision$\uparrow$ & Unsup.$\downarrow$ & \shortstack{VF error$\downarrow$} & BV/TV$\downarrow$ & Tb.Th$\downarrow$ \\
\midrule
Synth-only & 15.400 & 0.3393 & 0.5330 & 4.080 & 11.599 & 0.3723 & 0.2808 & 0.8152 & 3.01 & 0.0636 & 0.0436 & 7.365 \\
Real-only & 16.355 & \textbf{0.4143} & 0.4909 & 4.521 & 12.815 & 0.4682 & 0.3920 & 0.8299 & 2.29 & 0.0707 & 0.0374 & 1.720 \\
Direct Mix & \textbf{16.394} & 0.4107 & 0.4825 & 4.509 & 12.646 & 0.4735 & 0.3921 & 0.8294 & 2.26 & 0.0720 & 0.0349 & 1.646 \\
PIN & 15.090 & 0.3878 & \textbf{0.4514} & \textbf{0.995} & \textbf{3.473} & \textbf{0.7046} & \textbf{0.6137} & \textbf{0.8574} & \textbf{1.60} & \textbf{0.0411} & \textbf{0.0185} & \textbf{1.425} \\
\bottomrule
\end{tabular}
}
\typeout{PIN_TABLE_2_WIDTH=\the\wd\tablebox; LIMIT=\the\textwidth}
\usebox{\tablebox}
\par\vspace{3pt}{\footnotesize All strategies use the shared SR backbone. Metric conventions follow Table~\ref{tab:1}. \textbf{Bold}: best.}
\end{table*}

\begin{table*}[t]
\caption{\textbf{Component ablation and optional correction.}}
\label{tab:3}
\centering
\fontsize{9}{10.8}\selectfont
\setlength{\tabcolsep}{2.5pt}
\renewcommand{\arraystretch}{1.13}
\sbox{\tablebox}{\begin{tabular}{lrrrrrrrrrrrr}
\toprule
 & \multicolumn{3}{c}{Image} & \multicolumn{4}{c}{Surface geometry} & \multicolumn{3}{c}{Material fidelity} & \multicolumn{2}{c}{Morphology} \\
\cmidrule(lr){2-4}\cmidrule(lr){5-8}\cmidrule(lr){9-11}\cmidrule(lr){12-13}
Method & PSNR$\uparrow$ & SSIM$\uparrow$ & LPIPS$\downarrow$ & ASSD$\downarrow$ & HD95$\downarrow$ & SDice$\uparrow$ & HRcov$\uparrow$ & Precision$\uparrow$ & Unsup.$\downarrow$ & \shortstack{VF error$\downarrow$} & BV/TV$\downarrow$ & Tb.Th$\downarrow$ \\
\midrule
A0 & 15.062 & 0.3872 & 0.4520 & \textbf{0.988} & \textbf{3.360} & 0.7012 & 0.6096 & 0.8565 & 1.64 & 0.0409 & 0.0197 & 1.655 \\
A1 & 14.971 & 0.3535 & 0.4848 & 1.028 & 3.598 & 0.6944 & 0.5989 & 0.8556 & 1.67 & 0.0374 & 0.0201 & 1.982 \\
A2 & 14.523 & 0.3564 & 0.4625 & 1.071 & 3.698 & 0.6809 & 0.5873 & 0.8456 & 1.96 & \textbf{0.0327} & 0.0269 & 2.467 \\
A3 & \textbf{15.967} & \textbf{0.4096} & 0.4586 & 2.857 & 9.060 & 0.5381 & 0.4498 & 0.8336 & 2.20 & 0.0558 & 0.0430 & 2.276 \\
PIN & 15.090 & 0.3878 & \textbf{0.4514} & 0.995 & 3.473 & \textbf{0.7046} & \textbf{0.6137} & \textbf{0.8574} & \textbf{1.60} & 0.0411 & \textbf{0.0185} & \textbf{1.425} \\
\bottomrule
\end{tabular}
}
\typeout{PIN_TABLE_3_WIDTH=\the\wd\tablebox; LIMIT=\the\textwidth}
\usebox{\tablebox}
\secondarytablenotes[A0 retains all three core paths with correction off; A1--A3 remove synthetic supervision, the HR anchor, and the LR anchor, respectively. PIN includes the optional correction.]
\end{table*}

Table~\ref{tab:1} summarizes the representative external benchmark. Across all 30 external configurations, PIN achieves the best full-volume LPIPS (0.4514) and outperforms all synthetic-trained baselines in SSIM while maintaining comparable PSNR. Although real-trained baselines reach higher PSNR and SSIM by directly fitting registered target intensities, they suffer from severe surface-geometry degradation.

PIN is the only configuration in this benchmark with finite surface distances in all 40 ROIs. Against synthetic-trained mDCSRN, MTVNet, and RRDBNet3D, PIN reduces ASSD by 42.3\%, 45.9\%, and 49.5\%, and HD95 by 40.7\%, 43.8\%, and 46.9\%, respectively. PIN also leads all 30 external configurations in Surface Dice (Fig.~\ref{fig:3}a, left) and HR-covered at 0.5, 1, and 2~\HRvox{} (Suppl. Sec.~\siref{sec:external}). Paired comparisons on common finite support preserve this surface-distance advantage (Suppl. Sec.~\siref{sec:common-support}). The illustrative views in Fig.~\ref{fig:3}b are consistent with the surface measurements: PIN better preserves thin separations and porous interface patterns visible in the registered HR, while several baselines appear smoother or merge nearby structures.

For material fidelity, PIN leads all 30 baselines in both voxel Precision and Unsupported solid volume (cutting false-positive unsupported material to 1.60\%), while maintaining competitive overall VF accuracy. To quantify reference-pore erasure, we identify HR-defined void components whose voxels are entirely labeled as solid in SR. In the three-dimensional evaluation, only 10.4\% of eligible reference-pore volume is fully erased by PIN, compared with 37.3--72.9\% across all 30 external configurations on the same 40 ROI cores. Such pore filling goes unpenalized by reference-solid coverage metrics (Recall and Missing), allowing the baselines to appear superior on these measures despite their substantially poorer preservation of porous structure (Suppl. Secs.~\siref{sec:external-results-detail} and~\siref{sec:material-topology}).

On the three predefined bone-profile units, PIN's BV/TV absolute error (0.0185) is 52.2--53.4\% lower than those of synthetic-trained RRDBNet3D (0.0388) and mDCSRN (0.0394), and real-trained SuperFormer (0.0397). On the two whole held-out bone units, PIN's absolute Tb.Th error is 1.425~\HRvox{}, compared with 1.465--5.971~\HRvox{} for the external configurations in Table~\ref{tab:1} (Suppl. Sec.~\siref{sec:morphology}). PIN yields finite local solid-diameter measurements on all eight applicable units, compared with one to seven for external configurations. On paired common supports of six or seven units, it more than halves the errors of the three strongest ASSD baselines (Suppl. Tables~\siref{tab:si-compact-thickness-eligibility}--\siref{tab:si-compact-paired-thickness}).

Threshold-sensitivity analysis confirms that PIN consistently maintains its leading interface-recovery performance across five segmentation-threshold offsets and three distance tolerances (Suppl. Sec.~\siref{sec:thresholds}).

\subsection{Supervision strategy results}
\label{sec:strategies}

Under the shared SR backbone, conventional supervision strategies expose a clear trade-off (Table~\ref{tab:2}). While Synth-only struggles to generalize to acquired LR inputs, Real-only improves pixelwise PSNR/SSIM at the cost of severe interface oversmoothing (yielding the worst ASSD and HD95). Notably, naively combining both point-target losses (Direct Mix) barely improves surface distance over Real-only (ASSD of 4.509 vs. 4.521~\HRvox{}), indicating that direct voxelwise regression on registered pairs dominates and blurs fine interfaces even when mixed with synthetic targets.

By decoupling physical-pair supervision into complementary roles, PIN reduces ASSD by 77.9\% and boosts Surface Dice by 48.8\% over Direct Mix, while achieving the best LPIPS and bone-morphology accuracy across all strategies. In material recovery, PIN lowers absolute VF error by 42.9\%, cuts unsupported solid volume by nearly 30\%, and reduces the erased pore-volume fraction from 53.4\% to 10.4\% (Suppl. Sec.~\siref{sec:material-topology}). PIN's higher material selectivity is accompanied by lower solid Recall than Direct Mix (0.7736 vs. 0.9466). Much of this coverage deficit lies close to reconstructed solid: Missing falls from 22.64\% at 0.5~\HRvox{} to 5.42\% at 2~\HRvox{} (Suppl. Tables~\siref{tab:si-compact-strategy-voxel-reverse-coverage}--\siref{tab:si-compact-strategy-material}).

These gains remain stable across three random seeds (Suppl. Sec.~\siref{sec:seeds}): PIN consistently outperforms Direct Mix in Surface Dice ($0.7042\pm0.0023$ vs. $0.4733\pm0.0033$) and preserves the same method rankings across all image, surface, material, and BV/TV metrics, with only a minor reversal in Tb.Th at seed~2.

\subsection{Component ablation results}
\label{sec:components}

Comparing full PIN with correction-off A0 shows nearly identical reconstruction metrics (Table~\ref{tab:3}), confirming that the three-path core drives the primary performance gains, while the optional safeguard supports further development and adaptation as a diagnostic monitor during supervision design (e.g., flagging abnormal high-frequency copying in A2; Suppl. Secs.~\siref{sec:si-components} and~\siref{sec:diagnostics}).

Removing individual core paths from A0 reveals their distinct roles (Table~\ref{tab:3}). A1 and A2 each retain the LR anchor and one of the two HR-supervision paths. Dropping synthetic supervision (A1) moderately increases LPIPS and raises Tb.Th error by 19.7\%, showing that exact synthetic pairs still provide a useful correspondence prior. Removing the HR anchor (A2) causes a much sharper drop in structural fidelity: BV/TV and Tb.Th errors jump by 36.9\% and 49.0\%, accompanied by visible interface artifacts across bone and wood specimens (Suppl. Fig.~\siref{fig:a2-comparison}). Because the LR anchor operates only after low-pass proxy projection, high-frequency HR variations remain underconstrained without HR-anchor replay. Notably, A2 achieves the lowest aggregate VF error in Table~\ref{tab:3} despite its severely degraded interfaces, demonstrating that overall volume balance does not guarantee spatial structural accuracy.

Finally, removing the LR anchor (A3) causes the largest deterioration in geometric accuracy: ASSD increases from 0.988 to 2.857~\HRvox{}, Surface Dice drops by 23.3\%, and BV/TV error more than doubles, despite higher PSNR and SSIM than A0. The mean filtered LR residual also rises by 114.9\% across the test set (Suppl. Sec.~\siref{sec:lr-residual}). Direct observation-domain constraints on the acquired-LR path are therefore critical for anchoring reconstructed geometry to the physical scan.

\section{Discussion}
\label{sec:discussion}

For real-world volumetric CT SR, physically acquired LR--HR pairs are not inherently better supervision than synthetic pairs. Their value lies in acquisition-specific information, whose contribution depends on the supervisory role assigned to each observation. Real-only achieves strong image agreement but weak interface recovery, while Direct Mix retains surface behavior close to Real-only despite combining synthetic and physical point targets (Sec.~\ref{sec:strategies}). PIN assigns registered HR to structural teaching and acquired LR to observation compatibility, demonstrating the importance of complementary supervision roles for reliable fine-structure recovery.

CT super-resolution should not be evaluated by PSNR and SSIM alone~\cite{Jhuboo2022BoneMetrics,Chan2023BoneDiffusion,Klos2025LatticeCT}. High PSNR/SSIM and voxel overlap can still mask poor porous-interface reconstruction. The material diagnostics show that PIN substantially reduces reference-pore erasure and retains more opposing-face reference pore pathways, alongside stronger interface localization and material selectivity (Suppl. Sec.~\siref{sec:material-topology}). Surface geometry, material balance, and opposing-face pore connectivity capture complementary structural properties, while BV/TV and Tb.Th errors show how reconstruction affects measurements of bone morphology. For scientific volumetric CT SR, these structural and morphological evaluations provide the primary evidence of reconstruction quality, with PSNR and SSIM serving as supporting metrics.

Tailoring degradation operators to their supervision roles may help preserve complementary supervision. Relying on a single learned operator could propagate its errors and blind spots across paths, steering the SR model toward reconstructions favored by that operator. PIN therefore retains simple trilinear downsampling for synthetic supervision and uses a learned HR-to-LR proxy $D_\phi$ for its real-pair anchors. On this dataset, the learned $D_\phi$ achieves lower filtered LR error than fixed downsampling operators (Suppl. Sec.~\siref{sec:proxy-validation}).

A practical limitation of PIN is that training requires registered physical LR--HR pairs, which can be costly and difficult to acquire, although inference requires only acquired LR. Its robustness across different scanners, acquisition protocols, and registration quality remains to be established.

\section{Conclusion}
\label{sec:conclusion}

In real-world volumetric CT super-resolution, registered physical scans contain valuable acquisition-specific information but remain distinct observations rather than exact voxelwise targets. PIN addresses this mismatch by combining exact synthetic supervision with two physical-pair paths: registered-HR teaching via pair-conditioned spectral replay and acquired-LR regularization via affine-calibrated low-pass projection. Across 30 external baselines and same-backbone controls, PIN improves interface localization, substantially reduces reference-pore erasure, and lowers bone-morphology errors relative to direct real-pair regression and naive mixing. More broadly, the divergence between PSNR/SSIM and structural accuracy shows that scientific CT SR requires evaluation grounded in geometry, pore retention, and morphology.

\FloatBarrier
\label{main-text-end}
\section*{LLM Usage Statement}

During the preparation of this work, the authors used ChatGPT (OpenAI) to edit text and figures, check citations, and analyze experimental results. Cursor (SpaceX) also assisted with research code development. The authors reviewed and revised the resulting materials and code, verified the reported results, and take full responsibility for the final manuscript.

{\small
\setbiblabelwidth{50}
\bibliographystyle{ieeenat_fullname}
\bibliography{references}
}
\onecolumn
\makeatletter
\let\PINbackout\Hy@backout
\renewcommand*\Hy@backout[1]{%
  \begingroup\let\@auxout\@mainaux\PINbackout{#1}\endgroup}
\makeatother
\setcounter{section}{0}
\setcounter{equation}{0}
\setcounter{table}{0}
\setcounter{figure}{0}
\renewcommand{\theHsection}{supp.\arabic{section}}
\renewcommand{\theHsubsection}{\theHsection.\arabic{subsection}}
\renewcommand{\theHsubsubsection}{\theHsubsection.\arabic{subsubsection}}
\renewcommand{\theHequation}{supp.\arabic{equation}}
\renewcommand{\theHtable}{supp.\arabic{table}}
\renewcommand{\theHfigure}{supp.\arabic{figure}}
\renewcommand{\thesection}{S\arabic{section}}
\renewcommand{\thesubsection}{\thesection.\arabic{subsection}}
\renewcommand{\theequation}{S\arabic{equation}}
\renewcommand{\thetable}{S\arabic{table}}
\renewcommand{\thefigure}{S\arabic{figure}}
\setlength{\LTpre}{5pt}
\setlength{\LTpost}{8pt}
\makeatletter
\renewcommand{\cvprsect}[1]{\cvprsection[#1]{\texorpdfstring{\hskip -1em.~}{}#1}}
\patchcmd{\l@section}{\setlength\@tempdima{1.5em}}{\setlength\@tempdima{2em}}{}{\PackageError{supplementary}{Contents spacing patch failed}{}}
\makeatother
\pdfbookmark[0]{Supplementary Material}{supplement-start}
\addtocontents{toc}{\protect\setcounter{tocdepth}{1}}
\begin{center}
  {\Large\bfseries \thetitle\par}
  \vspace{0.5em}
  {\large\bfseries Supplementary Material\par}
\end{center}
This supplement provides PIN's implementation details and the analyses supporting the main manuscript. Sections S1--S2 define training, evaluation, aggregation, and support. Sections S3--S6 report the complete external inventory, supervision strategies and components, repeated seeds, focused threshold and morphology checks, and observation-domain diagnostics. Section~\ref{sec:a2-gallery} presents the A2--A0--PIN qualitative comparison after HR-anchor removal.

The main results use seed~0. Additional PIN and Direct Mix seeds describe training variability in image, surface, material, and main morphology results on their predefined evaluation supports. ROI means are nested within evaluation units. Image quality, surface geometry, material recovery, and morphology form four complementary evaluation groups. The LR anchor's observation-compatibility diagnostics have a separate mechanism-level role; registered HR is a separate physical acquisition.

\setcounter{tocdepth}{1}
\tableofcontents
\clearpage
\section{Implementation details and training protocol}
\label{sec:implementation}

\subsection{Data, sampling, and shared SR model}
Training uses the PIN split derived from VoDaSuRe~\citesupp{si-ref1}, with 13 volumes and 32 patch indices per volume. A Femur\_74 volume acquired at 80~kV is held out for validation and excluded from training. VoDaSuRe holds out whole femur and vertebrae test scans, while other specimens are split axially between training and test~\citesupp{si-ref1}. These axial splits do not establish specimen-disjointness for every material.

This gives 416 nominal dataset entries. With two distributed ranks and batch size one per rank, each active data stream's rank-local loader contains 208 batches. Each epoch makes 208 optimizer-update attempts, each accumulating two successive microbatches per rank and drawing once from each active stream per microbatch; the iterators cycle when exhausted. The epoch therefore consumes 416 batches per rank per stream, or 832 crop presentations per stream across both ranks, with four crop presentations per stream per update attempt. Full PIN has synthetic and physical-pair streams; the HR and LR anchors share the physical-pair crop.

The shared SR model has a 3D residual-dense encoder and a LIIF-style coordinate decoder~\citesupp{si-ref12}. The encoder has 96 features and ten residual-dense blocks, each containing four dense layers with growth rate 64. The decoder has three 384-unit hidden layers. Local ensemble, 3D feature unfolding, and cell decoding are enabled. Each LR input is encoded into a feature grid at LR resolution. At each HR query, the decoder combines local LR features with relative coordinates and cell sizes to predict intensity. The three paths are evaluated separately using shared encoder and decoder parameters.

At the evaluated fourfold scale, a $48^3$ LR crop corresponds to a $192^3$ HR grid over the same physical field. Write $F_\theta(z;\mathcal P)$ for its prediction array at a coordinate list or grid $\mathcal P$, and $F_\theta(z)$ for dense-grid reconstruction. Each synthetic training crop supplies $Q=32{,}768$ continuous HR queries $q_i=(q_{i,z},q_{i,y},q_{i,x})$ drawn uniformly from $[-1,1)^3$. Here $y_{\mathrm{real}}$ denotes the HR training crop being queried. For HR dimensions $(n_z,n_y,n_x)=(D,H,W)$ and axis $\alpha\in\{z,y,x\}$, define the zero-based voxel-index map and target entries by
\[
\pi_\alpha(q_i)=\clip_{[0,n_\alpha-1]}\!\left(\left\lfloor\frac{n_\alpha(q_{i,\alpha}+1)}{2}\right\rfloor\right),
\qquad
(y_{\mathcal Q})_i=y_{\mathrm{real}}[\pi_z(q_i),\pi_y(q_i),\pi_x(q_i)].
\]
The target is obtained by this integer lookup. The cell width is determined by the HR grid: $(2/D,2/H,2/W)$ for an HR crop of size $D\times H\times W$. For matching prediction and target arrays $p,t$ with $N$ scalar entries, the normalized $L_1$ discrepancy is $N^{-1}\|p-t\|_1$, where $\|\cdot\|_1$ sums absolute scalar entries. The tensor reduction includes all batch, query or grid, and channel entries; this run uses one output channel and a per-GPU batch size of one.

The routine for the two anchors additionally samples a $32^3$ LR subcube and the corresponding $128^3$ HR subcube within the larger pair. Denote this paired anchor subcrop by $(x_b,y_b)$, with LR and HR grids $\mathcal G_{\mathrm{LR},b}$ and $\mathcal G_{\mathrm{HR},b}$. Their voxel counts are $N_{\mathrm{LR}}=|\mathcal G_{\mathrm{LR},b}|=32^3$ and $N_{\mathrm{HR}}=|\mathcal G_{\mathrm{HR},b}|=128^3$. Both replay and acquired-LR predictions are rendered on all $128^3$ HR voxel centers, whose coordinates along an axis of length $n$ are $(2k+1)/n-1$, $k=0,\ldots,n-1$, with cell width $2/n$. Thus the point-query crop and the LR-anchor calibration crop have different sizes.

The outer $48^3$ LR/$192^3$ HR sampler requests foreground with probability 0.95. Foreground occupies at least 0.01 of the full LR-equivalent crop mask: acquired LR intensities above 0.05 define the real-stream mask, and $4^3$ HR-block means above 0.05 define the synthetic-stream proxy mask. Up to 64 uniformly drawn in-bounds integer origins are tested for the requested class. If none matches, selection uses the requested coarse-grid origin pool (stride 8 LR voxels, at most 4,096 origins per class), then the other class pool if needed, then a fresh uniform origin if both pools are empty. The $32^3$ anchor sampler makes its own 0.95 foreground request and tests up to 64 origins using the fraction of all LR subcrop voxels that are finite and above 0.05, with foreground defined by a fraction of at least 0.01. It returns the first matching origin; after an unsuccessful search it uses the last foreground candidate if available, otherwise the last background candidate, with the zero origin as a final fallback. The foreground flag for foreground-only fit QC is recomputed from the selected anchor subcrop, so fallback can change the realized foreground proportion.

\subsection{Shared learned HR-to-LR proxy}
Both anchors use the same learned 3D convolutional HR-to-LR proxy (\texttt{discrete\_cnn3d}). The proxy uses fourfold scale, 64 features, 16 blocks, and four checkpoint segments. During PIN training, proxy parameters and evaluation statistics are frozen, while derivatives with respect to its input are retained on the SR-to-LR path. The HR-anchor replay base is evaluated under stop-gradient. Freezing the proxy therefore preserves its fixed comparison operator while allowing the LR anchor to update the shared SR model.

Here $\sg$ denotes automatic-differentiation stop-gradient: it preserves the value of its argument and sets its derivative with respect to that argument to zero.

The proxy was pretrained on the 13 registered VoDaSuRe training volumes, with Femur\_74 held out for validation, by minimizing voxelwise $L_1$ between its HR-to-LR projection and acquired LR. The 100-epoch run used Adam with an initial learning rate of $3\times10^{-4}$ and cosine decay to $10^{-6}$, with a global batch size of four across two GPUs. PIN freezes the terminal epoch-100 checkpoint; the validation-best checkpoint occurred at epoch~97. The operator comparison in Sec.~\ref{sec:proxy-validation} evaluates this fixed operator.

\subsection{Exact synthetic supervision and HR anchor}
Synthetic supervision is jointly optimized with the HR and LR anchors. Fourfold trilinear downsampling $D_{\mathrm{syn}}$ constructs the LR input receiving its registered-HR point target, using \texttt{align\_corners=False} and clipping its output to $[0,1]$. The active synthetic term is $N_{\mathrm{syn}}^{-1}\|p-t\|_1$ over its $N_{\mathrm{syn}}=Q$ query predictions $p$ and integer-lookup targets $t$, without an added gradient loss.

For the anchor formulas below, abbreviate $x=x_b$ and $y=y_b$. For HR-anchor replay, let $L_0=\sg(D_\phi(y))$ and $r=x-L_0$ on the LR subcube used by the anchors. PIN retains $\mu_r=N_{\mathrm{LR}}^{-1}\sum_{i=1}^{N_{\mathrm{LR}}}r_i$ and measures four band strengths $s_r=\operatorname{Stats}_4(r-\mu_r)\in\mathbb R_{\geq0}^{4}$. The spectral estimator applies a separable, nonperiodic 3D Hann window $W$ and an orthonormal float32 real FFT $\mathcal F_{\mathrm R}$. For crop side $S=32$, $W$ is the product of the one-dimensional weights $\tfrac12[1-\cos(2\pi k/(S-1))]$, $k=0,\ldots,S-1$. With $f_z,f_y,f_x$ in cycles per LR voxel, normalized radial frequency is
\begin{equation}
\nu=\frac{\sqrt{f_z^2+f_y^2+f_x^2}}{\sqrt{3}/2}.
\label{eq:si-frequency}
\end{equation}
The bands are $[0,0.25)$, $[0.25,0.5)$, $[0.5,0.75)$, and $[0.75,1]$. Let $M_k(\omega)$ be their masks on the real-FFT half spectrum $\Omega_{\mathrm R}$, with the last upper edge implemented as $1+10^{-6}$. The Hermitian weight $\eta(\omega)$ is one at $f_x=0$ and the even-size Nyquist slice, and two elsewhere. For a supplied LR residual array $a$,
\[
\begin{aligned}
\mathcal E_k(a)&=\left[\frac{1}{S^3}\sum_{\omega\in\Omega_{\mathrm R}}
 \eta(\omega)M_k(\omega)|\mathcal F_{\mathrm R}(W\odot a)(\omega)|^2\right]^{1/2},\\
\operatorname{Stats}_4(a)&=(\mathcal E_0(a),\mathcal E_1(a),\mathcal E_2(a),\mathcal E_3(a)).
\end{aligned}
\]
Here $\odot$ is entrywise multiplication. The four windowed radial-band RMS contributions provide the magnitude summaries retained for replay. DC is excluded from the synthesized replay residual because the retained mean is added separately.

The replay mapping $\widetilde r=\mathcal R(s_r;\xi_b)$ uses the seeded draw $\xi_b$ to generate a mean-subtracted float32 Gaussian array $n$ on CPU. With $\epsilon=10^{-8}$, let $\varphi=\mathcal F_{\mathrm R}(W\odot n)/\max(|\mathcal F_{\mathrm R}(W\odot n)|,\epsilon)$ and $M_k^0(\omega)=M_k(\omega)\mathbf 1[\omega\ne0]$. Here the new draw supplies phase, while the Hermitian-weighted bin-count factor sets the per-band amplitude normalization. The initial target spectrum and its inverse are
\[
Y(\omega)=\varphi(\omega)\sum_{k=0}^{3}(s_r)_k
 \sqrt{\frac{S^3}{\max(\sum_{\omega'}\eta(\omega')M_k^0(\omega'),\epsilon)}}\,M_k^0(\omega),
\qquad T=\mathcal F_{\mathrm R}^{-1}(Y).
\]
Thus $T$ targets the windowed field $W\odot R$ used by the estimator. The finite updates below compensate for this window, while mean removal keeps the sampled component separate from $\mu_r$, which is added back later. Starting at $R^{(0)}=T$, the generator takes 12 window-consistent Landweber steps,
\[
R^{(t+1)}=\mathcal C\!\left(R^{(t)}+1.4W\odot(T-W\odot R^{(t)})\right),
\qquad [\mathcal C(a)]_i=a_i-\frac{1}{N_{\mathrm{LR}}}\sum_{j=1}^{N_{\mathrm{LR}}}a_j.
\]
Finite window compensation can leave a mismatch in the measured band strengths. When needed, the correction below rescales the current windowed spectrum once by the target-to-achieved strength ratios, without a new random draw. If any band has $|\mathcal E_k(R^{(12)})-(s_r)_k|/\max((s_r)_k,\epsilon)>0.02$, it replaces $T$ by
\[
\mathcal F_{\mathrm R}^{-1}\!\left[\sum_{k=0}^{3}M_k^0\,\mathcal F_{\mathrm R}(W\odot R^{(12)})
 \frac{(s_r)_k}{\max(\mathcal E_k(R^{(12)}),\epsilon)}\right]
\]
and takes six further steps of the same update, starting from $R^{(12)}$. The final $R$ is $\widetilde r$; a zero target band contributes zero spectral amplitude. These finite steps define the replay construction before clipping, without imposing an exact post-clipping band-strength equality. Its dedicated CPU random generator is keyed by the SHA-256 hash of the global seed, epoch, seed counter $2U+m$, rank, volume key, and LR crop origin, where $U$ is the trainer's global update counter and $m\in\{0,1\}$ is the accumulation microbatch index. The first eight digest bytes are interpreted as an unsigned little-endian integer and reduced modulo $2^{63}$. Batches in the seed-0 configuration provide no volume identifier, so the volume-key field uses the fixed literal $\texttt{unknown}$. This field only seeds replay; training volumes are selected separately by dataset index. The input is
\begin{equation}
x_{\mathrm{rep}}=\clip_{[0,1]}(L_0+\mu_r+\widetilde r).
\label{eq:si-replay}
\end{equation}
Band matching describes the residual construction before clipping. The spatial realization is newly sampled. If the complete measured residual were used instead, $L_0+r=x$ before clipping, restoring the acquired LR as the registered-HR point-supervised input. Replay avoids that identity while retaining selected pair-conditioned statistics.

\subsection{Detached affine calibration and filtering}
Each batch for the two anchors fits $A_b(z)=a_bz+b_b$ between $D_\phi(y)$ and acquired $x$ on its $32^3$ LR subcube. For flattened projection entries $d_i=[D_\phi(y)]_i$ and acquired entries $x_i$, the fitting mask is $m_i=\mathbf 1[\operatorname{finite}(d_i),\operatorname{finite}(x_i),0.05<x_i<0.98]$. The fit uses float32 Huber iteratively reweighted least squares with 25 iterations and $\delta=0.05$. With design rows $X_i=(d_i,1)$, $\mathsf W^{(0)}=\operatorname{diag}(m_i)$, and $I_2$ the $2\times2$ identity, the updates for finite fitting arrays and $t=0,\ldots,24$ are
\[
\begin{aligned}
\beta^{(t+1)}&=\operatorname{solve}\!\left(X^\top\mathsf W^{(t)}X+10^{-8}I_2,\;X^\top\mathsf W^{(t)}x\right),\\
\mathsf W^{(t+1)}_{ii}&=m_i\min\!\left(1,\frac{0.05}{\max(|x_i-X_i\beta^{(t+1)}|,10^{-12})}\right).
\end{aligned}
\]
Before these updates, validity requires at least 64 masked voxels and predictor variance greater than $10^{-12}$. A failed solve or nonfinite normal matrix or coefficients makes the fit invalid. After a successful solve, the normal matrix used in the last solve must have condition number at most $10^8$, and the slope must lie within $[0.1,10]$. A valid fit has $\chi_b=1$ and $(a_b,b_b)=\sg(\beta^{(25)})$; an invalid fit has $\chi_b=0$ and finite placeholders $(a_b,b_b)=(1,0)$. Thus both fitted coefficients are detached.

The fit estimates a crop-dependent intensity scale and offset; it does not deliver a spatial HR point target to the acquired-LR prediction. The differentiable proxy projection of that prediction is calibrated as $\widehat L=A_b(D_\phi(h_{\mathrm{meas}}))$, where $h_{\mathrm{meas}}=F_\theta(x;\mathcal G_{\mathrm{HR},b})$, and compared with acquired LR after identical separable 3D Gaussian filtering. The filter has $\sigma=1$ LR voxel, radius three, seven taps per axis, and zero padding. This comparison emphasizes a lower-frequency LR band. The fitting mask controls coefficient estimation only; the active LR objective is an unmasked mean over every voxel in the filtered LR crop. In the raw-loss notation,
\[
\ell_{\mathrm{LR}}(h)=
\begin{cases}
\frac{1}{N_{\mathrm{LR}}}\|G_\sigma\ast A_b(D_\phi(h))-G_\sigma\ast x\|_1,&\chi_b=1,\\
0,&\chi_b=0.
\end{cases}
\]

An invalid fit makes the LR anchor and all correction components exactly zero for that crop while the HR anchor stays connected. For seed~0, the foreground-only quality-control summary uses the last accumulation microbatch of each update attempt at epochs~2--60. On this QC support, the maximum invalid-fit fraction is 2.12\% at epoch~15 on rank~0 (4/189) and 2.48\% at epoch~44 on rank~1 (5/202). At the final epoch, the rank-specific counts are 1/194 and 2/194. Each denominator counts foreground checks on that rank in that epoch. The principal reasons are fewer than 64 valid voxels or a fitted slope outside $[0.1,10]$.

\subsection{Calibrated objective and accumulation}
Let $\ell_{\mathrm{HR}}=N_{\mathrm{HR}}^{-1}\|F_\theta(x_{\mathrm{rep}};\mathcal G_{\mathrm{HR},b})-y\|_1$ and $\ell_{\mathrm{LR}}=\ell_{\mathrm{LR}}(h_{\mathrm{meas}})$ be the raw normalized $L_1$ terms for the HR anchor and LR anchor, respectively. The anchor scale factors were determined by an offline, train-only calibration using two crops from each of the 13 training volumes at four SR model states (104 crop--state evaluations). For each raw anchor loss, its scale factor is the reciprocal of the median of the positive global $L_2$ norms of its gradients with respect to the shared SR parameters $\theta$. The fixed scale factors are
\begin{equation}
\bar L_{\mathrm{HR}}=2.364342842\ell_{\mathrm{HR}},\qquad
\bar L_{\mathrm{LR}}=0.8824529122\ell_{\mathrm{LR}}.
\label{eq:si-calibration}
\end{equation}
The core is $L_{\mathrm{synth}}+0.7\bar L_{\mathrm{HR}}+0.3\bar L_{\mathrm{LR}}$. With two-step accumulation, each micro-backward uses
\begin{equation}
\tfrac12L_{\mathrm{synth}}+0.35\bar L_{\mathrm{HR}}+0.15\bar L_{\mathrm{LR}}
=\tfrac12L_{\mathrm{synth}}+0.8275199947\ell_{\mathrm{HR}}+0.1323679368\ell_{\mathrm{LR}}.
\label{eq:si-microbackward}
\end{equation}
The loss-scale factors and 70/30 anchor coefficients remain fixed during training. These are scalar coefficients, not guaranteed instantaneous gradient-norm fractions. Each factor and the accumulation divisor are applied once.

In particular, the LR scalar supplied to the cotangent calculation already has multiplier $m_{\mathrm{LR}}=0.3c_{\mathrm{LR}}/2=0.13236793683$, with a measurement coefficient of one.

Synth-only uses the synthetic point target. Real-only uses acquired-LR-to-registered-HR point supervision. Direct Mix gives both raw point losses coefficient one, with equal coordinate query counts; that coefficient is not a data-sampling ratio. A0 retains all three calibrated core terms with correction disabled. Correction-off A1, A2, and A3 remove synthetic supervision, the HR anchor, and the LR anchor, respectively, from A0 without renormalizing the retained coefficients. These component controls disable both correction and copy-alarm termination, while retaining the copy thresholds for monitoring. This allows component-removal behavior to remain observable at epoch~60.

\subsection{Optimization and checkpoint selection}
SR training uses Adam with initial learning rate $10^{-4}$, cosine $T_{\max}=60$ and terminal rate $10^{-6}$, batch size one per GPU, two-GPU distributed training, automatic mixed precision, gradient clipping at 1.0, and two-step accumulation. Epoch validation does not use metric-based early stopping; full PIN retains the separate copy-alarm termination safeguard described in Sec.~\ref{sec:correction}. The sampling schedule above gives 12,480 optimizer-update attempts over 60 epochs. Seed~0 completes 60/60 epochs in 117.1~hours. Six AMP skips leave 12,474 successful optimizer updates. The final zero-based update index is 12,473, whereas the checkpoint stores the post-increment count 12,474.

Evaluation uses the final epoch-60 checkpoint; the validation-best checkpoint is from epoch~55. Checkpoints are saved at epochs 50, 52, 54, 56, 58, and 60. A0--A3 also complete epoch~60, with final update counters of 12,474, 12,475, 12,476, and 12,475, respectively. These runs use the same optimizer, schedule, accumulation, and final-checkpoint convention.

PIN seeds~1 and~2 repeat the same training protocol. Their evaluation results and support are reported in Sec.~\ref{sec:seeds}.

\subsection{Optional bounded cotangent correction and diagnostic safeguards}
\label{sec:correction}
The LR anchor evaluates predictions after HR-to-LR proxy projection, affine calibration, and Gaussian filtering, leaving some HR-space structural variations weakly constrained. The optional correction therefore uses four frozen diagnostics to provide bounded output-space guidance, alongside a separate copy-alarm monitoring and persistent-termination mechanism. The diagnostics assess geometric consistency, high-frequency copying, structural responses, and spectral-energy behavior. The formulation below specifies the optional safeguard used in full PIN; the three-path core can be trained without cotangent correction, as in A0 (Sec.~\ref{sec:si-components}).

The structural diagnostic $\Phi_{\mathrm{struct}}$ is a context-conditioned energy model pretrained on registered LR--HR pairs from the 13-volume training split. Its acceptance calibration used the Femur\_74 validation volume. Its pretraining construction defines a positive example as a trilinearly upsampled registered-LR context paired with the registered-HR residual relative to that context. Negative examples retain the context and use either the positive residual augmented with a degradation-derived perturbation or a residual swapped from a matched training positive. These two negative families are sampled equally. A mean pairwise softplus ranking objective with margin one favors lower energy for positive examples than for negatives. Pretraining ran for 10,000 steps on two ranks, with 64 ranking pairs from each negative family per rank (256 pairs per optimizer update). The structural diagnostic remains frozen during PIN training.

The spatial normalization center is the median of the within-volume positive-example medians. The MAD is the median across volumes of the within-volume median absolute deviations from that common center; the calibration support comprises 13 volumes and 416 patches. The spectral normalization uses per-coordinate positive-example medians and MADs with equal volume weight, using 416 views of side 32 and 52 views of side 64, with 13 volumes represented at each scale. These spatial and spectral statistics are fixed before PIN training and remain frozen with the structural model.

For the diagnostics below, let $L_{\mathrm{ref}}=\sg[A_b(D_\phi(y))]$ be the calibrated reference on the LR crop, and let $[z]_+=\max(z,0)$ and $\epsilon=10^{-8}$. The one-sided Huber function for $z\geq0$ is
\[
\rho_\beta(z)=\begin{cases}z^2/(2\beta),&0\leq z\leq\beta,\\z-\beta/2,&z>\beta,\end{cases}
\qquad\beta=0.1.
\]

The geometry component uses six signed finite-difference relations, indexed by $k=(\alpha,d)$ for axis $\alpha\in\{x,y,z\}$ and lag $d\in\{1,2\}$ LR voxels. For an LR array $q$ and site $i$, define
\[
\Delta_k q(i)=(G_\sigma\ast q)(i+d e_\alpha)-(G_\sigma\ast q)(i),\qquad\sigma=1,
\]
where $e_\alpha$ is the unit vector along axis $\alpha$ and $G_\sigma$ is the fixed Gaussian filter defined above. The detached mask $c_{k,i}$ is one when the site and its shifted endpoint lie inside the crop and both acquired-LR intensities are finite and strictly between 0.05 and 0.98; it is zero otherwise. For each relation, let $\sigma_k$ be the detached median absolute deviation of $\Delta_k x(i)$ over $c_{k,i}=1$, floored at $\epsilon$. Here MAD is $\operatorname{median}|t-\operatorname{median}(t)|$, without a 1.4826 multiplier; each median takes the lower middle value for an even number of entries, and empty valid support gives $\sigma_k=\epsilon$. The detached sign and eligibility gate are
\[
\begin{aligned}
s_{k,i}&=\sg[\operatorname{sign}(\Delta_k x(i))],\\
m_{k,i}&=\sg\!\left[c_{k,i}\,\mathbf1[|\Delta_k x(i)|>2\sigma_k]\,
 \mathbf1[\operatorname{sign}(\Delta_k x(i))=\operatorname{sign}(\Delta_k L_{\mathrm{ref}}(i))]\right].
\end{aligned}
\]
The reference votes on sign rather than setting the predicted difference magnitude. The margin and normalized positive excess are
\[
\mu_k=0.15\sigma_k,\qquad
\sigma_{\mathrm{eff},k}=\max(\sigma_k,\sigma_{\mathrm{floor},k}),\qquad
z_{k,i}=\frac{[\mu_k-s_{k,i}\Delta_k\widehat L(i)]_+}{\sigma_{\mathrm{eff},k}}.
\]
The geometry reference scales and floors are calibrated before PIN training using 32 training crops drawn with foreground-biased sampling from the 13-volume training split, excluding validation and test data. For each relation, the reference scale is the median of its per-crop acquired-LR finite-difference MADs, and the floor is their 10th percentile. The resulting values are locked throughout training.

The fixed reference scales and floors are
\begin{center}
\begin{tabular}{crr}
\toprule
$(\alpha,d)$ & $\sigma_{\mathrm{ref},k}$ & $\sigma_{\mathrm{floor},k}$ \\
\midrule
$(x,1)$ & 0.02407899498939514 & 0.012193804979324341 \\
$(x,2)$ & 0.04160907864570618 & 0.020985060930252077 \\
$(y,1)$ & 0.027589667588472366 & 0.013156029582023622 \\
$(y,2)$ & 0.048863932490348816 & 0.02243545949459076 \\
$(z,1)$ & 0.015386998653411865 & 0.009187796711921692 \\
$(z,2)$ & 0.02703644335269928 & 0.0164840966463089 \\
\bottomrule
\end{tabular}
\end{center}
With $N_{\mathrm{geo}}=\sum_{k,i}m_{k,i}$, the compensated scalar and its normalized severity pool all eligible relations:
\[
L_{\mathrm{geo}}=\frac{\sum_{k,i}m_{k,i}(\sigma_{\mathrm{eff},k}/\sigma_{\mathrm{ref},k})\rho_\beta(z_{k,i})}{\max(N_{\mathrm{geo}},1)},\qquad
z_{\mathrm{geo}}=\frac{\sum_{k,i}m_{k,i}z_{k,i}}{\max(N_{\mathrm{geo}},1)}.
\]
The denominator counts eligible relations, including those whose margin is already met. Empty eligible sets or no positive eligible excess give zero loss and gradient. Signs, gates, MAD estimates, floors, and reference scales are detached; differentiation follows only the predicted differences through $\widehat L$.

In the copy diagnostic, $\widehat L=A_b(D_\phi(h_{\mathrm{meas}}))$ is the calibrated LR projection defined above. The operator $B_{b3}$ is the highest radial-band orthonormal real-FFT filter
\[
B_{b3}(a)=\mathcal F_{\mathrm R}^{-1}\!\left(\mathbf 1[0.75\leq\nu\leq1+10^{-6}]\,\mathcal F_{\mathrm R}(a)\right),
\]
where $\mathcal F_{\mathrm R}$ and $\mathcal F_{\mathrm R}^{-1}$ are the orthonormal real FFT and its inverse on the LR crop. This operator applies no Hann window.

For copy, the detached high-frequency residual is $n_{\mathrm{hp}}=\operatorname{sg}[B_{b3}(x-A_b(D_\phi(y)))]$. The coefficient used by the correction is evaluated without a high-gradient exclusion mask:
\[
\widehat\alpha=
\frac{\left\langle B_{b3}\!\left(\widehat L-\operatorname{sg}[A_b(D_\phi(y))]\right),n_{\mathrm{hp}}\right\rangle}
{\|n_{\mathrm{hp}}\|_2^2+\epsilon}.
\]
The inner product and norm include all LR-crop voxels. An optional high-gradient exclusion mask is used only for an auxiliary stop-gradient diagnostic, not for the correction or the fixed-panel copy alarm. The correction's normalized excess is $[(\widehat\alpha-0.2792616636)/(0.4476819009-0.2792616636+\epsilon)]_+$. The upper threshold normalizes the crop-level hinge severity.

The copy thresholds are fixed by a separate training-only, crop-level receiver-operating-characteristic calibration of the high-frequency copy coefficient. Four crops from each of 13 training volumes yield 52 copy-positive and 104 copy-negative scores across the reference reconstruction states. Candidate thresholds are the distinct linearly interpolated empirical quantiles at 0\%, 1\%, \ldots{}, 100\% of the pooled positive and negative scores. Scores at or above a candidate threshold are classified as positive. Among operating points with false-positive rate at most 0.10 and true-positive rate at least 0.70, the warning threshold maximizes the true-positive rate, with ties resolved by the lower false-positive rate. Any remaining tie is resolved by the lower threshold. The upper threshold is the larger of the copy-positive 90th percentile and the warning threshold plus $10^{-4}$. Both thresholds remain fixed during PIN training.

The structural component uses an acquired-LR context $C=\sg[U(x)]$, where $U$ is trilinear upsampling to the $128^3$ HR grid with \texttt{align\_corners=False} and no clipping. The evaluator receives context $C$ and HR residual $h-C$. Let $P_{64}$ extract the central $64^3$ cube (zero-based indices 32--95 along each axis), and let $P_j$, $j=1,\ldots,8$, extract its eight nonoverlapping $32^3$ cubes. Every local cube shares the same central $64^3$ view. Prediction and reference energies are
\[
\begin{aligned}
S_{\mathrm{pred},j}&=\Phi_{\mathrm{struct}}\!\left(P_jC,P_j(h_{\mathrm{meas}}-C);P_{64}C,P_{64}(h_{\mathrm{meas}}-C)\right),\\
S_{\mathrm{ref},j}&=\sg\!\left[\Phi_{\mathrm{struct}}\!\left(P_jC,P_j(y-C);P_{64}C,P_{64}(y-C)\right)\right].
\end{aligned}
\]
The positive excess is $z_{\mathrm{struct}}=[\tfrac18\sum_{j=1}^{8}(S_{\mathrm{pred},j}-S_{\mathrm{ref},j})/s_{\mathrm{SF}}]_+$, with fixed $s_{\mathrm{SF}}=2.478597640991211$. The same detached context is used on both sides; model parameters and normalization statistics are frozen while derivatives through the prediction input are retained.

Within $\Phi_{\mathrm{struct}}$, context and residual are first normalized separately as $(a-\mu)/\max(1.4826\delta,\epsilon)$ using fixed training-positive medians and MADs: $(\mu,\delta)=(0.4603760987520218,0.10077778995037079)$ for context and $(0,0.15583860874176025)$ for residual. The spatial branch receives the two normalized $32^3$ arrays. The two spectral views, at sides $q=32$ and $q=64$, concatenate their context and residual transforms
\[
\mathcal T_q(a)=\log\!\left(1+\left|\operatorname{fftshift}\mathcal F_q(W_q\odot a)\right|^2\right),
\]
where $\mathcal F_q$ is the full orthonormal 3D FFT and $W_q$ is a separable nonperiodic Hann window. Each spectral coordinate is further normalized by its frozen training-positive median and MAD, using denominator $\mathrm{MAD}+\epsilon$. Spatial and spectral encoders use four $3\times3\times3$ convolution--GroupNorm(8)--SiLU stages with strides $(1,2,2,2)$, unit zero padding, and no convolutional bias. Their respective channel widths are $(32,64,128,128)$ and $(16,32,64,64)$; the spectral encoder is shared across the two scales. Global average pooling gives 128 spatial and two 64-entry spectral embeddings. Their concatenation is mapped to one patch energy by a $256\to128\to1$ fully connected network with an intermediate SiLU.

The energy component compares four radial-band RMS values: $E_{\mathrm{pred},k}=[\operatorname{Stats}_4(\widehat L-x)]_k$ and \newline $E_{\mathrm{ref},k}=[\sg(\operatorname{Stats}_4(A_b(D_\phi(y))-x))]_k$, for band index $k=0,\ldots,3$. These residuals are not mean-subtracted. Its normalized excess is \newline $z_k=[(E_{\mathrm{pred},k}-E_{\mathrm{ref},k}-\kappa s_k)/(s_k+\epsilon)]_+$, with $\kappa=1.4856691903$ and fixed \newline $s_k=(0.0048675425,0.0068104388,0.0034260806,0.0004298408)$. It averages the per-band one-sided Huber values with $s_k/0.00414681155$ compensation. The correction components use Huber parameter 0.1 and fixed reference-scale compensation.

The four fixed band scales $s_k$ originate from an earlier residual-band MAD calibration. The excess factor $\kappa$ was determined as the larger of two fold-specific thresholds calibrated using 156 crops from 13 training volumes. The compensation reference is the median of the four band scales. These constants were fixed before PIN training. The structural reference scale reuses the MAD from a separate calibration using 32 training-pair crops. These reference values remain fixed alongside the frozen spatial and spectral normalization statistics.

Let $z_{\mathrm{copy}}$ denote the normalized copy excess above. The differentiable diagnostic scalars are
\[
\begin{aligned}
L_{\mathrm{copy}}&=\rho_\beta(z_{\mathrm{copy}}),\\
L_{\mathrm{struct}}&=(s_{\mathrm{SF}}/s_{\mathrm{SF,ref}})\rho_\beta(z_{\mathrm{struct}})
\end{aligned}
\]
with $s_{\mathrm{SF,ref}}=2.478597640991211$, and
\[
L_{\mathrm{energy}}=\frac14\sum_{k=0}^{3}\frac{s_k}{0.00414681155}\rho_\beta(z_k),\qquad
z_{\mathrm{energy}}=\frac14\sum_{k=0}^{3}z_k.
\]
These component scalars are distinct from their nonnegative normalized severities $z_c$ used for amplitude weighting.

For cotangent construction, an independent float32 output leaf $h=\sg(h_{\mathrm{meas}})$ is made differentiable. All prediction-dependent projections and diagnostic scalars above, together with $\ell_{\mathrm{LR}}$, are reevaluated in float32 using $h$ in place of $h_{\mathrm{meas}}$. Define
\[
v_{\mathrm{LR}}=\nabla_h\!\left(m_{\mathrm{LR}}\ell_{\mathrm{LR}}(h)\right),\qquad
v_c=\nabla_h L_c(h),\quad c\in\{\mathrm{geo,copy,struct,energy}\}.
\]
All norms below are Euclidean norms over the output entries. For an active component's positive normalized severity $z_c$, the implementation clips the amplitude at $z_{\max}=4$ and forms
\begin{equation}
w=\sum_{c\ \mathrm{active}}\min(z_c,4)\frac{v_c}{\|v_c\|+\epsilon}.
\label{eq:si-cotangent-sum}
\end{equation}
With $\lambda_{\mathrm{safe}}=1$ and $\rho_h=0.1$,
\begin{equation}
u=\lambda_{\mathrm{safe}}\rho_h\|v_{\mathrm{LR}}\|
\min\left(1,\frac{\|w\|}{4}\right)\frac{w}{\|w\|+\epsilon},
\qquad
\|u\|\leq\lambda_{\mathrm{safe}}\rho_h\|v_{\mathrm{LR}}\|.
\label{eq:si-cotangent-cap}
\end{equation}
No valid LR anchor or no active component gives zero correction. A stop-gradient carrier adds $\langle h_{\mathrm{meas}},\sg(u)\rangle$ to the weighted LR scalar's backward route, giving the LR-path parameter derivative
\[
\begin{aligned}
g_\theta^{\mathrm{LR}}&=\nabla_\theta\!\left[m_{\mathrm{LR}}\ell_{\mathrm{LR}}(h_{\mathrm{meas}})+\langle h_{\mathrm{meas}},\sg(u)\rangle\right]\\
&=m_{\mathrm{LR}}\nabla_\theta\ell_{\mathrm{LR}}(h_{\mathrm{meas}})
  +\left(\frac{\partial h_{\mathrm{meas}}}{\partial\theta}\right)^\top u.
\end{aligned}
\]
The correction already includes the micro-step LR scaling through $v_{\mathrm{LR}}$ and is not multiplied by another accumulation or safety factor. The cap applies at the output leaf and does not bound the parameter derivative after the Jacobian. The correction enters a single SR-decoder backward pass. The scalar \texttt{L\_safe} is a value-only diagnostic on this cotangent path, not an additional optimized safety-loss term. The frozen-output diagnostics give a maximum measured norm ratio of 0.099999, within the 0.1 cap. This bounded intervention does not calculate an exact observation null space. The PIN--A0 comparison and frozen-output diagnostics characterize its empirical scope in Sec.~\ref{sec:diagnostics}.

The fixed validation panel contains eight crops from the held-out Femur\_74 volume and is selected independently of PIN and the test set. Full PIN terminates training if the fixed validation panel's median $\widehat\alpha$ reaches or exceeds 0.2792616636 in two consecutive evaluations, or its 90th percentile reaches or exceeds 0.4476819009 in two consecutive evaluations. The same fixed thresholds therefore enter both crop-level correction scaling and panel-level termination, with different aggregation and persistence logic. Training-crop scores are monitored but do not authorize termination on their own. For A0--A3, correction and termination are both disabled; copy alarms remain available for monitoring.

\section{Evaluation definitions, regions, and support}
\label{sec:metrics}

\subsection{Shared intensity and segmentation reference}
Intensity values are normalized from uint16 by division by 65,535. The morphology evaluator derives one reference from the HR ROI cores and applies it to both HR and SR. It uses the robust $P_{0.5}$--$P_{99.5}$ intensity range and 4,096-bin Otsu thresholding. The shared HR-derived threshold prevents each reconstruction from selecting its own segmentation criterion. Threshold sensitivity uses the offsets in Sec.~\ref{sec:thresholds}.

Geometric and length-valued morphology measurements use isotropic unit spacing on each specimen's native HR grid. Distances, thickness errors, and tolerance radii are therefore reported in HR-voxel units (HR vox). Native physical voxel pitch varies across VoDaSuRe specimens, so a fixed HR-grid tolerance corresponds to different physical distances across specimens.

Each test object supplies four nonoverlapping primary ROI cores of side 192 HR voxels, with field-of-view margin 32. Thickness analysis uses side-128 ROIs and margin 64. Surface calculations add an initial halo of 32 HR voxels per axis, with extent at least twice the unit grid spacing. The evaluator expands truncated halos up to the 64-voxel cap and filters the resulting surface elements back to the core. This preserves contextual distances without counting the halo as an extra evaluation region. The largest requested distance tolerance must fit within the halo.

\subsection{Image measurements}
PSNR-2D, SSIM-2D, and LPIPS are measured on HR-valid axial slices of the normalized arrays, with border shave zero. A slice is valid when its HR intensity range is at least $10^{-10}$. Valid slices are scored separately and equally averaged within each volume; no valid slice gives an unavailable value. The global summary then equally averages finite per-volume scores, so a volume with more valid slices does not receive more weight. LPIPS requires in-plane dimensions of at least 32 and is evaluated on CPU. The full-volume summaries and ROI-core image summaries remain separate. Image similarity describes agreement with the acquired registered HR realization, rather than a direct surface or anatomical measurement.

\subsection{Area-weighted surface distances and coverage}
Let $d_i^H$ be the nearest-SR distance from HR surface element $i$ of area $a_i^H$, and $d_j^S$ the nearest-HR distance from SR element $j$ of area $a_j^S$. Distances are reported in \HRvox{}. The directional averages are
\begin{equation}
\mathrm{ASD}_{H\to S}=\frac{\sum_i a_i^H d_i^H}{\sum_i a_i^H},\qquad
\mathrm{ASD}_{S\to H}=\frac{\sum_j a_j^S d_j^S}{\sum_j a_j^S}.
\label{eq:si-directional-asd}
\end{equation}
ASSD pools the two sets by their areas:
\begin{equation}
\mathrm{ASSD}=\frac{\sum_i a_i^H d_i^H+\sum_j a_j^S d_j^S}{\sum_i a_i^H+\sum_j a_j^S}.
\label{eq:si-assd}
\end{equation}
It is not the unweighted mean of the directional ASDs. HD95 is the larger of the two directional area-weighted 95th percentiles, following the evaluator's robust Hausdorff implementation:
\begin{equation}
\mathrm{HD}_{95}=\max\{Q_{0.95}^{a_H}(d^H),Q_{0.95}^{a_S}(d^S)\}.
\label{eq:si-hd95}
\end{equation}
The percentile is calculated within each direction rather than after pooling the distances.

At HR-grid tolerance $\tau$, HR-covered and SR-supported report the fraction of surface area within that nearest-surface distance:
\begin{equation}
C_H(\tau)=\frac{\sum_i a_i^H\mathbf{1}[d_i^H\leq\tau]}{\sum_i a_i^H},\qquad
C_S(\tau)=\frac{\sum_j a_j^S\mathbf{1}[d_j^S\leq\tau]}{\sum_j a_j^S}.
\label{eq:si-directional-coverage}
\end{equation}
Surface Dice pools the covered areas over both total areas:
\begin{equation}
\mathrm{SurfaceDice}(\tau)=\frac{\sum_i a_i^H\mathbf{1}[d_i^H\leq\tau]+\sum_j a_j^S\mathbf{1}[d_j^S\leq\tau]}{\sum_i a_i^H+\sum_j a_j^S}.
\label{eq:si-surface-dice}
\end{equation}
HR-covered asks whether reference interfaces have nearby reconstructed counterparts. SR-supported asks whether reconstructed interfaces have nearby reference support. Their directions are retained at 0.5, 1, and 2~\HRvox{} rather than combined into a single completeness claim.

An empty HR core makes the surface quantities unavailable. With nonempty HR and empty SR, HR-covered and Surface Dice are zero; SR-supported and distance measures are unavailable. These rules are why finite distance support can be smaller than coverage support. The evaluator's default base tolerance is the unit grid spacing, increased if an explicit epsilon is larger; the tables state their reported HR-grid tolerances.

\subsection{Voxel and reference-relative material measurements}
\label{sec:material-metrics}
Dice, Jaccard, Precision, and Recall use the shared HR-derived solid masks on the ROI cores. For HR solid $H$, SR solid $S$, and intersection $I=H\cap S$, their positive-denominator definitions are $2|I|/(|H|+|S|)$, $|I|/|H\cup S|$, $|I|/|S|$, and $|I|/|H|$, respectively. Precision concerns reconstructed solid overlap and Recall concerns recovered reference solid.

Unsupported and Missing use HR-grid Euclidean distance transforms calculated with halo context and counted only on the core. Let $H_c=H$ and $S_c=S$ be the core solids defined above, and let $H_h$ and $S_h$ be the solid sets in the halo-expanded regions containing the corresponding cores:
\begin{equation}
U(\tau)=100\frac{\#\{s\in S_c:\operatorname{dist}(s,H_h)>\tau\}}{|S_c|},\qquad
M(\tau)=100\frac{\#\{h\in H_c:\operatorname{dist}(h,S_h)>\tau\}}{|H_c|}.
\label{eq:si-material-rates}
\end{equation}
An empty HR core is excluded by reference quality control, making both material rates unavailable. The primary evaluation additionally excludes fully solid HR cores. For an eligible HR core, an empty $S_c$ makes Unsupported unavailable, while Missing still uses distances to $S_h$; if $S_h$ is empty, Missing is 100\%. The tables retain 0.5, 1, and 2~\HRvox{} tolerances; the main columns use 1~\HRvox{}. These voxel-volume rates remain separate from the surface-area quantities and from each other, because reconstructed material selectivity and reference material completeness have different denominators and meanings.

For HR and SR solid masks \(H\) and \(S\) over the same ROI core of \(N\) voxels, the absolute solid-volume-fraction error is \(e_{\mathrm{VF}}=\bigl||S|/N-|H|/N\bigr|\). The absolute error is computed for each ROI before averaging within evaluation units and then equally across units (Sec.~\ref{sec:nested-aggregation}). For the same binary segmentation, this quantity also equals the absolute porosity-fraction error, since porosity is one minus solid-volume fraction. VF error has no distance tolerance.

\subsection{Nested aggregation and finite support}
\label{sec:nested-aggregation}
For ROI-based structural quantities, the summary first averages the finite ROI values within an object and then averages the finite object means with equal object weight. Reported object and ROI counts accompany each metric. Distance values are not assigned to failed or empty-surface cases. Method-level finite summaries therefore need not have identical ROI support, and a paired comparison uses only ROIs finite in both compared outputs before producing its common-support summaries.

The protocol has ten test evaluation units, each with nested ROIs. Repeated seeds revisit the same units and do not create new specimens. The across-seed sample SD in Sec.~\ref{sec:seeds} describes variability among three trained endpoints. The tables explicitly report unavailable entries, finite support, and quality-control eligibility; missing values are not imputed.

\section{Complete external comparison}
\label{sec:external}

The complete comparison contains 15 method families trained under two source relations, synthetic-only and registered physical-pair, plus PIN. The following tables report all 30 external configurations and PIN's seed-0 results on the ten test units. For external baselines, the Real column denotes real-pair (\realyes{}) or synthetic-only (\realno{}) training; a dash marks PIN and the named supervision or component controls. Main Table~1 displays the eight configurations per relation with the lowest method-level ASSD; the full inventory is reported here.

The seven two-dimensional INR families CiaoSR, HIIF, LIIF, LINF, LTE, MetaSR, and SRNO are each trained separately under synthetic-only and real-pair supervision, giving 14 formal external configurations, each with a 2D EDSR encoder. Their shared settings are fixed $4\times$ scale, configured seed 0, an L1 objective, batch size 32, and 100 epochs with 520 updates per epoch (52,000 updates in total). The learning rate starts at $4\times10^{-4}$ and is halved after epochs 20, 40, 60, and 80, giving $2.5\times10^{-5}$ for the final 20 epochs. Optimization uses first- and second-moment coefficients $(\beta_1,\beta_2)=(0.9,0.999)$, $\epsilon=10^{-8}$, and zero weight decay. Each run is evaluated using its checkpoint with the highest validation PSNR. For the 14 independently trained 2D INR configurations, each LR axial slice is super-resolved in plane, and nearest-neighbor axial upsampling completes the HR-depth volume before the common volumetric evaluator. The three direct-$4\times$ retraining controls below use their separate protocols.

SDice, HRcov, and SRsup abbreviate Surface Dice, HR-covered, and SR-supported. Numeric suffixes 0.5, 1, and 2 denote HR-grid tolerances in \HRvox{}. PSNR is in dB; distances are in \HRvox{}; Unsupported and Missing are percentages; overlap and surface-coverage values are fractions. PSNR, SSIM, overlap, and coverage are higher-is-better; LPIPS, distances, and material-error rates are lower-is-better. Unless a cell carries a smaller support suffix, non-distance ROI summaries have finite values on all ten units and 40 ROIs. Distance support is stated separately. All means follow Sec.~\ref{sec:metrics}.

\subsection{Directional geometry and paired common support}
\label{sec:common-support}
Surface recovery has directional meaning. HR-to-SR ASD and HR-covered describe how well the SR surface reaches reference interfaces; SR-to-HR ASD and SR-supported describe how much reconstructed surface has nearby reference support. PIN's stronger HR coverage and symmetric surface scores do not imply improvement in every reverse-direction measure. On common finite support with real-trained SuperFormer, SR-to-HR ASD is 0.542~\HRvox{} for PIN and 0.364~\HRvox{} for the baseline (ten units, 33 ROIs). On common support with synthetic-trained mDCSRN, ASSD is 0.968 versus 1.724~\HRvox{}, HD95 is 3.302 versus 5.855~\HRvox{}, and HR-to-SR ASD is 1.194 versus 2.105~\HRvox{} (ten units, 38 ROIs).

\begingroup\fontsize{9}{10.8}\selectfont\setlength{\tabcolsep}{4pt}\renewcommand{\arraystretch}{1.12}\renewcommand{\SITableCaption}{\textbf{External distance comparisons on paired common support.} Each row has its own shared finite object/ROI set. Each distance cell gives PIN / baseline, in \HRvox{}. Directions and support are preserved; these values are not differences between independently supported method means.}\begin{longtable}{@{}>{\raggedright\arraybackslash}p{0.850in}>{\centering\arraybackslash}p{0.259in}>{\centering\arraybackslash}p{0.650in}>{\centering\arraybackslash}p{1.060in}>{\centering\arraybackslash}p{1.060in}>{\centering\arraybackslash}p{1.060in}>{\centering\arraybackslash}p{1.060in}@{}}
\caption{\SITableCaption}\label{tab:si-compact-external-paired-distances}\\
\toprule
Baseline & Real & \shortstack{Common\\obj./ROI} & \shortstack{ASD$\downarrow$ HR$\to$SR\\PIN / baseline} & \shortstack{ASD$\downarrow$ SR$\to$HR\\PIN / baseline} & \shortstack{ASSD$\downarrow$\\PIN / baseline} & \shortstack{HD95$\downarrow$\\PIN / baseline} \\
\midrule
\endfirsthead
\multicolumn{7}{l}{\textit{Table \thetable{} continued}}\\
\toprule
Baseline & Real & \shortstack{Common\\obj./ROI} & \shortstack{ASD$\downarrow$ HR$\to$SR\\PIN / baseline} & \shortstack{ASD$\downarrow$ SR$\to$HR\\PIN / baseline} & \shortstack{ASSD$\downarrow$\\PIN / baseline} & \shortstack{HD95$\downarrow$\\PIN / baseline} \\
\midrule
\endhead
\midrule\multicolumn{7}{r}{Continued on next page}\\
\endfoot
\bottomrule
\endlastfoot
MetaSR & \realyes{} & 10/32 & 1.187 / 5.632 & 0.544 / 0.428 & 0.965 / 5.114 & 3.280 / 13.721 \\
MetaSR & \realno{} & 10/33 & 1.185 / 4.642 & 0.542 / 0.544 & 0.963 / 4.084 & 3.276 / 12.042 \\
LIIF & \realyes{} & 10/31 & 1.164 / 5.856 & 0.545 / 0.400 & 0.943 / 5.288 & 3.248 / 14.297 \\
LIIF & \realno{} & 10/33 & 1.185 / 4.654 & 0.542 / 0.545 & 0.963 / 4.095 & 3.276 / 12.100 \\
LTE & \realyes{} & 10/30 & 1.167 / 5.697 & 0.547 / 0.373 & 0.946 / 5.107 & 3.255 / 14.102 \\
LTE & \realno{} & 10/33 & 1.185 / 4.572 & 0.542 / 0.543 & 0.963 / 4.019 & 3.276 / 11.873 \\
SRNO & \realyes{} & 10/27 & 1.160 / 7.307 & 0.549 / 0.387 & 0.941 / 6.654 & 3.239 / 17.319 \\
SRNO & \realno{} & 10/33 & 1.185 / 4.583 & 0.542 / 0.545 & 0.963 / 4.029 & 3.276 / 11.894 \\
CiaoSR & \realyes{} & 10/30 & 1.167 / 5.829 & 0.547 / 0.379 & 0.946 / 5.234 & 3.255 / 14.326 \\
CiaoSR & \realno{} & 10/33 & 1.185 / 4.425 & 0.542 / 0.544 & 0.963 / 3.878 & 3.276 / 11.475 \\
HIIF & \realyes{} & 10/31 & 1.189 / 5.557 & 0.546 / 0.425 & 0.967 / 4.990 & 3.287 / 13.765 \\
HIIF & \realno{} & 10/33 & 1.185 / 5.562 & 0.542 / 0.554 & 0.963 / 4.958 & 3.276 / 13.969 \\
LINF & \realyes{} & 10/27 & 1.162 / 7.112 & 0.549 / 0.481 & 0.942 / 6.402 & 3.239 / 17.932 \\
LINF & \realno{} & 10/33 & 1.185 / 4.362 & 0.542 / 0.538 & 0.963 / 3.819 & 3.276 / 11.320 \\
RCAN & \realyes{} & 10/31 & 1.189 / 6.961 & 0.546 / 0.544 & 0.967 / 6.290 & 3.287 / 16.993 \\
RCAN & \realno{} & 10/38 & 1.194 / 2.884 & 0.551 / 1.604 & 0.968 / 2.759 & 3.302 / 10.984 \\
HAT & \realyes{} & 10/31 & 1.186 / 4.694 & 0.545 / 0.613 & 0.964 / 4.194 & 3.275 / 11.688 \\
HAT & \realno{} & 10/38 & 1.194 / 3.341 & 0.551 / 0.529 & 0.968 / 2.853 & 3.302 / 8.936 \\
EDDSR & \realyes{} & 10/28 & 1.161 / 6.264 & 0.552 / 0.488 & 0.943 / 5.633 & 3.244 / 15.995 \\
EDDSR & \realno{} & 10/38 & 1.194 / 2.588 & 0.551 / 0.497 & 0.968 / 2.175 & 3.302 / 6.889 \\
MFER & \realyes{} & 10/28 & 1.161 / 5.542 & 0.552 / 0.397 & 0.943 / 4.915 & 3.244 / 14.357 \\
MFER & \realno{} & 10/37 & 1.193 / 2.513 & 0.552 / 0.547 & 0.968 / 2.105 & 3.303 / 6.742 \\
mDCSRN & \realyes{} & 10/29 & 1.162 / 5.269 & 0.548 / 0.401 & 0.943 / 4.694 & 3.239 / 13.651 \\
mDCSRN & \realno{} & 10/38 & 1.194 / 2.105 & 0.551 / 0.485 & 0.968 / 1.724 & 3.302 / 5.855 \\
RRDBNet3D & \realyes{} & 10/32 & 1.163 / 4.557 & 0.544 / 0.380 & 0.942 / 3.998 & 3.253 / 12.094 \\
RRDBNet3D & \realno{} & 10/38 & 1.194 / 2.376 & 0.551 / 0.477 & 0.968 / 1.968 & 3.302 / 6.538 \\
SuperFormer & \realyes{} & 10/33 & 1.185 / 4.970 & 0.542 / 0.364 & 0.963 / 4.467 & 3.276 / 12.076 \\
SuperFormer & \realno{} & 10/38 & 1.194 / 2.623 & 0.551 / 0.476 & 0.968 / 2.195 & 3.302 / 7.119 \\
MTVNet & \realyes{} & 10/30 & 1.190 / 5.744 & 0.547 / 0.347 & 0.968 / 5.186 & 3.289 / 14.005 \\
MTVNet & \realno{} & 10/38 & 1.194 / 2.230 & 0.551 / 0.497 & 0.968 / 1.839 & 3.302 / 6.183 \\
\end{longtable}
\endgroup

\begingroup\fontsize{9}{10.8}\selectfont\setlength{\tabcolsep}{4pt}\renewcommand{\arraystretch}{1.12}\renewcommand{\SITableCaption}{\textbf{Complete external surface-distance results.} Distances are in \HRvox{} and use each configuration's own finite support. The object/ROI counts show complete-case evaluability and differ from paired support above.}\begin{longtable}{@{}>{\raggedright\arraybackslash}p{0.850in}>{\centering\arraybackslash}p{0.259in}>{\centering\arraybackslash}p{0.813in}>{\centering\arraybackslash}p{0.813in}>{\centering\arraybackslash}p{0.813in}>{\centering\arraybackslash}p{0.813in}>{\centering\arraybackslash}p{0.813in}>{\centering\arraybackslash}p{0.813in}@{}}
\caption{\SITableCaption}\label{tab:si-compact-external-distances}\\
\toprule
Configuration & Real & \shortstack{ASD$\downarrow$\\HR$\to$SR} & \shortstack{ASD$\downarrow$\\SR$\to$HR} & ASSD$\downarrow$ & HD95$\downarrow$ & \shortstack{Finite\\objects} & \shortstack{Finite\\ROIs} \\
\midrule
\endfirsthead
\multicolumn{8}{l}{\textit{Table \thetable{} continued}}\\
\toprule
Configuration & Real & \shortstack{ASD$\downarrow$\\HR$\to$SR} & \shortstack{ASD$\downarrow$\\SR$\to$HR} & ASSD$\downarrow$ & HD95$\downarrow$ & \shortstack{Finite\\objects} & \shortstack{Finite\\ROIs} \\
\midrule
\endhead
\midrule\multicolumn{8}{r}{Continued on next page}\\
\endfoot
\bottomrule
\endlastfoot
MetaSR & \realyes{} & \shortstack{5.632\\(32/40 ROI)} & \shortstack{0.428\\(32/40 ROI)} & \shortstack{5.114\\(32/40 ROI)} & \shortstack{13.721\\(32/40 ROI)} & 10/10 & 32/40 \\
MetaSR & \realno{} & \shortstack{4.642\\(33/40 ROI)} & \shortstack{0.544\\(33/40 ROI)} & \shortstack{4.084\\(33/40 ROI)} & \shortstack{12.042\\(33/40 ROI)} & 10/10 & 33/40 \\
LIIF & \realyes{} & \shortstack{5.856\\(31/40 ROI)} & \shortstack{0.400\\(31/40 ROI)} & \shortstack{5.288\\(31/40 ROI)} & \shortstack{14.297\\(31/40 ROI)} & 10/10 & 31/40 \\
LIIF & \realno{} & \shortstack{4.654\\(33/40 ROI)} & \shortstack{0.545\\(33/40 ROI)} & \shortstack{4.095\\(33/40 ROI)} & \shortstack{12.100\\(33/40 ROI)} & 10/10 & 33/40 \\
LTE & \realyes{} & 5.697 & 0.373 & 5.107 & 14.102 & 10/10 & 30/40 \\
LTE & \realno{} & \shortstack{4.572\\(33/40 ROI)} & \shortstack{0.543\\(33/40 ROI)} & \shortstack{4.019\\(33/40 ROI)} & \shortstack{11.873\\(33/40 ROI)} & 10/10 & 33/40 \\
SRNO & \realyes{} & \shortstack{7.307\\(27/40 ROI)} & \shortstack{0.387\\(27/40 ROI)} & \shortstack{6.654\\(27/40 ROI)} & \shortstack{17.319\\(27/40 ROI)} & 10/10 & 27/40 \\
SRNO & \realno{} & \shortstack{4.583\\(33/40 ROI)} & \shortstack{0.545\\(33/40 ROI)} & \shortstack{4.029\\(33/40 ROI)} & \shortstack{11.894\\(33/40 ROI)} & 10/10 & 33/40 \\
CiaoSR & \realyes{} & 5.829 & 0.379 & 5.234 & 14.326 & 10/10 & 30/40 \\
CiaoSR & \realno{} & 4.425 & 0.544 & 3.878 & 11.475 & 10/10 & 33/40 \\
HIIF & \realyes{} & \shortstack{5.557\\(31/40 ROI)} & \shortstack{0.425\\(31/40 ROI)} & \shortstack{4.990\\(31/40 ROI)} & \shortstack{13.765\\(31/40 ROI)} & 10/10 & 31/40 \\
HIIF & \realno{} & \shortstack{5.562\\(33/40 ROI)} & \shortstack{0.554\\(33/40 ROI)} & \shortstack{4.958\\(33/40 ROI)} & \shortstack{13.969\\(33/40 ROI)} & 10/10 & 33/40 \\
LINF & \realyes{} & \shortstack{7.112\\(27/40 ROI)} & \shortstack{0.481\\(27/40 ROI)} & \shortstack{6.402\\(27/40 ROI)} & \shortstack{17.932\\(27/40 ROI)} & 10/10 & 27/40 \\
LINF & \realno{} & \shortstack{4.362\\(33/40 ROI)} & \shortstack{0.538\\(33/40 ROI)} & \shortstack{3.819\\(33/40 ROI)} & \shortstack{11.320\\(33/40 ROI)} & 10/10 & 33/40 \\
RCAN & \realyes{} & 6.961 & 0.544 & 6.290 & 16.993 & 10/10 & 31/40 \\
RCAN & \realno{} & 2.884 & 1.604 & 2.759 & 10.984 & 10/10 & 38/40 \\
HAT & \realyes{} & 4.694 & 0.613 & 4.194 & 11.688 & 10/10 & 31/40 \\
HAT & \realno{} & 3.341 & 0.529 & 2.853 & 8.936 & 10/10 & 38/40 \\
EDDSR & \realyes{} & 6.264 & 0.488 & 5.633 & 15.995 & 10/10 & 28/40 \\
EDDSR & \realno{} & 2.588 & 0.497 & 2.175 & 6.889 & 10/10 & 38/40 \\
MFER & \realyes{} & 5.542 & 0.397 & 4.915 & 14.357 & 10/10 & 28/40 \\
MFER & \realno{} & 2.513 & 0.547 & 2.105 & 6.742 & 10/10 & 37/40 \\
mDCSRN & \realyes{} & 5.269 & 0.401 & 4.694 & 13.651 & 10/10 & 29/40 \\
mDCSRN & \realno{} & 2.105 & 0.485 & 1.724 & 5.855 & 10/10 & 38/40 \\
RRDBNet3D & \realyes{} & 4.557 & 0.380 & 3.998 & 12.094 & 10/10 & 32/40 \\
RRDBNet3D & \realno{} & 2.376 & 0.477 & 1.968 & 6.538 & 10/10 & 38/40 \\
SuperFormer & \realyes{} & 4.970 & 0.364 & 4.467 & 12.076 & 10/10 & 33/40 \\
SuperFormer & \realno{} & 2.623 & 0.476 & 2.195 & 7.119 & 10/10 & 38/40 \\
MTVNet & \realyes{} & \shortstack{5.744\\(30/40 ROI)} & \shortstack{0.347\\(30/40 ROI)} & \shortstack{5.186\\(30/40 ROI)} & \shortstack{14.005\\(30/40 ROI)} & 10/10 & 30/40 \\
MTVNet & \realno{} & \shortstack{2.230\\(38/40 ROI)} & \shortstack{0.497\\(38/40 ROI)} & \shortstack{1.839\\(38/40 ROI)} & \shortstack{6.183\\(38/40 ROI)} & 10/10 & 38/40 \\
PIN & \textendash{} & 1.231 & 0.556 & 0.995 & 3.473 & 10/10 & 40/40 \\
\end{longtable}
\endgroup

\begingroup\fontsize{9}{10.8}\selectfont\setlength{\tabcolsep}{4pt}\renewcommand{\arraystretch}{1.12}\renewcommand{\SITableCaption}{\textbf{External surface-distance QC.} Counts report boundary, halo, empty-SR, and other-invalid categories; flags can overlap. Censored distances remain unavailable and do not receive artificial penalties.}\begin{longtable}{@{}>{\raggedright\arraybackslash}p{0.850in}>{\centering\arraybackslash}p{0.259in}>{\centering\arraybackslash}p{0.976in}>{\centering\arraybackslash}p{0.976in}>{\centering\arraybackslash}p{0.976in}>{\centering\arraybackslash}p{0.976in}>{\centering\arraybackslash}p{0.976in}@{}}
\caption{\SITableCaption}\label{tab:si-compact-external-distance-qc}\\
\toprule
Configuration & Real & \shortstack{Finite\\ROIs} & \shortstack{Boundary\\QC} & \shortstack{Insufficient\\halo} & \shortstack{Empty\\SR} & \shortstack{Other\\invalid} \\
\midrule
\endfirsthead
\multicolumn{7}{l}{\textit{Table \thetable{} continued}}\\
\toprule
Configuration & Real & \shortstack{Finite\\ROIs} & \shortstack{Boundary\\QC} & \shortstack{Insufficient\\halo} & \shortstack{Empty\\SR} & \shortstack{Other\\invalid} \\
\midrule
\endhead
\midrule\multicolumn{7}{r}{Continued on next page}\\
\endfoot
\bottomrule
\endlastfoot
MetaSR & \realyes{} & 32/40 & 8 & 3 & 0 & 0 \\
MetaSR & \realno{} & 33/40 & 7 & 0 & 0 & 0 \\
LIIF & \realyes{} & 31/40 & 9 & 4 & 0 & 0 \\
LIIF & \realno{} & 33/40 & 7 & 0 & 0 & 0 \\
LTE & \realyes{} & 30/40 & 10 & 4 & 0 & 0 \\
LTE & \realno{} & 33/40 & 7 & 0 & 0 & 0 \\
SRNO & \realyes{} & 27/40 & 13 & 5 & 0 & 0 \\
SRNO & \realno{} & 33/40 & 7 & 0 & 0 & 0 \\
CiaoSR & \realyes{} & 30/40 & 10 & 3 & 0 & 0 \\
CiaoSR & \realno{} & 33/40 & 7 & 0 & 0 & 0 \\
HIIF & \realyes{} & 31/40 & 9 & 4 & 0 & 0 \\
HIIF & \realno{} & 33/40 & 7 & 1 & 0 & 0 \\
LINF & \realyes{} & 27/40 & 13 & 7 & 0 & 0 \\
LINF & \realno{} & 33/40 & 7 & 0 & 0 & 0 \\
RCAN & \realyes{} & 31/40 & 9 & 3 & 0 & 0 \\
RCAN & \realno{} & 38/40 & 2 & 0 & 0 & 0 \\
HAT & \realyes{} & 31/40 & 9 & 2 & 0 & 0 \\
HAT & \realno{} & 38/40 & 2 & 0 & 0 & 0 \\
EDDSR & \realyes{} & 28/40 & 12 & 5 & 0 & 0 \\
EDDSR & \realno{} & 38/40 & 2 & 0 & 0 & 0 \\
MFER & \realyes{} & 28/40 & 12 & 4 & 0 & 0 \\
MFER & \realno{} & 37/40 & 2 & 1 & 0 & 0 \\
mDCSRN & \realyes{} & 29/40 & 11 & 6 & 0 & 0 \\
mDCSRN & \realno{} & 38/40 & 2 & 0 & 0 & 0 \\
RRDBNet3D & \realyes{} & 32/40 & 8 & 3 & 0 & 0 \\
RRDBNet3D & \realno{} & 38/40 & 2 & 0 & 0 & 0 \\
SuperFormer & \realyes{} & 33/40 & 7 & 2 & 0 & 0 \\
SuperFormer & \realno{} & 38/40 & 2 & 0 & 0 & 0 \\
MTVNet & \realyes{} & 30/40 & 10 & 1 & 0 & 0 \\
MTVNet & \realno{} & 38/40 & 2 & 0 & 0 & 0 \\
PIN & \textendash{} & 40/40 & 0 & 0 & 0 & 0 \\
\end{longtable}
\endgroup

\subsection{Image, interface, and material results}
\label{sec:external-results-detail}
Image similarity, interface recovery, and material measurements characterize complementary aspects of the external benchmark. PIN achieves the highest voxel Precision and lowest Unsupported solid volume at 1~\HRvox{} across all 30 external configurations, alongside the strongest Surface Dice and HR-covered at the reported tolerances. The complete voxel-overlap and distance-tolerant material results remain in Tables~\ref{tab:si-compact-external-images}--\ref{tab:si-compact-external-material}. Solid Recall and Missing quantify reference-solid coverage without penalizing the filling of reference pores. The higher Recall and lower Missing at 1~\HRvox{} of the external configurations (Tables~\ref{tab:si-compact-external-voxel-reverse-coverage}--\ref{tab:si-compact-external-material}) coexist with fully erased reference-pore volume shares of 37.33--72.92\%, compared with 10.41\% for PIN (Sec.~\ref{sec:material-topology}, Table~\ref{tab:si-pore-topology}). These coverage measures should therefore be interpreted jointly with interface recovery and pore preservation.

\begingroup\fontsize{9}{10.8}\selectfont\setlength{\tabcolsep}{4pt}\renewcommand{\arraystretch}{1.12}\renewcommand{\SITableCaption}{\textbf{Complete external image results.} Volume columns use the full-volume slice protocol; ROI columns use the ROI cores. PSNR is in dB; SSIM and LPIPS follow Sec.~\ref{sec:metrics}. All 31 configurations have ten finite evaluation-unit summaries.}\begin{longtable}{@{}>{\raggedright\arraybackslash}p{0.850in}>{\centering\arraybackslash}p{0.259in}>{\centering\arraybackslash}p{0.813in}>{\centering\arraybackslash}p{0.813in}>{\centering\arraybackslash}p{0.813in}>{\centering\arraybackslash}p{0.813in}>{\centering\arraybackslash}p{0.813in}>{\centering\arraybackslash}p{0.813in}@{}}
\caption{\SITableCaption}\label{tab:si-compact-external-images}\\
\toprule
Configuration & Real & \shortstack{Volume\\PSNR$\uparrow$} & \shortstack{Volume\\SSIM$\uparrow$} & \shortstack{Volume\\LPIPS$\downarrow$} & \shortstack{ROI\\PSNR$\uparrow$} & \shortstack{ROI\\SSIM$\uparrow$} & \shortstack{ROI\\LPIPS$\downarrow$} \\
\midrule
\endfirsthead
\multicolumn{8}{l}{\textit{Table \thetable{} continued}}\\
\toprule
Configuration & Real & \shortstack{Volume\\PSNR$\uparrow$} & \shortstack{Volume\\SSIM$\uparrow$} & \shortstack{Volume\\LPIPS$\downarrow$} & \shortstack{ROI\\PSNR$\uparrow$} & \shortstack{ROI\\SSIM$\uparrow$} & \shortstack{ROI\\LPIPS$\downarrow$} \\
\midrule
\endhead
\midrule\multicolumn{8}{r}{Continued on next page}\\
\endfoot
\bottomrule
\endlastfoot
MetaSR & \realyes{} & 16.038 & 0.3976 & 0.4996 & 18.536 & 0.5451 & 0.3829 \\
MetaSR & \realno{} & 15.295 & 0.3383 & 0.5406 & 17.223 & 0.4753 & 0.4206 \\
LIIF & \realyes{} & 16.011 & 0.3849 & 0.4954 & 18.411 & 0.5210 & 0.3813 \\
LIIF & \realno{} & 15.283 & 0.3377 & 0.5432 & 17.165 & 0.4745 & 0.4211 \\
LTE & \realyes{} & 16.028 & 0.3977 & 0.4856 & 18.461 & 0.5407 & 0.3691 \\
LTE & \realno{} & 15.281 & 0.3384 & 0.5455 & 17.165 & 0.4751 & 0.4204 \\
SRNO & \realyes{} & 16.141 & 0.3891 & 0.4767 & 17.911 & 0.4455 & 0.3818 \\
SRNO & \realno{} & 15.282 & 0.3380 & 0.5452 & 17.165 & 0.4747 & 0.4203 \\
CiaoSR & \realyes{} & 16.412 & 0.4275 & 0.4692 & 18.563 & 0.5553 & 0.3580 \\
CiaoSR & \realno{} & 15.272 & 0.3385 & 0.5459 & 17.140 & 0.4751 & 0.4192 \\
HIIF & \realyes{} & 15.933 & 0.3850 & 0.4727 & 18.174 & 0.4946 & 0.3631 \\
HIIF & \realno{} & 15.277 & 0.3251 & 0.5647 & 17.342 & 0.4670 & 0.4489 \\
LINF & \realyes{} & 15.767 & 0.2288 & 0.5000 & 17.872 & 0.2597 & 0.3926 \\
LINF & \realno{} & 15.298 & 0.3355 & 0.5355 & 17.271 & 0.4735 & 0.4121 \\
RCAN & \realyes{} & 15.659 & 0.3200 & 0.5378 & 17.608 & 0.4206 & 0.4288 \\
RCAN & \realno{} & 14.915 & 0.3336 & 0.5082 & 16.386 & 0.4513 & 0.4188 \\
HAT & \realyes{} & 15.686 & 0.3466 & 0.5342 & 17.516 & 0.4578 & 0.4336 \\
HAT & \realno{} & 14.942 & 0.3075 & 0.5331 & 15.729 & 0.3310 & 0.4767 \\
EDDSR & \realyes{} & 15.908 & 0.3782 & 0.5114 & 18.024 & 0.5046 & 0.4043 \\
EDDSR & \realno{} & 15.320 & 0.3702 & 0.4781 & 17.436 & 0.5023 & 0.3739 \\
MFER & \realyes{} & 16.165 & 0.4021 & 0.4793 & 18.514 & 0.5379 & 0.3668 \\
MFER & \realno{} & 14.933 & 0.3504 & 0.4770 & 17.188 & 0.5008 & 0.3777 \\
mDCSRN & \realyes{} & 16.273 & 0.4071 & 0.4789 & 18.581 & 0.5398 & 0.3683 \\
mDCSRN & \realno{} & 15.201 & 0.3840 & 0.4666 & 17.546 & 0.5268 & 0.3547 \\
RRDBNet3D & \realyes{} & 16.353 & 0.4191 & 0.4788 & 18.628 & 0.5484 & 0.3672 \\
RRDBNet3D & \realno{} & 15.230 & 0.3806 & 0.4772 & 17.707 & 0.5393 & 0.3596 \\
SuperFormer & \realyes{} & 16.412 & 0.4229 & 0.4751 & 18.763 & 0.5548 & 0.3636 \\
SuperFormer & \realno{} & 15.348 & 0.3741 & 0.4631 & 17.469 & 0.4661 & 0.3691 \\
MTVNet & \realyes{} & 16.402 & 0.4096 & 0.4830 & 19.020 & 0.5602 & 0.3645 \\
MTVNet & \realno{} & 15.042 & 0.3478 & 0.4890 & 16.552 & 0.4013 & 0.4162 \\
PIN & \textendash{} & 15.090 & 0.3878 & 0.4514 & 17.250 & 0.5221 & 0.3452 \\
\end{longtable}
\endgroup

\begingroup\fontsize{9}{10.8}\selectfont\setlength{\tabcolsep}{4pt}\renewcommand{\arraystretch}{1.12}\renewcommand{\SITableCaption}{\textbf{Complete external reference-interface coverage.} SDice and HRcov are area fractions at 0.5, 1, and 2~\HRvox{}. Both have finite support on ten units and 40 ROIs for every configuration, including defined zero scores for empty predictions.}\begin{longtable}{@{}>{\raggedright\arraybackslash}p{0.850in}>{\centering\arraybackslash}p{0.259in}>{\centering\arraybackslash}p{0.813in}>{\centering\arraybackslash}p{0.813in}>{\centering\arraybackslash}p{0.813in}>{\centering\arraybackslash}p{0.813in}>{\centering\arraybackslash}p{0.813in}>{\centering\arraybackslash}p{0.813in}@{}}
\caption{\SITableCaption}\label{tab:si-compact-external-reference-coverage}\\
\toprule
Configuration & Real & \shortstack{SDice$\uparrow$\\0.5} & \shortstack{SDice$\uparrow$\\1} & \shortstack{SDice$\uparrow$\\2} & \shortstack{HRcov$\uparrow$\\0.5} & \shortstack{HRcov$\uparrow$\\1} & \shortstack{HRcov$\uparrow$\\2} \\
\midrule
\endfirsthead
\multicolumn{8}{l}{\textit{Table \thetable{} continued}}\\
\toprule
Configuration & Real & \shortstack{SDice$\uparrow$\\0.5} & \shortstack{SDice$\uparrow$\\1} & \shortstack{SDice$\uparrow$\\2} & \shortstack{HRcov$\uparrow$\\0.5} & \shortstack{HRcov$\uparrow$\\1} & \shortstack{HRcov$\uparrow$\\2} \\
\midrule
\endhead
\midrule\multicolumn{8}{r}{Continued on next page}\\
\endfoot
\bottomrule
\endlastfoot
MetaSR & \realyes{} & 0.2423 & 0.4213 & 0.5011 & 0.1834 & 0.3486 & 0.4368 \\
MetaSR & \realno{} & 0.1981 & 0.3772 & 0.4912 & 0.1334 & 0.2877 & 0.4067 \\
LIIF & \realyes{} & 0.2324 & 0.3980 & 0.4709 & 0.1751 & 0.3288 & 0.4084 \\
LIIF & \realno{} & 0.1989 & 0.3777 & 0.4938 & 0.1340 & 0.2879 & 0.4090 \\
LTE & \realyes{} & 0.2378 & 0.4006 & 0.4744 & 0.1783 & 0.3293 & 0.4103 \\
LTE & \realno{} & 0.2012 & 0.3815 & 0.4996 & 0.1357 & 0.2912 & 0.4149 \\
SRNO & \realyes{} & 0.2123 & 0.3622 & 0.4307 & 0.1593 & 0.2969 & 0.3711 \\
SRNO & \realno{} & 0.2000 & 0.3804 & 0.4988 & 0.1347 & 0.2902 & 0.4142 \\
CiaoSR & \realyes{} & 0.2360 & 0.3976 & 0.4723 & 0.1775 & 0.3281 & 0.4090 \\
CiaoSR & \realno{} & 0.2026 & 0.3850 & 0.5048 & 0.1365 & 0.2940 & 0.4196 \\
HIIF & \realyes{} & 0.2405 & 0.4249 & 0.5083 & 0.1797 & 0.3502 & 0.4430 \\
HIIF & \realno{} & 0.1727 & 0.3357 & 0.4429 & 0.1143 & 0.2522 & 0.3613 \\
LINF & \realyes{} & 0.1984 & 0.3592 & 0.4402 & 0.1431 & 0.2861 & 0.3723 \\
LINF & \realno{} & 0.2079 & 0.3957 & 0.5121 & 0.1406 & 0.3034 & 0.4265 \\
RCAN & \realyes{} & 0.2097 & 0.3846 & 0.4793 & 0.1537 & 0.3127 & 0.4127 \\
RCAN & \realno{} & 0.2632 & 0.4933 & 0.6281 & 0.1865 & 0.3979 & 0.5493 \\
HAT & \realyes{} & 0.2202 & 0.4053 & 0.5074 & 0.1603 & 0.3280 & 0.4364 \\
HAT & \realno{} & 0.2523 & 0.4750 & 0.6067 & 0.1741 & 0.3744 & 0.5183 \\
EDDSR & \realyes{} & 0.2062 & 0.3595 & 0.4413 & 0.1526 & 0.2838 & 0.3733 \\
EDDSR & \realno{} & 0.2921 & 0.5175 & 0.6500 & 0.2127 & 0.4106 & 0.5620 \\
MFER & \realyes{} & 0.2656 & 0.4277 & 0.4998 & 0.2060 & 0.3500 & 0.4313 \\
MFER & \realno{} & 0.2838 & 0.5109 & 0.6545 & 0.2070 & 0.4046 & 0.5644 \\
mDCSRN & \realyes{} & 0.2611 & 0.4164 & 0.4871 & 0.2039 & 0.3410 & 0.4206 \\
mDCSRN & \realno{} & 0.3247 & 0.5669 & 0.7046 & 0.2408 & 0.4568 & 0.6172 \\
RRDBNet3D & \realyes{} & 0.2913 & 0.4648 & 0.5431 & 0.2252 & 0.3810 & 0.4701 \\
RRDBNet3D & \realno{} & 0.3095 & 0.5393 & 0.6708 & 0.2272 & 0.4308 & 0.5828 \\
SuperFormer & \realyes{} & 0.2793 & 0.4412 & 0.5136 & 0.2171 & 0.3632 & 0.4460 \\
SuperFormer & \realno{} & 0.3014 & 0.5240 & 0.6520 & 0.2199 & 0.4157 & 0.5623 \\
MTVNet & \realyes{} & 0.2629 & 0.4097 & 0.4727 & 0.2046 & 0.3365 & 0.4090 \\
MTVNet & \realno{} & 0.3135 & 0.5537 & 0.6928 & 0.2315 & 0.4453 & 0.6057 \\
PIN & \textendash{} & 0.4021 & 0.7046 & 0.8687 & 0.3282 & 0.6137 & 0.8107 \\
\end{longtable}
\endgroup

\begingroup\fontsize{9}{10.8}\selectfont\setlength{\tabcolsep}{4pt}\renewcommand{\arraystretch}{1.12}\renewcommand{\SITableCaption}{\textbf{Complete external voxel overlap and reconstructed-surface support.} Voxel overlap has no distance tolerance. SRsup is an area fraction at the stated tolerances in \HRvox{}. Cells with reduced support retain their finite-ROI suffix; other cells use ten units and 40 ROIs.}\begin{longtable}{@{}>{\raggedright\arraybackslash}p{0.850in}>{\centering\arraybackslash}p{0.259in}>{\centering\arraybackslash}p{0.680in}>{\centering\arraybackslash}p{0.680in}>{\centering\arraybackslash}p{0.680in}>{\centering\arraybackslash}p{0.680in}>{\centering\arraybackslash}p{0.680in}>{\centering\arraybackslash}p{0.680in}>{\centering\arraybackslash}p{0.680in}@{}}
\caption{\SITableCaption}\label{tab:si-compact-external-voxel-reverse-coverage}\\
\toprule
Configuration & Real & Dice$\uparrow$ & Jaccard$\uparrow$ & Precision$\uparrow$ & Recall$\uparrow$ & \shortstack{SRsup$\uparrow$\\0.5} & \shortstack{SRsup$\uparrow$\\1} & \shortstack{SRsup$\uparrow$\\2} \\
\midrule
\endfirsthead
\multicolumn{9}{l}{\textit{Table \thetable{} continued}}\\
\toprule
Configuration & Real & Dice$\uparrow$ & Jaccard$\uparrow$ & Precision$\uparrow$ & Recall$\uparrow$ & \shortstack{SRsup$\uparrow$\\0.5} & \shortstack{SRsup$\uparrow$\\1} & \shortstack{SRsup$\uparrow$\\2} \\
\midrule
\endhead
\midrule\multicolumn{9}{r}{Continued on next page}\\
\endfoot
\bottomrule
\endlastfoot
MetaSR & \realyes{} & 0.8750 & 0.7792 & 0.8208 & 0.9424 & 0.6145 & 0.9319 & 0.9908 \\
MetaSR & \realno{} & 0.8491 & 0.7403 & 0.8134 & 0.8928 & 0.5526 & 0.8846 & 0.9809 \\
LIIF & \realyes{} & 0.8762 & 0.7811 & 0.8204 & 0.9466 & 0.6415 & 0.9345 & 0.9889 \\
LIIF & \realno{} & 0.8495 & 0.7410 & 0.8131 & 0.8941 & 0.5523 & 0.8832 & 0.9810 \\
LTE & \realyes{} & 0.8775 & 0.7832 & 0.8194 & 0.9506 & 0.6539 & 0.9370 & 0.9909 \\
LTE & \realno{} & 0.8494 & 0.7408 & 0.8133 & 0.8936 & 0.5531 & 0.8839 & 0.9816 \\
SRNO & \realyes{} & 0.8780 & 0.7839 & 0.8125 & 0.9601 & 0.6371 & 0.9359 & 0.9905 \\
SRNO & \realno{} & 0.8496 & 0.7411 & 0.8129 & 0.8946 & 0.5521 & 0.8829 & 0.9809 \\
CiaoSR & \realyes{} & 0.8774 & 0.7829 & 0.8192 & 0.9509 & 0.6537 & 0.9355 & 0.9900 \\
CiaoSR & \realno{} & 0.8492 & 0.7405 & 0.8130 & 0.8934 & 0.5526 & 0.8832 & 0.9812 \\
HIIF & \realyes{} & 0.8708 & 0.7729 & 0.8218 & 0.9322 & 0.6162 & 0.9300 & 0.9911 \\
HIIF & \realno{} & 0.8505 & 0.7423 & 0.8089 & 0.9023 & 0.5446 & 0.8796 & 0.9794 \\
LINF & \realyes{} & 0.8698 & 0.7714 & 0.8180 & 0.9376 & 0.5913 & 0.9161 & 0.9867 \\
LINF & \realno{} & 0.8474 & 0.7380 & 0.8157 & 0.8865 & 0.5544 & 0.8884 & 0.9819 \\
RCAN & \realyes{} & 0.8621 & 0.7598 & 0.8218 & 0.9139 & 0.5654 & 0.8891 & 0.9764 \\
RCAN & \realno{} & 0.8408 & 0.7281 & 0.8163 & 0.8710 & 0.5322 & 0.8441 & 0.9307 \\
HAT & \realyes{} & 0.8581 & 0.7537 & 0.8232 & 0.9034 & 0.5669 & 0.8934 & 0.9810 \\
HAT & \realno{} & 0.8476 & 0.7382 & 0.8215 & 0.8795 & 0.5635 & 0.8921 & 0.9824 \\
EDDSR & \realyes{} & 0.8709 & 0.7731 & 0.8187 & 0.9391 & 0.5957 & 0.9188 & 0.9864 \\
EDDSR & \realno{} & 0.8480 & 0.7391 & 0.8289 & 0.8722 & 0.5786 & 0.9076 & 0.9866 \\
MFER & \realyes{} & 0.8792 & 0.7859 & 0.8245 & 0.9476 & 0.6465 & 0.9392 & 0.9892 \\
MFER & \realno{} & 0.8400 & 0.7272 & 0.8258 & 0.8586 & 0.5541 & 0.8849 & 0.9804 \\
mDCSRN & \realyes{} & 0.8814 & 0.7894 & 0.8248 & 0.9520 & 0.6451 & 0.9432 & 0.9913 \\
mDCSRN & \realno{} & 0.8474 & 0.7382 & 0.8330 & 0.8665 & 0.5860 & 0.9116 & 0.9878 \\
RRDBNet3D & \realyes{} & 0.8809 & 0.7886 & 0.8267 & 0.9479 & 0.6576 & 0.9423 & 0.9912 \\
RRDBNet3D & \realno{} & 0.8515 & 0.7442 & 0.8321 & 0.8760 & 0.5914 & 0.9158 & 0.9887 \\
SuperFormer & \realyes{} & 0.8833 & 0.7925 & 0.8281 & 0.9516 & 0.6662 & 0.9508 & 0.9936 \\
SuperFormer & \realno{} & 0.8510 & 0.7436 & 0.8293 & 0.8783 & 0.5929 & 0.9140 & 0.9882 \\
MTVNet & \realyes{} & 0.8848 & 0.7949 & 0.8274 & 0.9567 & 0.6765 & 0.9572 & 0.9947 \\
MTVNet & \realno{} & 0.8464 & 0.7368 & 0.8329 & 0.8648 & 0.5789 & 0.9072 & 0.9873 \\
PIN & \textendash{} & 0.8109 & 0.6873 & 0.8574 & 0.7736 & 0.5429 & 0.8781 & 0.9796 \\
\end{longtable}
\endgroup

\begingroup\fontsize{9}{10.8}\selectfont\setlength{\tabcolsep}{4pt}\renewcommand{\arraystretch}{1.12}\renewcommand{\SITableCaption}{\textbf{Complete external reference-relative material rates.} Unsupported uses predicted-solid volume as denominator; Missing uses reference-solid volume. Both are percentages at the three HR-grid tolerances in \HRvox{}. Reduced finite support is attached to the affected cells.}\begin{longtable}{@{}>{\raggedright\arraybackslash}p{0.850in}>{\centering\arraybackslash}p{0.259in}>{\centering\arraybackslash}p{0.813in}>{\centering\arraybackslash}p{0.813in}>{\centering\arraybackslash}p{0.813in}>{\centering\arraybackslash}p{0.813in}>{\centering\arraybackslash}p{0.813in}>{\centering\arraybackslash}p{0.813in}@{}}
\caption{\SITableCaption}\label{tab:si-compact-external-material}\\
\toprule
Configuration & Real & \shortstack{Unsupported$\downarrow$\\0.5} & \shortstack{Unsupported$\downarrow$\\1} & \shortstack{Unsupported$\downarrow$\\2} & \shortstack{Missing$\downarrow$\\0.5} & \shortstack{Missing$\downarrow$\\1} & \shortstack{Missing$\downarrow$\\2} \\
\midrule
\endfirsthead
\multicolumn{8}{l}{\textit{Table \thetable{} continued}}\\
\toprule
Configuration & Real & \shortstack{Unsupported$\downarrow$\\0.5} & \shortstack{Unsupported$\downarrow$\\1} & \shortstack{Unsupported$\downarrow$\\2} & \shortstack{Missing$\downarrow$\\0.5} & \shortstack{Missing$\downarrow$\\1} & \shortstack{Missing$\downarrow$\\2} \\
\midrule
\endhead
\midrule\multicolumn{8}{r}{Continued on next page}\\
\endfoot
\bottomrule
\endlastfoot
MetaSR & \realyes{} & 17.92 & 2.68 & 0.12 & 5.76 & 1.69 & 0.61 \\
MetaSR & \realno{} & 18.66 & 3.19 & 0.23 & 10.72 & 6.68 & 3.88 \\
LIIF & \realyes{} & 17.96 & 2.71 & 0.16 & 5.34 & 1.58 & 0.60 \\
LIIF & \realno{} & 18.69 & 3.21 & 0.23 & 10.59 & 6.54 & 3.73 \\
LTE & \realyes{} & 18.06 & 2.73 & 0.12 & 4.94 & 1.43 & 0.56 \\
LTE & \realno{} & 18.67 & 3.21 & 0.23 & 10.64 & 6.55 & 3.69 \\
SRNO & \realyes{} & 18.75 & 3.01 & 0.18 & 3.99 & 1.21 & 0.51 \\
SRNO & \realno{} & 18.71 & 3.23 & 0.23 & 10.54 & 6.47 & 3.63 \\
CiaoSR & \realyes{} & 18.08 & 2.73 & 0.11 & 4.91 & 1.48 & 0.60 \\
CiaoSR & \realno{} & 18.70 & 3.23 & 0.23 & 10.66 & 6.54 & 3.66 \\
HIIF & \realyes{} & 17.82 & 2.65 & 0.09 & 6.78 & 2.49 & 0.98 \\
HIIF & \realno{} & 19.11 & 3.40 & 0.28 & 9.77 & 6.08 & 3.57 \\
LINF & \realyes{} & 18.20 & 2.79 & 0.11 & 6.24 & 2.59 & 1.24 \\
LINF & \realno{} & 18.43 & 3.09 & 0.21 & 11.35 & 7.06 & 4.10 \\
RCAN & \realyes{} & 17.82 & 2.70 & 0.11 & 8.61 & 4.34 & 2.14 \\
RCAN & \realno{} & 18.37 & 3.75 & 1.12 & 12.90 & 7.15 & 3.67 \\
HAT & \realyes{} & 17.68 & 2.68 & 0.13 & 9.66 & 4.99 & 2.55 \\
HAT & \realno{} & 17.85 & 2.89 & 0.16 & 12.05 & 6.64 & 3.31 \\
EDDSR & \realyes{} & 18.13 & 2.80 & 0.13 & 6.09 & 2.68 & 1.26 \\
EDDSR & \realno{} & 17.11 & 2.58 & 0.13 & 12.78 & 7.43 & 3.90 \\
MFER & \realyes{} & 17.55 & 2.49 & 0.08 & 5.24 & 1.59 & 0.56 \\
MFER & \realno{} & 17.42 & 2.84 & 0.22 & 14.14 & 8.32 & 4.40 \\
mDCSRN & \realyes{} & 17.52 & 2.43 & 0.05 & 4.80 & 1.51 & 0.55 \\
mDCSRN & \realno{} & 16.70 & 2.41 & 0.09 & 13.35 & 7.49 & 3.85 \\
RRDBNet3D & \realyes{} & 17.33 & 2.35 & 0.05 & 5.21 & 1.62 & 0.63 \\
RRDBNet3D & \realno{} & 16.79 & 2.40 & 0.08 & 12.40 & 7.02 & 3.59 \\
SuperFormer & \realyes{} & 17.19 & 2.31 & 0.04 & 4.84 & 1.36 & 0.41 \\
SuperFormer & \realno{} & 17.07 & 2.53 & 0.10 & 12.17 & 6.99 & 3.70 \\
MTVNet & \realyes{} & 17.26 & 2.32 & 0.04 & 4.33 & 1.23 & 0.46 \\
MTVNet & \realno{} & 16.71 & 2.42 & 0.09 & 13.52 & 7.65 & 3.92 \\
PIN & \textendash{} & 14.26 & 1.60 & 0.02 & 22.64 & 12.29 & 5.42 \\
\end{longtable}
\endgroup

\subsection{Training split and scale curriculum sensitivity}
\label{sec:training-protocol-sensitivity}
To examine the effect of training-object selection and scale curriculum, we retrain the three strongest author-released configurations by ASSD using PIN's 13-object training split and direct $4\times$ training from random initialization. The released checkpoints use 14 training objects and staged $2\times$-to-$4\times$ training; the controls exclude Femur\_74 and use no $2\times$ pretraining. Each method retains its architecture, synthetic supervision, and method-specific optimization settings. We evaluate the released and retrained checkpoints on the same acquired-LR test set using the same evaluator. These controls change the training-object split and scale path together. Their optimization settings remain method-specific, and their results supplement the official 30-configuration benchmark.

All three retraining runs use seed~8400, a configured batch size of eight and accumulation factor one, bfloat16 precision, and Adam with $(\beta_1,\beta_2)=(0.9,0.999)$ and zero weight decay. The generator objective is L1 with coefficient one; the other configured generator loss coefficients are zero. Training targeted 100,000 iterations, with the stopping condition checked only between complete data-loader passes. All three configurations specify MultiStepLR with a milestone at 800,000 and a multiplier of 0.5. With 1,248 batches per pass, the runs completed 81 passes and ended at step 101,088; the final checkpoint was used for evaluation. The initial learning rates and configured warmup lengths are
\begin{center}
\begin{tabular}{lrr}
\toprule
Method & Initial learning rate & Configured warmup steps \\
\midrule
mDCSRN & $10^{-4}$ & 0 \\
MTVNet & $2\times10^{-4}$ & 3,000 \\
RRDBNet3D & $10^{-4}$ & 0 \\
\bottomrule
\end{tabular}
\end{center}

\begingroup\fontsize{9}{10.8}\selectfont\setlength{\tabcolsep}{4pt}\renewcommand{\arraystretch}{1.12}\renewcommand{\SITableCaption}{Training-protocol sensitivity of the strongest author-released external configurations. Released checkpoints use the original VoDaSuRe training split and staged $2\times$-to-$4\times$ training, whereas retrained checkpoints use PIN's 13-object split and direct $4\times$ training. All external configurations in this table use synthetic-only training. All results are evaluated under the same test protocol. Finite surface-distance support is reported explicitly.}\begin{longtable}{@{}llrrrrr@{}}
\caption{\SITableCaption}\label{tab:si-training-protocol-sensitivity}\\
\toprule
Configuration & Checkpoint & \shortstack{ASSD$\downarrow$\\(\HRvox{})} & \shortstack{HD95$\downarrow$\\(\HRvox{})} & \shortstack{Surface Dice$\uparrow$\\@1~\HRvox{}} & \shortstack{HR-covered$\uparrow$\\@1~\HRvox{}} & \shortstack{Finite\\ROIs} \\
\midrule
\endfirsthead
\multicolumn{7}{l}{\textit{Table \thetable{} continued}}\\
\toprule
Configuration & Checkpoint & \shortstack{ASSD$\downarrow$\\(\HRvox{})} & \shortstack{HD95$\downarrow$\\(\HRvox{})} & \shortstack{Surface Dice$\uparrow$\\@1~\HRvox{}} & \shortstack{HR-covered$\uparrow$\\@1~\HRvox{}} & \shortstack{Finite\\ROIs} \\
\midrule
\endhead
\midrule\multicolumn{7}{r}{Continued on next page}\\
\endfoot
\bottomrule
\endlastfoot
\multicolumn{7}{@{}l}{\textbf{Native support: bounded scores on 40/40 ROIs}} \\
\addlinespace[2pt]
mDCSRN & Released & 1.724 & 5.855 & 0.5669 & 0.4568 & 38/40 \\
mDCSRN & Direct $4\times$ retrained & 1.750 & 5.932 & 0.5662 & 0.4567 & 39/40 \\
MTVNet & Released & 1.839 & 6.183 & 0.5537 & 0.4453 & 38/40 \\
MTVNet & Direct $4\times$ retrained & 1.753 & 5.877 & 0.5628 & 0.4564 & 38/40 \\
RRDBNet3D & Released & 1.968 & 6.538 & 0.5393 & 0.4308 & 38/40 \\
RRDBNet3D & Direct $4\times$ retrained & 1.558 & 5.389 & 0.5965 & 0.4881 & 39/40 \\
PIN & Seed 0 & 0.995 & 3.473 & 0.7046 & 0.6137 & 40/40 \\
\addlinespace[5pt]
\multicolumn{7}{@{}l}{\textbf{Common finite ROI support: all four metrics on shared ROIs}} \\
\addlinespace[3pt]
\multicolumn{7}{@{}l}{mDCSRN / PIN} \\
PIN & Seed 0 & 0.968 & 3.302 & 0.7090 & 0.6188 & 38/40 \\
mDCSRN & Released & 1.724 & 5.855 & 0.5739 & 0.4640 & 38/40 \\
mDCSRN & Direct $4\times$ retrained & 1.682 & 5.722 & 0.5735 & 0.4638 & 38/40 \\
\addlinespace[3pt]
\multicolumn{7}{@{}l}{MTVNet / PIN} \\
PIN & Seed 0 & 0.968 & 3.302 & 0.7090 & 0.6188 & 38/40 \\
MTVNet & Released & 1.839 & 6.183 & 0.5607 & 0.4523 & 38/40 \\
MTVNet & Direct $4\times$ retrained & 1.753 & 5.877 & 0.5674 & 0.4611 & 38/40 \\
\addlinespace[3pt]
\multicolumn{7}{@{}l}{RRDBNet3D / PIN} \\
PIN & Seed 0 & 0.968 & 3.302 & 0.7090 & 0.6188 & 38/40 \\
RRDBNet3D & Released & 1.968 & 6.538 & 0.5459 & 0.4373 & 38/40 \\
RRDBNet3D & Direct $4\times$ retrained & 1.499 & 5.168 & 0.6034 & 0.4953 & 38/40 \\
\end{longtable}
\endgroup

With the same 13-object training split and direct-$4\times$ scale path, PIN retains lower ASSD and higher Surface Dice than all three retrained external baselines (Table~\ref{tab:si-training-protocol-sensitivity}).

\subsection{Model parameter counts}
\label{sec:model-parameters}
Table~\ref{tab:si-model-parameters} reports the parameter counts of the 16 SR architectures. PIN and the seven independently trained INR baselines use counts calculated from the full SR networks; the eight VoDaSuRe architectures use author-reported values. All PIN supervision variants and component ablations share the same SR network architecture.

\begingroup\fontsize{9}{10.8}\selectfont\setlength{\tabcolsep}{4pt}\renewcommand{\arraystretch}{1.12}\renewcommand{\SITableCaption}{Parameter counts of the 16 SR architectures, expressed in millions (M). The source column distinguishes counts calculated from the full SR networks from values reported in the VoDaSuRe supplementary material.}\begin{longtable}{@{}lcrl@{}}
\caption{\SITableCaption}\label{tab:si-model-parameters}\\
\toprule
SR architecture & Type & Parameters (M) & Source \\
\midrule
\endfirsthead
\multicolumn{4}{l}{\textit{Table \thetable{} continued}}\\
\toprule
SR architecture & Type & Parameters (M) & Source \\
\midrule
\endhead
\midrule\multicolumn{4}{r}{Continued on next page}\\
\endfoot
\bottomrule
\endlastfoot
PIN & 3D & 15.499361 & Counted from full SR network \\
MetaSR & 2D & 1.665560 & Counted from full SR network \\
LIIF & 2D & 1.567323 & Counted from full SR network \\
LTE & 2D & 1.714267 & Counted from full SR network \\
SRNO & 2D & 2.015323 & Counted from full SR network \\
CiaoSR & 2D & 2.685982 & Counted from full SR network \\
HIIF & 2D & 2.955511 & Counted from full SR network \\
LINF & 2D & 1.925656 & Counted from full SR network \\
RCAN & 2D & 15.6 & VoDaSuRe supplement, Table 2~\citesupp{si-ref1} \\
HAT & 2D & 20.8 & VoDaSuRe supplement, Table 2~\citesupp{si-ref1} \\
EDDSR & 3D & 0.8 & VoDaSuRe supplement, Table 2~\citesupp{si-ref1} \\
MFER & 3D & 1.7 & VoDaSuRe supplement, Table 2~\citesupp{si-ref1} \\
mDCSRN & 3D & 1.7 & VoDaSuRe supplement, Table 2~\citesupp{si-ref1} \\
RRDBNet3D & 3D & 26.1 & VoDaSuRe supplement, Table 2~\citesupp{si-ref1} \\
SuperFormer & 3D & 20.4 & VoDaSuRe supplement, Table 2~\citesupp{si-ref1} \\
MTVNet (L3) & 3D & 67.0 & VoDaSuRe supplement, Table 2~\citesupp{si-ref1} \\
\end{longtable}
\endgroup

\section{Strategy, component, and repeated-seed results}
\label{sec:internal-results}

\subsection{Supervision-strategy comparison}
\label{sec:si-strategies}
Synth-only, Real-only, Direct Mix, and PIN share the SR model and fourfold evaluation scale, enabling comparison of their supervision designs under the same backbone. The first two isolate the exact synthetic and registered physical-pair point-supervision routes. Direct Mix optimizes both raw point losses with coefficient one and equal coordinate query counts. PIN's HR anchor assigns registered-HR point teaching to replay, while its LR anchor constrains the actual acquired-LR prediction through observation compatibility, jointly with exact synthetic supervision. The following tables report all four strategies and their image, surface, voxel-overlap, and material measurements at all reported tolerances.

The seed-0 results show stronger interface recovery and material selectivity for PIN. Relative to Direct Mix, Surface Dice@1 increases from 0.4735 to 0.7046, HR-covered@1 from 0.3921 to 0.6137, and Precision from 0.8294 to 0.8574; Unsupported volume at 1~\HRvox{} decreases from 2.26\% to 1.60\%. Absolute VF error decreases from 0.0720 to 0.0411 on the shared ten-unit support. The material-error decomposition, reference-pore erasure, and opposing-face path results in Sec.~\ref{sec:material-topology} provide complementary measurements of material recovery and porous-structure retention.

\begingroup\fontsize{9}{10.8}\selectfont\setlength{\tabcolsep}{4pt}\renewcommand{\arraystretch}{1.12}\renewcommand{\SITableCaption}{\textbf{Supervision-strategy image results.} Full-volume and ROI-core scopes are separate; PSNR is in dB. All four rows use their seed-0 run and have ten finite evaluation-unit summaries.}\par\noindent\begin{minipage}{\textwidth}
\centering\captionsetup{hypcap=false}
\captionof{table}{\SITableCaption}\label{tab:si-compact-strategy-images}
\begin{tabular}{@{}>{\raggedright\arraybackslash}p{1.220in}>{\centering\arraybackslash}p{0.813in}>{\centering\arraybackslash}p{0.813in}>{\centering\arraybackslash}p{0.813in}>{\centering\arraybackslash}p{0.813in}>{\centering\arraybackslash}p{0.813in}>{\centering\arraybackslash}p{0.813in}@{}}
\toprule
Configuration & \shortstack{Volume\\PSNR$\uparrow$} & \shortstack{Volume\\SSIM$\uparrow$} & \shortstack{Volume\\LPIPS$\downarrow$} & \shortstack{ROI\\PSNR$\uparrow$} & \shortstack{ROI\\SSIM$\uparrow$} & \shortstack{ROI\\LPIPS$\downarrow$} \\
\midrule
Synth-only & 15.400 & 0.3393 & 0.5330 & 17.290 & 0.4762 & 0.4064 \\
Real-only & 16.355 & 0.4143 & 0.4909 & 18.805 & 0.5565 & 0.3774 \\
Direct Mix & 16.394 & 0.4107 & 0.4825 & 18.661 & 0.5397 & 0.3737 \\
PIN & 15.090 & 0.3878 & 0.4514 & 17.250 & 0.5221 & 0.3452 \\
\bottomrule
\end{tabular}
\end{minipage}\par\vspace{8pt}
\endgroup

\begingroup\fontsize{9}{10.8}\selectfont\setlength{\tabcolsep}{4pt}\renewcommand{\arraystretch}{1.12}\renewcommand{\SITableCaption}{\textbf{Supervision-strategy surface distances and support.} Directions remain separate. Distances are in \HRvox{}; finite object and ROI counts accompany every endpoint.}\par\noindent\begin{minipage}{\textwidth}
\centering\captionsetup{hypcap=false}
\captionof{table}{\SITableCaption}\label{tab:si-compact-strategy-distances}
\begin{tabular}{@{}>{\raggedright\arraybackslash}p{1.220in}>{\centering\arraybackslash}p{0.813in}>{\centering\arraybackslash}p{0.813in}>{\centering\arraybackslash}p{0.813in}>{\centering\arraybackslash}p{0.813in}>{\centering\arraybackslash}p{0.813in}>{\centering\arraybackslash}p{0.813in}@{}}
\toprule
Configuration & \shortstack{ASD$\downarrow$\\HR$\to$SR} & \shortstack{ASD$\downarrow$\\SR$\to$HR} & ASSD$\downarrow$ & HD95$\downarrow$ & \shortstack{Finite\\objects} & \shortstack{Finite\\ROIs} \\
\midrule
Synth-only & 4.610 & 0.531 & 4.080 & 11.599 & 10/10 & 34/40 \\
Real-only & 5.055 & 0.391 & 4.521 & 12.815 & 10/10 & 30/40 \\
Direct Mix & 5.024 & 0.381 & 4.509 & 12.646 & 10/10 & 33/40 \\
PIN & 1.231 & 0.556 & 0.995 & 3.473 & 10/10 & 40/40 \\
\bottomrule
\end{tabular}
\end{minipage}\par\vspace{8pt}
\endgroup

\begingroup\fontsize{9}{10.8}\selectfont\setlength{\tabcolsep}{4pt}\renewcommand{\arraystretch}{1.12}\renewcommand{\SITableCaption}{\textbf{Supervision-strategy reference-interface coverage.} SDice and HRcov are area fractions at 0.5, 1, and 2~\HRvox{}. All cells use ten units and 40 ROIs.}\par\noindent\begin{minipage}{\textwidth}
\centering\captionsetup{hypcap=false}
\captionof{table}{\SITableCaption}\label{tab:si-compact-strategy-reference-coverage}
\begin{tabular}{@{}>{\raggedright\arraybackslash}p{1.220in}>{\centering\arraybackslash}p{0.813in}>{\centering\arraybackslash}p{0.813in}>{\centering\arraybackslash}p{0.813in}>{\centering\arraybackslash}p{0.813in}>{\centering\arraybackslash}p{0.813in}>{\centering\arraybackslash}p{0.813in}@{}}
\toprule
Configuration & \shortstack{SDice$\uparrow$\\0.5} & \shortstack{SDice$\uparrow$\\1} & \shortstack{SDice$\uparrow$\\2} & \shortstack{HRcov$\uparrow$\\0.5} & \shortstack{HRcov$\uparrow$\\1} & \shortstack{HRcov$\uparrow$\\2} \\
\midrule
Synth-only & 0.2033 & 0.3723 & 0.4814 & 0.1412 & 0.2808 & 0.3968 \\
Real-only & 0.2899 & 0.4682 & 0.5398 & 0.2299 & 0.3920 & 0.4759 \\
Direct Mix & 0.2934 & 0.4735 & 0.5499 & 0.2294 & 0.3921 & 0.4812 \\
PIN & 0.4021 & 0.7046 & 0.8687 & 0.3282 & 0.6137 & 0.8107 \\
\bottomrule
\end{tabular}
\end{minipage}\par\vspace{8pt}
\endgroup

\begingroup\fontsize{9}{10.8}\selectfont\setlength{\tabcolsep}{4pt}\renewcommand{\arraystretch}{1.12}\renewcommand{\SITableCaption}{\textbf{Supervision-strategy voxel overlap and reconstructed-surface support.} Voxel overlap has no distance tolerance; SRsup uses the three HR-grid tolerances in \HRvox{}. All cells use ten units and 40 ROIs.}\par\noindent\begin{minipage}{\textwidth}
\centering\captionsetup{hypcap=false}
\captionof{table}{\SITableCaption}\label{tab:si-compact-strategy-voxel-reverse-coverage}
\begin{tabular}{@{}>{\raggedright\arraybackslash}p{1.220in}>{\centering\arraybackslash}p{0.680in}>{\centering\arraybackslash}p{0.680in}>{\centering\arraybackslash}p{0.680in}>{\centering\arraybackslash}p{0.680in}>{\centering\arraybackslash}p{0.680in}>{\centering\arraybackslash}p{0.680in}>{\centering\arraybackslash}p{0.680in}@{}}
\toprule
Configuration & Dice$\uparrow$ & Jaccard$\uparrow$ & Precision$\uparrow$ & Recall$\uparrow$ & \shortstack{SRsup$\uparrow$\\0.5} & \shortstack{SRsup$\uparrow$\\1} & \shortstack{SRsup$\uparrow$\\2} \\
\midrule
Synth-only & 0.8534 & 0.7468 & 0.8152 & 0.9004 & 0.5585 & 0.8938 & 0.9842 \\
Real-only & 0.8802 & 0.7877 & 0.8299 & 0.9421 & 0.6442 & 0.9466 & 0.9929 \\
Direct Mix & 0.8823 & 0.7911 & 0.8294 & 0.9466 & 0.6487 & 0.9490 & 0.9943 \\
PIN & 0.8109 & 0.6873 & 0.8574 & 0.7736 & 0.5429 & 0.8781 & 0.9796 \\
\bottomrule
\end{tabular}
\end{minipage}\par\vspace{8pt}
\endgroup

\begingroup\fontsize{9}{10.8}\selectfont\setlength{\tabcolsep}{4pt}\renewcommand{\arraystretch}{1.12}\renewcommand{\SITableCaption}{\textbf{Supervision-strategy reference-relative material rates.} Unsupported and Missing are percentages with predicted-solid and reference-solid denominators, respectively. The headers give HR-grid tolerances in \HRvox{}. All cells use ten units and 40 ROIs.}\par\noindent\begin{minipage}{\textwidth}
\centering\captionsetup{hypcap=false}
\captionof{table}{\SITableCaption}\label{tab:si-compact-strategy-material}
\begin{tabular}{@{}>{\raggedright\arraybackslash}p{1.220in}>{\centering\arraybackslash}p{0.813in}>{\centering\arraybackslash}p{0.813in}>{\centering\arraybackslash}p{0.813in}>{\centering\arraybackslash}p{0.813in}>{\centering\arraybackslash}p{0.813in}>{\centering\arraybackslash}p{0.813in}@{}}
\toprule
Configuration & \shortstack{Unsupported$\downarrow$\\0.5} & \shortstack{Unsupported$\downarrow$\\1} & \shortstack{Unsupported$\downarrow$\\2} & \shortstack{Missing$\downarrow$\\0.5} & \shortstack{Missing$\downarrow$\\1} & \shortstack{Missing$\downarrow$\\2} \\
\midrule
Synth-only & 18.48 & 3.01 & 0.16 & 9.96 & 6.40 & 3.83 \\
Real-only & 17.01 & 2.29 & 0.09 & 5.79 & 1.49 & 0.39 \\
Direct Mix & 17.06 & 2.26 & 0.06 & 5.34 & 1.29 & 0.27 \\
PIN & 14.26 & 1.60 & 0.02 & 22.64 & 12.29 & 5.42 \\
\bottomrule
\end{tabular}
\end{minipage}\par\vspace{8pt}
\endgroup

\subsection{Correction-off component controls}
\label{sec:si-components}
A0 retains exact synthetic supervision and both the HR anchor and LR anchor, with the optional correction disabled. A1, A2, and A3 remove synthetic supervision, the HR anchor, and the LR anchor, respectively, from A0. Retained terms keep their original coefficients. PIN adds the optional bounded cotangent correction and retains copy-alarm termination. A0--A3 disable both interventions while preserving the fixed copy thresholds for monitoring, so component-removal behavior can be evaluated at the final checkpoint (Sec.~\ref{sec:correction}). These controls assess the HR and LR anchors as complete components; affine calibration, filtering, and invalid-fit handling are not ablated separately.

\begin{figure}[!ht]
\centering
\includegraphics[width=\textwidth,height=0.84\textheight,keepaspectratio]{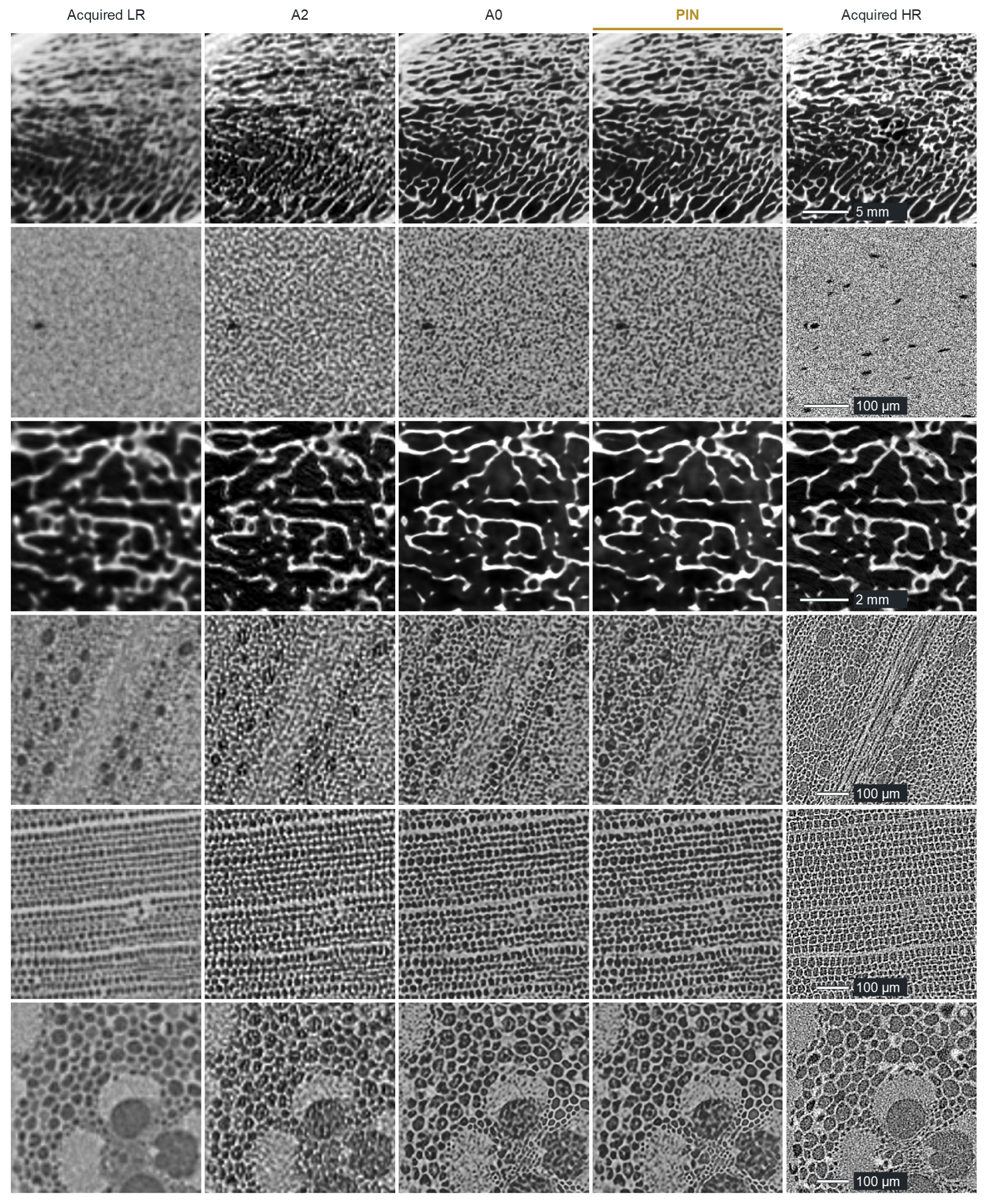}
\caption{\textbf{Artifacts after HR-anchor removal.} A2 develops image artifacts and irregular interfaces across the evaluated objects; six representative bone and wood examples are shown. Columns: acquired LR, A2, A0, PIN, acquired HR. Rows (top to bottom): Femur 01, Ox bone A, Vertebrae D, Oak, Cypress, and Bamboo. A2 retains exact synthetic supervision and the LR anchor at unchanged coefficients; A0 restores the HR anchor with correction off; PIN adds the optional cotangent correction. Within each row, columns share slice, orientation, a $360\times360$ HR-voxel view, and intensity window; LR is interpolated for display.}
\label{fig:a2-comparison}
\end{figure}

A2 has measurements for all ten objects and 40 finite surface-distance ROIs. Removing the HR anchor produces evident artifacts and irregular interfaces across the evaluated objects; Sec.~\ref{sec:a2-gallery} and Fig.~\ref{fig:a2-comparison} show six representative bone and wood examples. Relative to A0, its Surface Dice@1 decreases from 0.7012 to 0.6809, HR-covered@1 from 0.6096 to 0.5873, and precision from 0.8565 to 0.8456; LPIPS increases from 0.4520 to 0.4625. BV/TV absolute error rises from 0.0197 to 0.0269 ($n=3$), and Tb.Th error from 1.655 to 2.467~\HRvox{} on the fixed two-unit support (Sec.~\ref{sec:morphology}). A2 also has a lower absolute VF error than A0 (0.0327 versus 0.0409), despite poorer interface and bone-morphology measurements. This comparison demonstrates that aggregate volume-fraction accuracy and spatial structural accuracy capture different reconstruction properties. Copy-alarm monitoring detects the A2 component-removal anomaly, while the evaluated control retains monitoring without authorizing correction or training termination.

Removing the LR anchor (A3) gives substantially weaker reference-interface recovery than A0, including ASSD 2.857 versus 0.988~\HRvox{} and Surface Dice@1 0.5381 versus 0.7012. The fixed-operator LR residual check in Sec.~\ref{sec:lr-residual} tests the corresponding observation-compatibility behavior. PIN and A0 show small, mixed differences: Surface Dice and HR-covered improve slightly, while A0 has slightly lower ASSD and HD95. Their close agreement indicates that the three-path core $L_{\mathrm{core}}$ provides the primary geometric and morphological gains without cotangent correction, while copy-alarm monitoring detects abnormal copying in A2 (Secs.~\ref{sec:correction} and~\ref{sec:diagnostics}).

\begingroup\fontsize{9}{10.8}\selectfont\setlength{\tabcolsep}{4pt}\renewcommand{\arraystretch}{1.12}\renewcommand{\SITableCaption}{\textbf{Complete component image results.} PSNR is in dB; full-volume and ROI-core scopes remain separate. All five rows use seed-0 checkpoints and have ten finite evaluation-unit summaries.}\par\noindent\begin{minipage}{\textwidth}
\centering\captionsetup{hypcap=false}
\captionof{table}{\SITableCaption}\label{tab:si-compact-component-images}
\begin{tabular}{@{}>{\raggedright\arraybackslash}p{1.220in}>{\centering\arraybackslash}p{0.813in}>{\centering\arraybackslash}p{0.813in}>{\centering\arraybackslash}p{0.813in}>{\centering\arraybackslash}p{0.813in}>{\centering\arraybackslash}p{0.813in}>{\centering\arraybackslash}p{0.813in}@{}}
\toprule
Configuration & \shortstack{Volume\\PSNR$\uparrow$} & \shortstack{Volume\\SSIM$\uparrow$} & \shortstack{Volume\\LPIPS$\downarrow$} & \shortstack{ROI\\PSNR$\uparrow$} & \shortstack{ROI\\SSIM$\uparrow$} & \shortstack{ROI\\LPIPS$\downarrow$} \\
\midrule
A0 & 15.062 & 0.3872 & 0.4520 & 17.126 & 0.5214 & 0.3442 \\
A1 & 14.971 & 0.3535 & 0.4848 & 16.955 & 0.5038 & 0.3821 \\
A2 & 14.523 & 0.3564 & 0.4625 & 16.422 & 0.4885 & 0.3609 \\
A3 & 15.967 & 0.4096 & 0.4586 & 18.065 & 0.5386 & 0.3514 \\
PIN & 15.090 & 0.3878 & 0.4514 & 17.250 & 0.5221 & 0.3452 \\
\bottomrule
\end{tabular}
\end{minipage}\par\vspace{8pt}
\endgroup

\begingroup\fontsize{9}{10.8}\selectfont\setlength{\tabcolsep}{4pt}\renewcommand{\arraystretch}{1.12}\renewcommand{\SITableCaption}{\textbf{Complete component surface distances and support.} Distances are in \HRvox{} and directional ASD remains directional. A0--A3 and PIN retain their finite object/ROI counts.}\par\noindent\begin{minipage}{\textwidth}
\centering\captionsetup{hypcap=false}
\captionof{table}{\SITableCaption}\label{tab:si-compact-component-distances}
\begin{tabular}{@{}>{\raggedright\arraybackslash}p{1.220in}>{\centering\arraybackslash}p{0.813in}>{\centering\arraybackslash}p{0.813in}>{\centering\arraybackslash}p{0.813in}>{\centering\arraybackslash}p{0.813in}>{\centering\arraybackslash}p{0.813in}>{\centering\arraybackslash}p{0.813in}@{}}
\toprule
Configuration & \shortstack{ASD$\downarrow$\\HR$\to$SR} & \shortstack{ASD$\downarrow$\\SR$\to$HR} & ASSD$\downarrow$ & HD95$\downarrow$ & \shortstack{Finite\\objects} & \shortstack{Finite\\ROIs} \\
\midrule
A0 & 1.219 & 0.558 & 0.988 & 3.360 & 10/10 & 40/40 \\
A1 & 1.279 & 0.545 & 1.028 & 3.598 & 10/10 & 40/40 \\
A2 & 1.319 & 0.586 & 1.071 & 3.698 & 10/10 & 40/40 \\
A3 & 3.318 & 0.462 & 2.857 & 9.060 & 10/10 & 37/40 \\
PIN & 1.231 & 0.556 & 0.995 & 3.473 & 10/10 & 40/40 \\
\bottomrule
\end{tabular}
\end{minipage}\par\vspace{8pt}
\endgroup

\begingroup\fontsize{9}{10.8}\selectfont\setlength{\tabcolsep}{4pt}\renewcommand{\arraystretch}{1.12}\renewcommand{\SITableCaption}{\textbf{Complete component reference-interface coverage.} SDice and HRcov are area fractions at 0.5, 1, and 2~\HRvox{}. All cells use ten units and 40 ROIs.}\par\noindent\begin{minipage}{\textwidth}
\centering\captionsetup{hypcap=false}
\captionof{table}{\SITableCaption}\label{tab:si-compact-component-reference-coverage}
\begin{tabular}{@{}>{\raggedright\arraybackslash}p{1.220in}>{\centering\arraybackslash}p{0.813in}>{\centering\arraybackslash}p{0.813in}>{\centering\arraybackslash}p{0.813in}>{\centering\arraybackslash}p{0.813in}>{\centering\arraybackslash}p{0.813in}>{\centering\arraybackslash}p{0.813in}@{}}
\toprule
Configuration & \shortstack{SDice$\uparrow$\\0.5} & \shortstack{SDice$\uparrow$\\1} & \shortstack{SDice$\uparrow$\\2} & \shortstack{HRcov$\uparrow$\\0.5} & \shortstack{HRcov$\uparrow$\\1} & \shortstack{HRcov$\uparrow$\\2} \\
\midrule
A0 & 0.4005 & 0.7012 & 0.8669 & 0.3266 & 0.6096 & 0.8083 \\
A1 & 0.3968 & 0.6944 & 0.8549 & 0.3196 & 0.5989 & 0.7913 \\
A2 & 0.3785 & 0.6809 & 0.8493 & 0.3048 & 0.5873 & 0.7860 \\
A3 & 0.3190 & 0.5381 & 0.6368 & 0.2502 & 0.4498 & 0.5619 \\
PIN & 0.4021 & 0.7046 & 0.8687 & 0.3282 & 0.6137 & 0.8107 \\
\bottomrule
\end{tabular}
\end{minipage}\par\vspace{8pt}
\endgroup

\begingroup\fontsize{9}{10.8}\selectfont\setlength{\tabcolsep}{4pt}\renewcommand{\arraystretch}{1.12}\renewcommand{\SITableCaption}{\textbf{Complete component voxel overlap and reconstructed-surface support.} Voxel overlap has no distance tolerance; SRsup uses the HR-grid tolerances in \HRvox{}. All cells use ten units and 40 ROIs.}\par\noindent\begin{minipage}{\textwidth}
\centering\captionsetup{hypcap=false}
\captionof{table}{\SITableCaption}\label{tab:si-compact-component-voxel-reverse-coverage}
\begin{tabular}{@{}>{\raggedright\arraybackslash}p{1.220in}>{\centering\arraybackslash}p{0.680in}>{\centering\arraybackslash}p{0.680in}>{\centering\arraybackslash}p{0.680in}>{\centering\arraybackslash}p{0.680in}>{\centering\arraybackslash}p{0.680in}>{\centering\arraybackslash}p{0.680in}>{\centering\arraybackslash}p{0.680in}@{}}
\toprule
Configuration & Dice$\uparrow$ & Jaccard$\uparrow$ & Precision$\uparrow$ & Recall$\uparrow$ & \shortstack{SRsup$\uparrow$\\0.5} & \shortstack{SRsup$\uparrow$\\1} & \shortstack{SRsup$\uparrow$\\2} \\
\midrule
A0 & 0.8097 & 0.6857 & 0.8565 & 0.7722 & 0.5420 & 0.8765 & 0.9795 \\
A1 & 0.8152 & 0.6930 & 0.8556 & 0.7825 & 0.5493 & 0.8833 & 0.9810 \\
A2 & 0.8087 & 0.6834 & 0.8456 & 0.7793 & 0.5261 & 0.8676 & 0.9763 \\
A3 & 0.8688 & 0.7697 & 0.8336 & 0.9134 & 0.5996 & 0.9145 & 0.9862 \\
PIN & 0.8109 & 0.6873 & 0.8574 & 0.7736 & 0.5429 & 0.8781 & 0.9796 \\
\bottomrule
\end{tabular}
\end{minipage}\par\vspace{8pt}
\endgroup

\begingroup\fontsize{9}{10.8}\selectfont\setlength{\tabcolsep}{4pt}\renewcommand{\arraystretch}{1.12}\renewcommand{\SITableCaption}{\textbf{Complete component reference-relative material rates.} Unsupported and Missing retain their distinct volume denominators and are percentages at the HR-grid tolerances in \HRvox{}. All cells use ten units and 40 ROIs.}\par\noindent\begin{minipage}{\textwidth}
\centering\captionsetup{hypcap=false}
\captionof{table}{\SITableCaption}\label{tab:si-compact-component-material}
\begin{tabular}{@{}>{\raggedright\arraybackslash}p{1.220in}>{\centering\arraybackslash}p{0.813in}>{\centering\arraybackslash}p{0.813in}>{\centering\arraybackslash}p{0.813in}>{\centering\arraybackslash}p{0.813in}>{\centering\arraybackslash}p{0.813in}>{\centering\arraybackslash}p{0.813in}@{}}
\toprule
Configuration & \shortstack{Unsupported$\downarrow$\\0.5} & \shortstack{Unsupported$\downarrow$\\1} & \shortstack{Unsupported$\downarrow$\\2} & \shortstack{Missing$\downarrow$\\0.5} & \shortstack{Missing$\downarrow$\\1} & \shortstack{Missing$\downarrow$\\2} \\
\midrule
A0 & 14.35 & 1.64 & 0.03 & 22.78 & 12.47 & 5.56 \\
A1 & 14.44 & 1.67 & 0.03 & 21.75 & 11.86 & 5.43 \\
A2 & 15.44 & 1.96 & 0.05 & 22.07 & 11.73 & 4.86 \\
A3 & 16.64 & 2.20 & 0.05 & 8.66 & 2.80 & 0.95 \\
PIN & 14.26 & 1.60 & 0.02 & 22.64 & 12.29 & 5.42 \\
\bottomrule
\end{tabular}
\end{minipage}\par\vspace{8pt}
\endgroup

\subsection{Three-seed descriptive variability}
\label{sec:seeds}
PIN and Direct Mix are evaluated at seeds~0, 1, and~2 for image, surface, voxel-overlap, material, and main morphology endpoints. For each metric, the across-seed mean is the average of three seed-level aggregate estimates, and SD is their sample standard deviation ($n=3$, denominator $n-1$). No ROI or object is counted as an additional training repeat. These seeds revisit the same test units and support a descriptive variability statement.

Both methods retain fixed common morphology support across all three seeds: Femur~01, Ox bone~A, and Vertebrae~D for BV/TV (four valid ROIs per unit), and Femur~01 and Vertebrae~D for Tb.Th (one valid thickness ROI per unit).

PIN's Surface Dice@1 is $0.70425\pm0.00230$, compared with $0.47326\pm0.00333$ for Direct Mix. HR-covered@1 is $0.61300\pm0.00291$ versus $0.39213\pm0.00292$. The PIN--Direct Mix ordering is unchanged across the three seeds for every endpoint in Table~\ref{tab:si-compact-seeds-main} except Tb.Th absolute error. PIN has lower LPIPS, ASSD, HD95, unsupported volume, and BV/TV absolute error, and higher Surface Dice, HR-covered, and precision. Direct Mix has higher PSNR, SSIM, Dice, and recall, and lower missing volume. PIN has lower Tb.Th error at seeds~0 and~1, but slightly higher error at seed~2 (1.64285 versus 1.62147~\HRvox{}). These comparisons are descriptive and do not constitute a significance test.

PIN has finite distance values on all ten units and 40 ROIs at each seed. Direct Mix has finite distance values on all ten units but 33, 32, and 32 of 40 ROIs for seeds~0, 1, and~2. The designated main endpoints remain seed~0. Table~\ref{tab:si-compact-seeds-support} states image, distance, bounded-surface, and material support; the next two tables report every seed, mean, and SD using full-precision seed-level aggregates.

\begingroup\fontsize{9}{10.8}\selectfont\setlength{\tabcolsep}{4pt}\renewcommand{\arraystretch}{1.12}\renewcommand{\SITableCaption}{\textbf{Repeated-seed support.} Object and ROI denominators are ten and 40. Distance support applies to ASSD, HD95, and both directional ASDs; bounded-surface support applies to SDice, HRcov, and SRsup at all reported tolerances. Material and voxel measures have complete support.}\par\noindent\begin{minipage}{\textwidth}
\centering\captionsetup{hypcap=false}
\captionof{table}{\SITableCaption}\label{tab:si-compact-seeds-support}
\begin{tabular}{@{}>{\raggedright\arraybackslash}p{0.950in}>{\centering\arraybackslash}p{0.420in}>{\centering\arraybackslash}p{0.780in}>{\centering\arraybackslash}p{1.200in}>{\centering\arraybackslash}p{1.200in}>{\centering\arraybackslash}p{1.200in}@{}}
\toprule
Method & Seed & \shortstack{Image\\volumes} & \shortstack{Distance\\obj.; ROI} & \shortstack{SDice/HRcov\\obj.; ROI} & \shortstack{Material\\obj.; ROI} \\
\midrule
PIN & 0 & 10/10 & 10/10; 40/40 & 10/10; 40/40 & 10/10; 40/40 \\
PIN & 1 & 10/10 & 10/10; 40/40 & 10/10; 40/40 & 10/10; 40/40 \\
PIN & 2 & 10/10 & 10/10; 40/40 & 10/10; 40/40 & 10/10; 40/40 \\
Direct Mix & 0 & 10/10 & 10/10; 33/40 & 10/10; 40/40 & 10/10; 40/40 \\
Direct Mix & 1 & 10/10 & 10/10; 32/40 & 10/10; 40/40 & 10/10; 40/40 \\
Direct Mix & 2 & 10/10 & 10/10; 32/40 & 10/10; 40/40 & 10/10; 40/40 \\
\bottomrule
\end{tabular}
\end{minipage}\par\vspace{8pt}
\endgroup

\begingroup\fontsize{9}{10.8}\selectfont\setlength{\tabcolsep}{4pt}\renewcommand{\arraystretch}{1.12}\renewcommand{\SITableCaption}{\textbf{PIN and Direct Mix repeated seeds: evaluation endpoints.} Values are seed-level aggregates, followed by their mean and sample SD over three endpoints. Surface and material metrics marked @1, and unqualified Unsupported/Missing, use 1~\HRvox{}. Unsupported and Missing are percentages. BV/TV absolute error is dimensionless on the fixed three-unit support; Tb.Th absolute error is in \HRvox{} on the fixed two-unit support. Other metric support is given in the preceding table.}\begin{longtable}{@{}>{\raggedright\arraybackslash}p{1.600in}>{\centering\arraybackslash}p{0.780in}>{\centering\arraybackslash}p{0.720in}>{\centering\arraybackslash}p{0.720in}>{\centering\arraybackslash}p{0.720in}>{\centering\arraybackslash}p{0.720in}>{\centering\arraybackslash}p{0.720in}@{}}
\caption{\SITableCaption}\label{tab:si-compact-seeds-main}\\
\toprule
Metric & Method & Seed 0 & Seed 1 & Seed 2 & Mean & SD \\
\midrule
\endfirsthead
\multicolumn{7}{l}{\textit{Table \thetable{} continued}}\\
\toprule
Metric & Method & Seed 0 & Seed 1 & Seed 2 & Mean & SD \\
\midrule
\endhead
\midrule\multicolumn{7}{r}{Continued on next page}\\
\endfoot
\bottomrule
\endlastfoot
\shortstack{PSNR-2D$\uparrow$\\(dB)} & PIN & 15.09032 & 15.14337 & 15.17161 & 15.13510 & 0.04127 \\
SSIM-2D$\uparrow$ & PIN & 0.38776 & 0.38593 & 0.38995 & 0.38788 & 0.00201 \\
LPIPS$\downarrow$ & PIN & 0.45140 & 0.44769 & 0.44830 & 0.44913 & 0.00199 \\
\shortstack{ASSD$\downarrow$\\(\HRvox{})} & PIN & 0.99465 & 0.98236 & 1.00539 & 0.99414 & 0.01152 \\
\shortstack{HD95$\downarrow$\\(\HRvox{})} & PIN & 3.47293 & 3.39150 & 3.46872 & 3.44438 & 0.04585 \\
Surface Dice$\uparrow$@1 & PIN & 0.70460 & 0.70635 & 0.70179 & 0.70425 & 0.00230 \\
HR-covered$\uparrow$@1 & PIN & 0.61365 & 0.61552 & 0.60982 & 0.61300 & 0.00291 \\
Dice$\uparrow$ & PIN & 0.81087 & 0.81330 & 0.81543 & 0.81320 & 0.00228 \\
Precision$\uparrow$ & PIN & 0.85742 & 0.85543 & 0.85630 & 0.85638 & 0.00100 \\
Recall$\uparrow$ & PIN & 0.77359 & 0.77914 & 0.78261 & 0.77845 & 0.00455 \\
\shortstack{Unsupported$\downarrow$\\(\%)} & PIN & 1.60135 & 1.66296 & 1.63026 & 1.63153 & 0.03082 \\
\shortstack{Missing$\downarrow$\\(\%)} & PIN & 12.28644 & 11.77344 & 11.59774 & 11.88587 & 0.35785 \\
\shortstack{PSNR-2D$\uparrow$\\(dB)} & Direct Mix & 16.39358 & 16.39546 & 16.37484 & 16.38796 & 0.01140 \\
SSIM-2D$\uparrow$ & Direct Mix & 0.41066 & 0.41161 & 0.41120 & 0.41115 & 0.00048 \\
LPIPS$\downarrow$ & Direct Mix & 0.48255 & 0.47876 & 0.47677 & 0.47936 & 0.00293 \\
\shortstack{ASSD$\downarrow$\\(\HRvox{})} & Direct Mix & 4.50883 & 4.38230 & 3.91150 & 4.26754 & 0.31476 \\
\shortstack{HD95$\downarrow$\\(\HRvox{})} & Direct Mix & 12.64645 & 12.48308 & 11.62891 & 12.25281 & 0.54645 \\
Surface Dice$\uparrow$@1 & Direct Mix & 0.47355 & 0.46979 & 0.47643 & 0.47326 & 0.00333 \\
HR-covered$\uparrow$@1 & Direct Mix & 0.39207 & 0.38924 & 0.39508 & 0.39213 & 0.00292 \\
Dice$\uparrow$ & Direct Mix & 0.88231 & 0.88166 & 0.88136 & 0.88178 & 0.00049 \\
Precision$\uparrow$ & Direct Mix & 0.82942 & 0.83023 & 0.82853 & 0.82939 & 0.00085 \\
Recall$\uparrow$ & Direct Mix & 0.94663 & 0.94489 & 0.94551 & 0.94568 & 0.00089 \\
\shortstack{Unsupported$\downarrow$\\(\%)} & Direct Mix & 2.26480 & 2.24494 & 2.33213 & 2.28062 & 0.04570 \\
\shortstack{Missing$\downarrow$\\(\%)} & Direct Mix & 1.28714 & 1.40684 & 1.29902 & 1.33100 & 0.06595 \\
\shortstack{BV/TV\\absolute error$\downarrow$} & PIN & 0.01852 & 0.01947 & 0.02172 & 0.01990 & 0.00164 \\
\shortstack{Tb.Th absolute error$\downarrow$\\(\HRvox{})} & PIN & 1.42460 & 1.39604 & 1.64285 & 1.48783 & 0.13501 \\
\shortstack{BV/TV\\absolute error$\downarrow$} & Direct Mix & 0.03492 & 0.03790 & 0.03740 & 0.03674 & 0.00159 \\
\shortstack{Tb.Th absolute error$\downarrow$\\(\HRvox{})} & Direct Mix & 1.64588 & 1.46313 & 1.62147 & 1.57683 & 0.09922 \\
\end{longtable}
\endgroup

\begingroup\fontsize{9}{10.8}\selectfont\setlength{\tabcolsep}{4pt}\renewcommand{\arraystretch}{1.12}\renewcommand{\SITableCaption}{\textbf{PIN and Direct Mix repeated seeds: directional surfaces and tolerances.} The two ASDs are in \HRvox{}; bounded-surface values are area fractions and numeric suffixes give tolerances in \HRvox{}. Means and sample SDs use three seed-level estimates. Support is given in Table~\ref{tab:si-compact-seeds-support}.}\begin{longtable}{@{}>{\raggedright\arraybackslash}p{1.600in}>{\centering\arraybackslash}p{0.780in}>{\centering\arraybackslash}p{0.720in}>{\centering\arraybackslash}p{0.720in}>{\centering\arraybackslash}p{0.720in}>{\centering\arraybackslash}p{0.720in}>{\centering\arraybackslash}p{0.720in}@{}}
\caption{\SITableCaption}\label{tab:si-compact-seeds-directional}\\
\toprule
Metric & Method & Seed 0 & Seed 1 & Seed 2 & Mean & SD \\
\midrule
\endfirsthead
\multicolumn{7}{l}{\textit{Table \thetable{} continued}}\\
\toprule
Metric & Method & Seed 0 & Seed 1 & Seed 2 & Mean & SD \\
\midrule
\endhead
\midrule\multicolumn{7}{r}{Continued on next page}\\
\endfoot
\bottomrule
\endlastfoot
\shortstack{ASD$\downarrow$ HR to SR\\(\HRvox{})} & PIN & 1.23056 & 1.21325 & 1.24438 & 1.22940 & 0.01560 \\
\shortstack{ASD$\downarrow$ SR to HR\\(\HRvox{})} & PIN & 0.55568 & 0.55652 & 0.54682 & 0.55301 & 0.00538 \\
Surface Dice$\uparrow$@0.5 \HRvox{} & PIN & 0.40207 & 0.40209 & 0.40007 & 0.40141 & 0.00116 \\
Surface Dice$\uparrow$@1.0 \HRvox{} & PIN & 0.70460 & 0.70635 & 0.70179 & 0.70425 & 0.00230 \\
Surface Dice$\uparrow$@2.0 \HRvox{} & PIN & 0.86866 & 0.86992 & 0.86291 & 0.86717 & 0.00373 \\
HR-covered$\uparrow$@0.5 \HRvox{} & PIN & 0.32824 & 0.32804 & 0.32494 & 0.32708 & 0.00185 \\
HR-covered$\uparrow$@1.0 \HRvox{} & PIN & 0.61365 & 0.61552 & 0.60982 & 0.61300 & 0.00291 \\
HR-covered$\uparrow$@2.0 \HRvox{} & PIN & 0.81073 & 0.81227 & 0.80331 & 0.80877 & 0.00479 \\
SR-supported$\uparrow$@0.5 \HRvox{} & PIN & 0.54293 & 0.54256 & 0.54698 & 0.54416 & 0.00245 \\
SR-supported$\uparrow$@1.0 \HRvox{} & PIN & 0.87814 & 0.87841 & 0.88250 & 0.87969 & 0.00244 \\
SR-supported$\uparrow$@2.0 \HRvox{} & PIN & 0.97958 & 0.97932 & 0.98114 & 0.98001 & 0.00098 \\
\shortstack{ASD$\downarrow$ HR to SR\\(\HRvox{})} & Direct Mix & 5.02397 & 4.90480 & 4.42367 & 4.78415 & 0.31782 \\
\shortstack{ASD$\downarrow$ SR to HR\\(\HRvox{})} & Direct Mix & 0.38087 & 0.38281 & 0.38364 & 0.38244 & 0.00142 \\
Surface Dice$\uparrow$@0.5 \HRvox{} & Direct Mix & 0.29338 & 0.29062 & 0.29491 & 0.29297 & 0.00217 \\
Surface Dice$\uparrow$@1.0 \HRvox{} & Direct Mix & 0.47355 & 0.46979 & 0.47643 & 0.47326 & 0.00333 \\
Surface Dice$\uparrow$@2.0 \HRvox{} & Direct Mix & 0.54991 & 0.54627 & 0.55380 & 0.54999 & 0.00376 \\
HR-covered$\uparrow$@0.5 \HRvox{} & Direct Mix & 0.22943 & 0.22690 & 0.23070 & 0.22901 & 0.00194 \\
HR-covered$\uparrow$@1.0 \HRvox{} & Direct Mix & 0.39207 & 0.38924 & 0.39508 & 0.39213 & 0.00292 \\
HR-covered$\uparrow$@2.0 \HRvox{} & Direct Mix & 0.48123 & 0.47829 & 0.48525 & 0.48159 & 0.00349 \\
SR-supported$\uparrow$@0.5 \HRvox{} & Direct Mix & 0.64866 & 0.65083 & 0.64838 & 0.64929 & 0.00134 \\
SR-supported$\uparrow$@1.0 \HRvox{} & Direct Mix & 0.94898 & 0.94769 & 0.94670 & 0.94779 & 0.00114 \\
SR-supported$\uparrow$@2.0 \HRvox{} & Direct Mix & 0.99432 & 0.99335 & 0.99384 & 0.99384 & 0.00049 \\
\end{longtable}
\endgroup

Focused morphology summaries appear in Sec.~\ref{sec:morphology}.

\section{Threshold sensitivity and morphology}

\subsection{Segmentation threshold and distance tolerance}
\label{sec:thresholds}
The sensitivity analysis measures frozen outputs at five offsets from the object-specific HR-calibrated threshold: $-0.05$, $-0.025$, $0$, $+0.025$, and $+0.05$. The threshold shifts are applied to both HR and SR masks. Distance tolerances are 0.5, 1, and 2~\HRvox{}. It changes the measurement threshold without retraining a model or selecting a new endpoint.

PIN exceeds all 30 external configurations in Surface Dice and HR-covered at every threshold--tolerance setting. At 1~\HRvox{}, PIN's lowest Surface Dice across the tested offsets (0.6762 at $\Delta T=-0.050$) remains above the highest external value at any offset (0.6000 at $\Delta T=+0.050$), and its lowest HR-covered score (0.5926) likewise remains above the highest external value (0.4863). Tables~\ref{tab:si-compact-threshold-surface-dice} and~\ref{tab:si-compact-threshold-hr-covered} compare PIN with the largest external value at each setting, retaining the comparator identity. Synthetic-trained mDCSRN is the strongest external in 29 of the 30 metric-settings; at $\Delta T=+0.05$ and 0.5~\HRvox{}, the largest external HR-covered value is 0.2632 from real-trained RRDBNet3D, compared with PIN's 0.3330. This sensitivity result concerns the two stated bounded-surface measures. It does not assert unchanged ordering in reverse-direction support, unbounded distance, or every morphology metric.

Surface Dice pools covered HR and SR surface areas over their combined total area; HR-covered uses total HR surface area as its denominator. Their formulas are given in Sec.~\ref{sec:metrics}. Every setting retains ten object summaries and defined bounded-surface scores on all 40 ROIs for PIN and the displayed external comparator. Defined zero coverage from an empty SR is retained; bounded-score support remains distinct from finite surface-distance evaluability. Margins in the tables are percentage-point differences between the displayed four-decimal values.

\begingroup\fontsize{9}{10.8}\selectfont\setlength{\tabcolsep}{4pt}\renewcommand{\arraystretch}{1.12}\renewcommand{\SITableCaption}{\textbf{Surface Dice threshold sensitivity.} Each row selects the largest external Surface Dice among 30 configurations at the stated threshold offset and HR-grid tolerance. PIN and the displayed external retain ten object summaries and 40/40 defined ROI scores at each offset. Values are area fractions; margin is PIN minus external in percentage points (pp).}\begin{longtable}{@{}>{\raggedright\arraybackslash}p{0.650in}>{\centering\arraybackslash}p{0.680in}>{\centering\arraybackslash}p{0.800in}>{\centering\arraybackslash}p{1.080in}>{\centering\arraybackslash}p{0.259in}>{\centering\arraybackslash}p{0.850in}>{\centering\arraybackslash}p{0.750in}@{}}
\caption{\SITableCaption}\label{tab:si-compact-threshold-surface-dice}\\
\toprule
$\Delta T$ & $\tau$ (\HRvox{}) & PIN$\uparrow$ & Best external & Real & External$\uparrow$ & \shortstack{Margin\\(pp)} \\
\midrule
\endfirsthead
\multicolumn{7}{l}{\textit{Table \thetable{} continued}}\\
\toprule
$\Delta T$ & $\tau$ (\HRvox{}) & PIN$\uparrow$ & Best external & Real & External$\uparrow$ & \shortstack{Margin\\(pp)} \\
\midrule
\endhead
\midrule\multicolumn{7}{r}{Continued on next page}\\
\endfoot
\bottomrule
\endlastfoot
-0.050 & 0.5 & 0.3811 & mDCSRN & \realno{} & 0.2890 & +9.21 \\
-0.050 & 1 & 0.6762 & mDCSRN & \realno{} & 0.5017 & +17.45 \\
-0.050 & 2 & 0.8416 & mDCSRN & \realno{} & 0.6240 & +21.76 \\
-0.025 & 0.5 & 0.3930 & mDCSRN & \realno{} & 0.3096 & +8.34 \\
-0.025 & 1 & 0.6930 & mDCSRN & \realno{} & 0.5392 & +15.38 \\
-0.025 & 2 & 0.8579 & mDCSRN & \realno{} & 0.6703 & +18.76 \\
+0.000 & 0.5 & 0.4021 & mDCSRN & \realno{} & 0.3247 & +7.74 \\
+0.000 & 1 & 0.7046 & mDCSRN & \realno{} & 0.5669 & +13.77 \\
+0.000 & 2 & 0.8687 & mDCSRN & \realno{} & 0.7046 & +16.41 \\
+0.025 & 0.5 & 0.4089 & mDCSRN & \realno{} & 0.3349 & +7.40 \\
+0.025 & 1 & 0.7131 & mDCSRN & \realno{} & 0.5862 & +12.69 \\
+0.025 & 2 & 0.8758 & mDCSRN & \realno{} & 0.7293 & +14.65 \\
+0.050 & 0.5 & 0.4147 & mDCSRN & \realno{} & 0.3421 & +7.26 \\
+0.050 & 1 & 0.7199 & mDCSRN & \realno{} & 0.6000 & +11.99 \\
+0.050 & 2 & 0.8810 & mDCSRN & \realno{} & 0.7472 & +13.38 \\
\end{longtable}
\endgroup

\begingroup\fontsize{9}{10.8}\selectfont\setlength{\tabcolsep}{4pt}\renewcommand{\arraystretch}{1.12}\renewcommand{\SITableCaption}{\textbf{HR-covered threshold sensitivity.} Each row selects the largest external HR-covered among 30 configurations at the stated offset and tolerance. PIN and the displayed external retain ten object summaries and 40/40 defined ROI scores at each offset. Values are HR-area fractions; margin is in percentage points. Comparator changes are retained.}\begin{longtable}{@{}>{\raggedright\arraybackslash}p{0.650in}>{\centering\arraybackslash}p{0.680in}>{\centering\arraybackslash}p{0.800in}>{\centering\arraybackslash}p{1.080in}>{\centering\arraybackslash}p{0.259in}>{\centering\arraybackslash}p{0.850in}>{\centering\arraybackslash}p{0.750in}@{}}
\caption{\SITableCaption}\label{tab:si-compact-threshold-hr-covered}\\
\toprule
$\Delta T$ & $\tau$ (\HRvox{}) & PIN$\uparrow$ & Best external & Real & External$\uparrow$ & \shortstack{Margin\\(pp)} \\
\midrule
\endfirsthead
\multicolumn{7}{l}{\textit{Table \thetable{} continued}}\\
\toprule
$\Delta T$ & $\tau$ (\HRvox{}) & PIN$\uparrow$ & Best external & Real & External$\uparrow$ & \shortstack{Margin\\(pp)} \\
\midrule
\endhead
\midrule\multicolumn{7}{r}{Continued on next page}\\
\endfoot
\bottomrule
\endlastfoot
-0.050 & 0.5 & 0.3175 & mDCSRN & \realno{} & 0.2127 & +10.48 \\
-0.050 & 1 & 0.5926 & mDCSRN & \realno{} & 0.3954 & +19.72 \\
-0.050 & 2 & 0.7782 & mDCSRN & \realno{} & 0.5283 & +24.99 \\
-0.025 & 0.5 & 0.3241 & mDCSRN & \realno{} & 0.2291 & +9.50 \\
-0.025 & 1 & 0.6056 & mDCSRN & \realno{} & 0.4308 & +17.48 \\
-0.025 & 2 & 0.7978 & mDCSRN & \realno{} & 0.5791 & +21.87 \\
+0.000 & 0.5 & 0.3282 & mDCSRN & \realno{} & 0.2408 & +8.74 \\
+0.000 & 1 & 0.6137 & mDCSRN & \realno{} & 0.4568 & +15.69 \\
+0.000 & 2 & 0.8107 & mDCSRN & \realno{} & 0.6172 & +19.35 \\
+0.025 & 0.5 & 0.3308 & mDCSRN & \realno{} & 0.2484 & +8.24 \\
+0.025 & 1 & 0.6190 & mDCSRN & \realno{} & 0.4744 & +14.46 \\
+0.025 & 2 & 0.8195 & mDCSRN & \realno{} & 0.6445 & +17.50 \\
+0.050 & 0.5 & 0.3330 & RRDBNet3D & \realyes{} & 0.2632 & +6.98 \\
+0.050 & 1 & 0.6235 & mDCSRN & \realno{} & 0.4863 & +13.72 \\
+0.050 & 2 & 0.8260 & mDCSRN & \realno{} & 0.6639 & +16.21 \\
\end{longtable}
\endgroup

\subsection{Morphology eligibility and quality control}
\label{sec:morphology}
Thickness and local diameter use BoneJ2 ThicknessWrapper with 7.2.2 core components, both solid and void maps, and artefact masking. The local diameter at a phase voxel is the diameter of the largest sphere contained in that phase and containing the voxel. The trabecular profile names the solid measurement Tb.Th; other profiles name it local solid diameter. The shared HR-derived segmentation reference is applied to both HR and SR. Computation uses the separate $128^3$ \texttt{th\_01} core with a 64-HR-voxel field-of-view margin and halo context. The HR-grid diameter map $D$ for each HR or SR phase is used directly, without another factor of two or spacing conversion. For each HR or SR phase, $\overline D$ is the arithmetic mean of finite positive diameter voxels in the core; the absolute mean-diameter error is $|\overline D_{\mathrm{SR}}-\overline D_{\mathrm{HR}}|$.

Thickness halos start at 32 HR voxels per side and are capped at 64. For each HR/SR solid/void phase, $R_{99}=P_{99}(D)/2$ is computed from the same finite positive phase-core diameter samples and guides expansion. If one phase has $2R_{99}$ greater than the HR-grid width of the capped halo and none has a finite $2R_{99}$ greater than the initial HR-grid halo width but within that cap, the initial calculation retains only phases already fitting its halo without a second-halo comparison. Otherwise the halo doubles toward the cap where possible. A phase whose $2R_{99}$ exceeds that HR-grid halo cap is unavailable. Outside the initial shortcut, comparisons require equal core-sample counts and sorted diameter vectors satisfying $|D_{\mathrm{new}}-D_{\mathrm{old}}|\leq\max(10^{-4}\,\HRvox{},10^{-6}|D_{\mathrm{new}}|)$. At the cap, converged phases are accepted even with artificial crop-face contact; phases that fail the comparison are unavailable. If no comparison is possible outside the shortcut, a phase requires $2R_{99}$ to fit the available HR-grid halo width and contact with neither a true volume face nor an artificial crop face. QC is applied separately to HR/SR and solid/void; empty samples, failed computation, or failed phase QC give NaN on the affected side, and errors require both sides to be finite.

Morphology applicability is determined by the object profile. For an applicable metric, the HR-reference computation must return a finite value before an SR--HR error can be formed. Native support therefore has three stages: profile applicability, a finite HR result, and a finite SR result within that finite-HR set. An applicable case without a finite HR result is excluded from that run's error denominator; unavailable SR within the finite-HR set is an SR evaluability failure. Unavailable values are not imputed, and these support counts do not enlarge the number of independent test units.

Table~\ref{tab:si-compact-main-table-morphology} reproduces the seed-0 BV/TV and Tb.Th absolute errors in all three main tables. BV/TV uses the predefined three-unit bone-profile subset: Femur~01, Ox bone~A, and Vertebrae~D. Main-table Tb.Th uses a fixed two-unit bone-profile support: Femur~01 and Vertebrae~D, both eligible and evaluable across the displayed configurations. Tb.Th denotes trabecular thickness; solid $\varnothing$ denotes local solid diameter.

\begingroup\fontsize{9}{10.8}\selectfont\setlength{\tabcolsep}{4pt}\renewcommand{\arraystretch}{1.12}\renewcommand{\SITableCaption}{\textbf{Main-table morphology summary.} Rows reproduce the configurations in Main Tables~1--3. BV/TV absolute error is dimensionless and uses the predefined three-unit bone-profile subset; Tb.Th absolute error is in \HRvox{} and uses the fixed two-unit support. All values are seed-0 results.}\par\noindent\begin{minipage}{\textwidth}
\centering\captionsetup{hypcap=false}
\captionof{table}{\SITableCaption}\label{tab:si-compact-main-table-morphology}
\begin{tabular}{@{}lcrr@{}}
\toprule
Configuration & Real & BV/TV abs. error$\downarrow$ & Tb.Th abs. error$\downarrow$ (\HRvox{}) \\
\midrule
\multicolumn{4}{@{}l}{\textbf{Main Table 1: external configurations}} \\
mDCSRN & \realno{} & 0.0394 & 5.326 \\
MTVNet & \realno{} & 0.0402 & 5.804 \\
RRDBNet3D & \realno{} & 0.0388 & 5.713 \\
MFER & \realno{} & 0.0411 & 5.460 \\
EDDSR & \realno{} & 0.0409 & 5.445 \\
SuperFormer & \realno{} & 0.0414 & 5.971 \\
RCAN & \realno{} & 0.0429 & 4.968 \\
HAT & \realno{} & 0.0418 & 5.172 \\
RRDBNet3D & \realyes{} & 0.0436 & 1.590 \\
HAT & \realyes{} & 0.0431 & 2.054 \\
SuperFormer & \realyes{} & 0.0397 & 2.456 \\
mDCSRN & \realyes{} & 0.0433 & 1.778 \\
MFER & \realyes{} & 0.0451 & 1.724 \\
HIIF & \realyes{} & 0.0404 & 1.465 \\
LTE & \realyes{} & 0.0399 & 2.243 \\
MetaSR & \realyes{} & 0.0403 & 2.402 \\
PIN & \textendash{} & 0.0185 & 1.425 \\
\addlinespace[2pt]
\multicolumn{4}{@{}l}{\textbf{Main Table 2: supervision strategies}} \\
Synth-only & \textendash{} & 0.0436 & 7.365 \\
Real-only & \textendash{} & 0.0374 & 1.720 \\
Direct Mix & \textendash{} & 0.0349 & 1.646 \\
PIN & \textendash{} & 0.0185 & 1.425 \\
\addlinespace[2pt]
\multicolumn{4}{@{}l}{\textbf{Main Table 3: components}} \\
A0 & \textendash{} & 0.0197 & 1.655 \\
A1 & \textendash{} & 0.0201 & 1.982 \\
A2 & \textendash{} & 0.0269 & 2.467 \\
A3 & \textendash{} & 0.0430 & 2.276 \\
PIN & \textendash{} & 0.0185 & 1.425 \\
\bottomrule
\end{tabular}
\end{minipage}\par\vspace{8pt}
\endgroup

Across the full external inventory, the three lowest external BV/TV absolute errors on the predefined three-bone-unit support are 0.0388 for synthetic-trained RRDBNet3D, 0.0394 for synthetic-trained mDCSRN, and 0.0397 for real-trained SuperFormer. PIN's 0.0185 corresponds to a 52.2--53.4\% reduction. All three external configurations also appear in the representative Main Table~1.

Table~\ref{tab:si-compact-thickness-eligibility} reports native applicability and finite HR/SR support for each run. Tb.Th has two finite HR and two finite SR measurements in every row. Local solid diameter is applicable to eight objects in every row, all with finite HR results. Real-trained HAT on Cardboard and real-trained mDCSRN on Bamboo lack a valid paired local-diameter measurement because the SR phase fails QC. Each SR-finite fraction uses that run's finite-HR count as its denominator. Native complete-case means do not replace the separately defined main-table two-unit Tb.Th errors. Paired errors in Table~\ref{tab:si-compact-paired-thickness} retain common finite HR/SR support with the three strongest external distance comparators and each metric's own denominator.

\begingroup\fontsize{9}{10.8}\selectfont\setlength{\tabcolsep}{4pt}\renewcommand{\arraystretch}{1.12}\renewcommand{\SITableCaption}{\textbf{Native thickness and local-diameter applicability and finite support.} Applicability counts objects with the relevant profile; HR finite counts finite reference results. Each SR-finite fraction gives the number of finite SR results within that run's finite-HR set over the size of that set.}\begin{longtable}{@{}>{\raggedright\arraybackslash}p{0.850in}>{\centering\arraybackslash}p{0.259in}>{\centering\arraybackslash}p{0.813in}>{\centering\arraybackslash}p{0.813in}>{\centering\arraybackslash}p{0.813in}>{\centering\arraybackslash}p{0.813in}>{\centering\arraybackslash}p{0.813in}>{\centering\arraybackslash}p{0.813in}@{}}
\caption{\SITableCaption}\label{tab:si-compact-thickness-eligibility}\\
\toprule
Configuration & Real & \shortstack{Tb.Th\\applicable} & \shortstack{Tb.Th\\HR finite} & \shortstack{Tb.Th\\SR finite} & \shortstack{Solid $\varnothing$\\applicable} & \shortstack{Solid $\varnothing$\\HR finite} & \shortstack{Solid $\varnothing$\\SR finite} \\
\midrule
\endfirsthead
\multicolumn{8}{l}{\textit{Table \thetable{} continued}}\\
\toprule
Configuration & Real & \shortstack{Tb.Th\\applicable} & \shortstack{Tb.Th\\HR finite} & \shortstack{Tb.Th\\SR finite} & \shortstack{Solid $\varnothing$\\applicable} & \shortstack{Solid $\varnothing$\\HR finite} & \shortstack{Solid $\varnothing$\\SR finite} \\
\midrule
\endhead
\midrule\multicolumn{8}{r}{Continued on next page}\\
\endfoot
\bottomrule
\endlastfoot
MetaSR & \realyes{} & 2 & 2 & 2/2 & 8 & 8 & 2/8 \\
MetaSR & \realno{} & 2 & 2 & 2/2 & 8 & 8 & 2/8 \\
LIIF & \realyes{} & 2 & 2 & 2/2 & 8 & 8 & 3/8 \\
LIIF & \realno{} & 2 & 2 & 2/2 & 8 & 8 & 3/8 \\
LTE & \realyes{} & 2 & 2 & 2/2 & 8 & 8 & 3/8 \\
LTE & \realno{} & 2 & 2 & 2/2 & 8 & 8 & 3/8 \\
SRNO & \realyes{} & 2 & 2 & 2/2 & 8 & 8 & 3/8 \\
SRNO & \realno{} & 2 & 2 & 2/2 & 8 & 8 & 3/8 \\
CiaoSR & \realyes{} & 2 & 2 & 2/2 & 8 & 8 & 2/8 \\
CiaoSR & \realno{} & 2 & 2 & 2/2 & 8 & 8 & 3/8 \\
HIIF & \realyes{} & 2 & 2 & 2/2 & 8 & 8 & 3/8 \\
HIIF & \realno{} & 2 & 2 & 2/2 & 8 & 8 & 1/8 \\
LINF & \realyes{} & 2 & 2 & 2/2 & 8 & 8 & 1/8 \\
LINF & \realno{} & 2 & 2 & 2/2 & 8 & 8 & 1/8 \\
RCAN & \realyes{} & 2 & 2 & 2/2 & 8 & 8 & 1/8 \\
RCAN & \realno{} & 2 & 2 & 2/2 & 8 & 8 & 4/8 \\
HAT & \realyes{} & 2 & 2 & 2/2 & 8 & 8 & 1/8 \\
HAT & \realno{} & 2 & 2 & 2/2 & 8 & 8 & 4/8 \\
EDDSR & \realyes{} & 2 & 2 & 2/2 & 8 & 8 & 1/8 \\
EDDSR & \realno{} & 2 & 2 & 2/2 & 8 & 8 & 6/8 \\
MFER & \realyes{} & 2 & 2 & 2/2 & 8 & 8 & 2/8 \\
MFER & \realno{} & 2 & 2 & 2/2 & 8 & 8 & 7/8 \\
mDCSRN & \realyes{} & 2 & 2 & 2/2 & 8 & 8 & 3/8 \\
mDCSRN & \realno{} & 2 & 2 & 2/2 & 8 & 8 & 7/8 \\
RRDBNet3D & \realyes{} & 2 & 2 & 2/2 & 8 & 8 & 4/8 \\
RRDBNet3D & \realno{} & 2 & 2 & 2/2 & 8 & 8 & 6/8 \\
SuperFormer & \realyes{} & 2 & 2 & 2/2 & 8 & 8 & 3/8 \\
SuperFormer & \realno{} & 2 & 2 & 2/2 & 8 & 8 & 6/8 \\
MTVNet & \realyes{} & 2 & 2 & 2/2 & 8 & 8 & 2/8 \\
MTVNet & \realno{} & 2 & 2 & 2/2 & 8 & 8 & 6/8 \\
PIN & \textendash{} & 2 & 2 & 2/2 & 8 & 8 & 8/8 \\
\end{longtable}
\endgroup

\begingroup\fontsize{9}{10.8}\selectfont\setlength{\tabcolsep}{4pt}\renewcommand{\arraystretch}{1.12}\renewcommand{\SITableCaption}{\textbf{Focused paired morphology outcomes.} PIN and the three synthetic-trained main external distance comparators are compared only on objects with finite HR and SR measurements for both methods. Absolute thickness and local-diameter errors are in \HRvox{}; each measurement retains its own common-object count.}\par\noindent\begin{minipage}{\textwidth}
\centering\captionsetup{hypcap=false}
\captionof{table}{\SITableCaption}\label{tab:si-compact-paired-thickness}
\begin{tabular}{@{}>{\raggedright\arraybackslash}p{1.220in}>{\centering\arraybackslash}p{0.813in}>{\centering\arraybackslash}p{0.813in}>{\centering\arraybackslash}p{0.813in}>{\centering\arraybackslash}p{0.813in}>{\centering\arraybackslash}p{0.813in}>{\centering\arraybackslash}p{0.813in}@{}}
\toprule
Baseline & \shortstack{Tb.Th\\common obj.} & \shortstack{PIN\\abs. error$\downarrow$} & \shortstack{Baseline\\abs. error$\downarrow$} & \shortstack{Solid $\varnothing$\\common obj.} & \shortstack{PIN\\abs. error$\downarrow$} & \shortstack{Baseline\\abs. error$\downarrow$} \\
\midrule
mDCSRN & 2 & 1.425 & 5.326 & 7 & 4.626 & 10.912 \\
MTVNet & 2 & 1.425 & 5.804 & 6 & 4.874 & 11.251 \\
RRDBNet3D & 2 & 1.425 & 5.713 & 6 & 4.874 & 12.635 \\
\bottomrule
\end{tabular}
\end{minipage}\par\vspace{8pt}
\endgroup

The control summaries below supplement the material-recovery results with signed and absolute solid-fraction error and porosity error. BV/TV and SV/TV retain their profile-specific object supports, shown in the cells. This separates the direction of a volume-fraction change from its error magnitude.

\begingroup\fontsize{9}{10.8}\selectfont\setlength{\tabcolsep}{4pt}\renewcommand{\arraystretch}{1.12}\renewcommand{\SITableCaption}{\textbf{Focused strategy and component morphology summary.} BV/TV absolute error uses the three predefined bone-profile units; SV/TV signed and absolute errors use the seven nonbone units. VF/Porosity absolute error uses all ten evaluation units and 40 ROIs, with per-ROI absolute errors aggregated within and across units. For the same binary segmentation, absolute solid-volume-fraction and porosity errors are mathematically identical. All values are dimensionless.}\par\noindent\begin{minipage}{\textwidth}
\centering\captionsetup{hypcap=false}
\captionof{table}{\SITableCaption}\label{tab:si-compact-control-morphology}
\begin{tabular}{@{}>{\raggedright\arraybackslash}p{1.220in}>{\centering\arraybackslash}p{1.100in}>{\centering\arraybackslash}p{1.100in}>{\centering\arraybackslash}p{1.100in}>{\centering\arraybackslash}p{1.100in}@{}}
\toprule
Configuration & \shortstack{BV/TV\\abs. error$\downarrow$} & \shortstack{SV/TV\\signed error} & \shortstack{SV/TV\\abs. error$\downarrow$} & \shortstack{VF / Porosity\\abs. error$\downarrow$} \\
\midrule
Synth-only & \shortstack{0.0436\\(3/10 obj)} & \shortstack{0.0722\\(7/10 obj)} & \shortstack{0.0722\\(7/10 obj)} & 0.0636 \\
Real-only & \shortstack{0.0374\\(3/10 obj)} & \shortstack{0.0849\\(7/10 obj)} & \shortstack{0.0849\\(7/10 obj)} & 0.0707 \\
Direct Mix & \shortstack{0.0349\\(3/10 obj)} & \shortstack{0.0879\\(7/10 obj)} & \shortstack{0.0879\\(7/10 obj)} & 0.0720 \\
PIN & \shortstack{0.0185\\(3/10 obj)} & \shortstack{-0.0507\\(7/10 obj)} & \shortstack{0.0508\\(7/10 obj)} & 0.0411 \\
A0 & \shortstack{0.0197\\(3/10 obj)} & \shortstack{-0.0499\\(7/10 obj)} & \shortstack{0.0500\\(7/10 obj)} & 0.0409 \\
A1 & \shortstack{0.0201\\(3/10 obj)} & \shortstack{-0.0441\\(7/10 obj)} & \shortstack{0.0448\\(7/10 obj)} & 0.0374 \\
A2 & \shortstack{0.0269\\(3/10 obj)} & \shortstack{-0.0329\\(7/10 obj)} & \shortstack{0.0352\\(7/10 obj)} & 0.0327 \\
A3 & \shortstack{0.0430\\(3/10 obj)} & \shortstack{0.0613\\(7/10 obj)} & \shortstack{0.0613\\(7/10 obj)} & 0.0558 \\
\bottomrule
\end{tabular}
\end{minipage}\par\vspace{8pt}
\endgroup

\subsection{Material recovery and pore topology diagnostics}
\label{sec:material-topology}

To distinguish false-positive material from reference-solid omission, we decompose each binary HR/SR ROI into true-positive (TP), false-positive (FP), and false-negative (FN) solid voxels. For \(N\) core voxels, the signed solid-volume-fraction error is \((FP-FN)/N\), while the per-ROI absolute volume-fraction error is \(|FP-FN|/N\). Each metric is aggregated over ROIs within an evaluation unit and then equally across test units. These calculations use the shared HR-derived segmentation and the same 40 ROI cores as the benchmark.

\begingroup\fontsize{9}{10.8}\selectfont\setlength{\tabcolsep}{4pt}\renewcommand{\arraystretch}{1.12}\renewcommand{\SITableCaption}{Material error decomposition on the shared HR-derived binary segmentation. All entries use ten evaluation units and 40 ROI cores. FP/N and FN/N are percentages of all core voxels; signed solid-volume bias is reported in percentage points, and absolute VF error is dimensionless. Absolute VF error is computed within each ROI before nested aggregation. The signed bias equals FP/N minus FN/N after accounting for the displayed units. Counts are derived from the archived frozen evaluation masks.}\begin{longtable}{@{}>{\raggedright\arraybackslash}p{1.080in}>{\centering\arraybackslash}p{0.259in}rrrr@{}}
\caption{\SITableCaption}\label{tab:si-material-decomposition}\\
\toprule
Configuration & Real & FP/$N$$\downarrow$ (\%) & FN/$N$$\downarrow$ (\%) & \shortstack{Signed solid-volume\\bias (pp)} & \shortstack{Absolute VF error$\downarrow$\\(dimensionless)} \\
\midrule
\endfirsthead
\multicolumn{6}{l}{\textit{Table \thetable{} continued}}\\
\toprule
Configuration & Real & FP/$N$$\downarrow$ (\%) & FN/$N$$\downarrow$ (\%) & \shortstack{Signed solid-volume\\bias (pp)} & \shortstack{Absolute VF error$\downarrow$\\(dimensionless)} \\
\midrule
\endhead
\midrule\multicolumn{6}{r}{Continued on next page}\\
\endfoot
\bottomrule
\endlastfoot
\multicolumn{6}{l}{\textit{Supervision strategies and component controls}}\\
Synth-only & \textendash{} & 8.9018 & 3.2577 & 5.6441 & 0.0636 \\
Real-only & \textendash{} & 8.7519 & 1.9514 & 6.8005 & 0.0707 \\
Direct Mix & \textendash{} & 8.8074 & 1.7685 & 7.0389 & 0.0720 \\
PIN & \textendash{} & 5.5857 & 9.2413 & -3.6556 & 0.0411 \\
A0 & \textendash{} & 5.6194 & 9.2180 & -3.5987 & 0.0409 \\
A1 & \textendash{} & 5.7221 & 8.9279 & -3.2058 & 0.0374 \\
A2 & \textendash{} & 6.2458 & 8.6460 & -2.4002 & 0.0327 \\
A3 & \textendash{} & 8.1684 & 3.1989 & 4.9695 & 0.0558 \\
\midrule
\multicolumn{6}{l}{\textit{External configurations}}\\
CiaoSR & \realno{} & 8.8170 & 3.5643 & 5.2528 & 0.0591 \\
CiaoSR & \realyes{} & 9.3156 & 1.5562 & 7.7594 & 0.0819 \\
EDDSR & \realno{} & 7.6555 & 4.9310 & 2.7245 & 0.0377 \\
EDDSR & \realyes{} & 9.4436 & 1.6745 & 7.7691 & 0.0850 \\
HAT & \realno{} & 8.0926 & 4.4570 & 3.6356 & 0.0437 \\
HAT & \realyes{} & 8.9544 & 2.5940 & 6.3605 & 0.0712 \\
HIIF & \realno{} & 9.2494 & 3.0524 & 6.1970 & 0.0696 \\
HIIF & \realyes{} & 9.0593 & 2.0608 & 6.9984 & 0.0748 \\
LIIF & \realno{} & 8.8475 & 3.5232 & 5.3243 & 0.0603 \\
LIIF & \realyes{} & 9.2015 & 1.7132 & 7.4883 & 0.0794 \\
LINF & \realno{} & 8.6462 & 3.8287 & 4.8175 & 0.0563 \\
LINF & \realyes{} & 9.3651 & 1.8022 & 7.5629 & 0.0837 \\
LTE & \realno{} & 8.8231 & 3.5568 & 5.2663 & 0.0597 \\
LTE & \realyes{} & 9.2515 & 1.6176 & 7.6339 & 0.0803 \\
mDCSRN & \realno{} & 7.3223 & 5.4070 & 1.9154 & 0.0341 \\
mDCSRN & \realyes{} & 9.2391 & 1.3870 & 7.8520 & 0.0824 \\
MetaSR & \realno{} & 8.8412 & 3.5388 & 5.3024 & 0.0601 \\
MetaSR & \realyes{} & 9.1371 & 1.8083 & 7.3288 & 0.0781 \\
MFER & \realno{} & 7.6830 & 5.2846 & 2.3983 & 0.0369 \\
MFER & \realyes{} & 9.1090 & 1.6370 & 7.4721 & 0.0793 \\
MTVNet & \realno{} & 7.3081 & 5.4332 & 1.8750 & 0.0358 \\
MTVNet & \realyes{} & 9.1221 & 1.3193 & 7.8028 & 0.0817 \\
RCAN & \realno{} & 7.9670 & 4.9817 & 2.9853 & 0.0386 \\
RCAN & \realyes{} & 8.9285 & 2.6059 & 6.3226 & 0.0715 \\
RRDBNet3D & \realno{} & 7.4633 & 5.0003 & 2.4631 & 0.0378 \\
RRDBNet3D & \realyes{} & 8.8635 & 1.7968 & 7.0667 & 0.0744 \\
SRNO & \realno{} & 8.8401 & 3.5310 & 5.3091 & 0.0600 \\
SRNO & \realyes{} & 9.5668 & 1.3440 & 8.2229 & 0.0847 \\
SuperFormer & \realno{} & 7.5405 & 5.0472 & 2.4933 & 0.0386 \\
SuperFormer & \realyes{} & 8.9330 & 1.6112 & 7.3219 & 0.0763 \\
\end{longtable}
\endgroup

\begin{figure}[!ht]
\centering
\includegraphics[width=\textwidth]{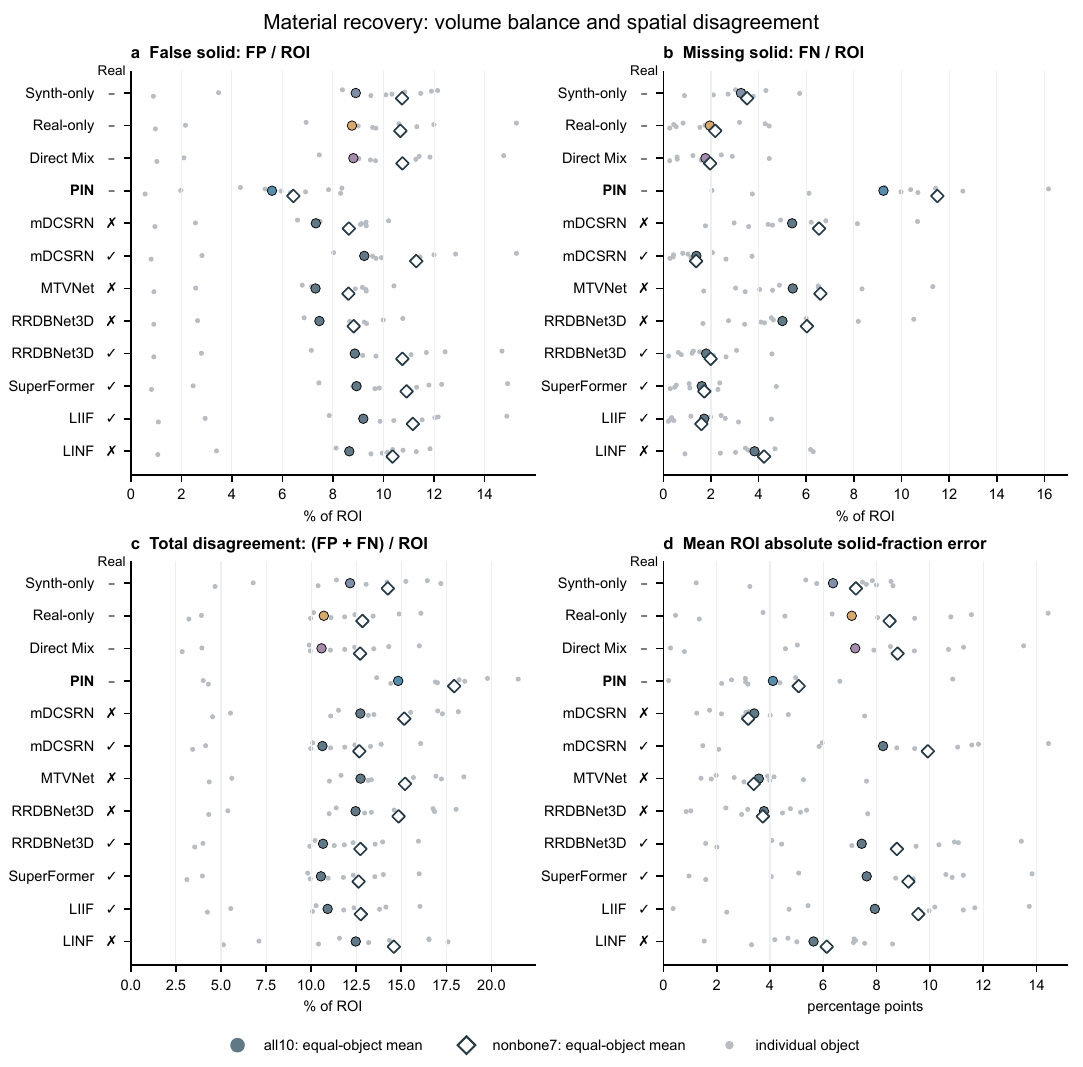}
\caption{Material error decomposition across supervision strategies. FP/N and FN/N quantify false-positive and false-negative solid voxels as percentages of ROI volume; total disagreement summarizes their sum, and absolute VF error measures the volume-fraction discrepancy. Object-level results and the all-ten-unit and seven-nonbone-unit summaries are shown on their respective supports. Real: \realyes{} denotes real-pair training and \realno{} synthetic-only training; dashes mark the named supervision strategies.}
\label{fig:si-material-decomposition}
\end{figure}

Across the four same-backbone strategies and all 30 external configurations, PIN has the lowest FP/N. Real-only and Direct Mix increase voxel coverage relative to Synth-only primarily through reduced FN, with no corresponding increase in FP/N. Compared with Direct Mix, PIN reduces FP/N from 8.81\% to 5.59\% but increases FN/N from 1.77\% to 9.24\%. Total voxel disagreement \((FP+FN)/N\) counts both error directions, whereas per-ROI absolute VF error \(|FP-FN|/N\) allows false-positive and false-negative volumes to offset. Neither metric captures the spatial arrangement of these errors, motivating their joint interpretation with interface recovery and pore preservation. PIN therefore records higher total voxel disagreement than Direct Mix (14.83\% versus 10.58\%), alongside lower absolute VF error (0.0411 versus 0.0720) and a lower fully erased reference-pore volume share (10.41\% versus 53.43\%; Tables~\ref{tab:si-material-decomposition}--\ref{tab:si-pore-topology}). PIN's negative signed solid-volume bias is concentrated on the seven non-bone units ($-5.07$~pp, compared with $-3.66$~pp overall), while its three-unit bone BV/TV absolute error is 0.0185 (Table~\ref{tab:si-compact-control-morphology}). For PIN, Missing decreases from 22.64\% at 0.5~\HRvox{} to 12.29\% at 1~\HRvox{} and 5.42\% at 2~\HRvox{} (Tables~\ref{tab:si-compact-external-material} and~\ref{tab:si-compact-strategy-material}). These correspond to relative reductions of 45.7\% and 76.1\% under the same object-equal aggregation, indicating that much of the reference-solid coverage deficit lies close to reconstructed solid. Figure~\ref{fig:si-material-decomposition} visualizes the material-error decomposition across supervision strategies.

We further evaluate reference-pore erasure and opposing-face path retention on three-dimensional HR/SR masks using 6-connected components and a 32-voxel halo. An HR-void component is eligible for erasure when it intersects the ROI core, is fully contained within the halo-expanded region, and contains at least 64 voxels across the full component. A reference pore is classified as fully erased when SR labels every voxel of that component as solid. Opposing-face path loss records reference-void connections between opposing core faces that are no longer connected in SR. Connectivity is evaluated in the halo-expanded ROI, allowing paths between opposing core-face endpoints to extend outside the core. Each measurement uses its predefined reference support and eligibility criteria.

The fully erased reference-pore volume share weights each eligible component by its HR-void volume within the ROI core. Numerator and denominator volumes are pooled across each object's ROIs, and the resulting shares are averaged equally across the ten objects. Opposing-face path losses are counted against 250 reference opportunities across the same 40 ROI cores. All 40 cores are processed, and 39 contain eligible reference components for the erasure analysis.

The spatial diagnostic covers all 30 external configurations from 15 method families and the four shared-backbone supervision strategies. Every configuration has exact regenerated-to-archived core-mask agreement across all 40 HR-aligned ROI cores. Complete per-configuration results are reported in Table~\ref{tab:si-pore-topology}.

\begingroup\fontsize{9}{10.8}\selectfont\setlength{\tabcolsep}{4pt}\renewcommand{\arraystretch}{1.12}\renewcommand{\SITableCaption}{Three-dimensional reference-pore erasure and opposing-face path retention on frozen HR/SR masks. Erasure uses fully contained HR-void components of at least 64 voxels across the halo-expanded region. Erasure shares use eligible HR core-volume weighting, within-object pooling, and equal-object averaging. Opposing-face path losses are counted against 250 reference opportunities. All 34 configurations have 40/40 exact core-mask agreement, with eligible reference components in 39/40 cores for the erasure analysis. Lower values indicate better performance for both metrics.}\begin{longtable}{@{}>{\raggedright\arraybackslash}p{0.880in}>{\centering\arraybackslash}p{0.259in}>{\centering\arraybackslash}p{3.600in}>{\centering\arraybackslash}p{1.100in}@{}}
\caption{\SITableCaption}\label{tab:si-pore-topology}\\
\toprule
Method & Real & \shortstack{Fully erased reference-pore\\volume share$\downarrow$} & \shortstack{Lost HR face\\paths$\downarrow$} \\
\midrule
\endfirsthead
\multicolumn{4}{l}{\textit{Table \thetable{} continued}}\\
\toprule
Method & Real & \shortstack{Fully erased reference-pore\\volume share$\downarrow$} & \shortstack{Lost HR face\\paths$\downarrow$} \\
\midrule
\endhead
\midrule\multicolumn{4}{r}{Continued on next page}\\
\endfoot
\bottomrule
\endlastfoot
\multicolumn{4}{l}{\textit{Shared-backbone supervision strategies}}\\
Synth-only & \textendash{} & 68.89\% & 108/250 \\
Real-only & \textendash{} & 53.90\% & 64/250 \\
Direct Mix & \textendash{} & 53.43\% & 65/250 \\
PIN & \textendash{} & 10.41\% & 39/250 \\
\midrule
\multicolumn{4}{l}{\textit{Complete external benchmark: 15 families, Synthetic and Real}}\\
CiaoSR & \realno{} & 68.77\% & 105/250 \\
CiaoSR & \realyes{} & 59.97\% & 77/250 \\
EDDSR & \realno{} & 45.59\% & 90/250 \\
EDDSR & \realyes{} & 63.57\% & 100/250 \\
HAT & \realno{} & 54.55\% & 88/250 \\
HAT & \realyes{} & 59.61\% & 103/250 \\
HIIF & \realno{} & 72.92\% & 135/250 \\
HIIF & \realyes{} & 60.33\% & 78/250 \\
LIIF & \realno{} & 69.37\% & 110/250 \\
LIIF & \realyes{} & 58.75\% & 77/250 \\
LINF & \realno{} & 68.37\% & 99/250 \\
LINF & \realyes{} & 64.40\% & 97/250 \\
LTE & \realno{} & 69.19\% & 108/250 \\
LTE & \realyes{} & 58.28\% & 78/250 \\
mDCSRN & \realno{} & 37.33\% & 82/250 \\
mDCSRN & \realyes{} & 58.41\% & 70/250 \\
MetaSR & \realno{} & 69.33\% & 111/250 \\
MetaSR & \realyes{} & 57.42\% & 77/250 \\
MFER & \realno{} & 43.92\% & 81/250 \\
MFER & \realyes{} & 60.01\% & 75/250 \\
MTVNet & \realno{} & 39.33\% & 84/250 \\
MTVNet & \realyes{} & 57.38\% & 72/250 \\
RCAN & \realno{} & 50.12\% & 80/250 \\
RCAN & \realyes{} & 61.65\% & 92/250 \\
RRDBNet3D & \realno{} & 40.22\% & 74/250 \\
RRDBNet3D & \realyes{} & 55.50\% & 64/250 \\
SRNO & \realno{} & 69.11\% & 107/250 \\
SRNO & \realyes{} & 63.30\% & 84/250 \\
SuperFormer & \realno{} & 43.99\% & 80/250 \\
SuperFormer & \realyes{} & 52.89\% & 62/250 \\
\end{longtable}
\endgroup

Across the four shared-backbone strategies, the fully erased reference-pore volume share is 68.89\% for Synth-only, 53.90\% for Real-only, and 53.43\% for Direct Mix, compared with 10.41\% for PIN. PIN also loses fewer reference opposing-face paths: 39 of 250, compared with 108, 64, and 65 for the three alternative strategies. These results demonstrate substantially reduced reference-pore erasure and better retention of opposing-face reference pathways.

Across the complete external benchmark, fully erased reference-pore volume shares range from 37.33\% to 72.92\% among the 30 external configurations, compared with PIN's 10.41\%. Lost reference opposing-face paths range from 62 to 135 out of the 250 reference opportunities, compared with 39 for PIN. All comparisons use the same 40 ROI cores and the metric-specific eligibility criteria defined above. PIN achieves the lowest erasure share and the fewest lost reference opposing-face paths among all 30 external configurations.

\section{Fixed-operator and frozen-output checks}

\subsection{Learned HR-to-LR proxy and fixed operators}
\label{sec:proxy-validation}
The operator analysis compares the frozen learned HR-to-LR proxy with fixed downsampling and blur operators on registered pairs. Here $(L_r,H_r)$ denotes the acquired-LR/registered-HR pair for evaluation volume $v$, and $D_k$ is the compared HR-to-LR operator indexed by $k$. The discrepancy is $N_v^{-1}\|G_\sigma*A_v(D_k(H_r))-G_\sigma*L_r\|_1$ with $\sigma=1$ LR voxel, where $N_v$ counts every voxel of the resulting LR volume. Validation and test support are reported separately. The volume-scoped affine map is fitted for each operator against acquired LR. These results characterize the comparison bridge for the present dataset.

The Gaussian-PSF operator uses zero-padded HR Gaussian filtering with radius $\lceil3\sigma_{\mathrm{HR}}\rceil$, followed by nonoverlapping $4^3$ box averaging. The frozen $\sigma_{\mathrm{HR}}=2$ HR voxels is selected from $\{0.5,1,1.5,2,2.5,3\}$ using the mean of finite training-volume filtered LR L1 values after the same affine fit. This HR operator blur is distinct from the shared $\sigma=1$ LR-voxel comparison filter.

The fixed MTF kernel is estimated from full training volumes after box downsampling. Whole-volume 3D real FFTs of box-averaged HR and acquired LR accumulate the real cross-spectrum and box-HR power over all 13 training volumes in 64 radial bins. The binned cross-spectrum is divided by the binned input power plus $\epsilon=10^{-8}\max(\overline P,10^{-12})$, where $\overline P$ is the mean binned box-HR power. The radial response is smoothed with a Gaussian of $\sigma=1$ bin using nearest-edge handling, clipped to $[0.05,1.25]$, and interpolated on the real-FFT grid of a $32^3$ volume. An inverse real FFT, centered $9^3$ crop, reversal symmetrization, and unit-sum normalization give the frozen LR-grid kernel, applied after box averaging.

On the test support, the frozen learned proxy has the lowest filtered LR L1 among the compared operators. The validation ordering differs: box-average downsampling gives 0.0191 compared with 0.0196 for the learned proxy. Gaussian-PSF test support is nine objects; the other operators have ten. The trabecular-bone profile in Table~\ref{tab:si-compact-proxy-profiles} comprises the two whole held-out test scans, Femur 01 and Vertebrae D. On these specimen-disjoint scans, Learned-D achieves a mean filtered LR $L_1$ error of 0.0329 versus 0.1440 for box-average downsampling, demonstrating that its advantage also holds beyond axial-split specimens.

\begingroup\fontsize{9}{10.8}\selectfont\setlength{\tabcolsep}{4pt}\renewcommand{\arraystretch}{1.12}\renewcommand{\SITableCaption}{\textbf{Frozen learned proxy and fixed-operator comparison.} The registered volume/object is the reporting unit. Validation and test support are separate. Filtered L1 is a normalized-intensity discrepancy; test mean and SD summarize the object-level results. Raw PSNR is in dB; raw SSIM remains auxiliary.}\par\noindent\begin{minipage}{\textwidth}
\centering\captionsetup{hypcap=false}
\captionof{table}{\SITableCaption}\label{tab:si-compact-proxy-comparison}
\begin{tabular}{@{}>{\raggedright\arraybackslash}p{1.670in}>{\centering\arraybackslash}p{0.320in}>{\centering\arraybackslash}p{0.640in}>{\centering\arraybackslash}p{0.320in}>{\centering\arraybackslash}p{1.050in}>{\centering\arraybackslash}p{0.620in}>{\centering\arraybackslash}p{0.630in}>{\centering\arraybackslash}p{0.630in}@{}}
\toprule
Operator & \shortstack{Val.\\$n$} & \shortstack{Val.\\filtered L1$\downarrow$} & \shortstack{Test\\$n$} & \shortstack{Test filtered L1$\downarrow$\\mean $\pm$ SD} & \shortstack{Test\\median$\downarrow$} & \shortstack{Raw\\PSNR$\uparrow$} & \shortstack{Raw\\SSIM$\uparrow$} \\
\midrule
Learned-D & 1 & 0.0196 & 10 & 0.0308 $\pm$ 0.0210 & 0.0237 & 23.848 & 0.6051 \\
Box-average + x4 & 1 & 0.0191 & 10 & 0.0901 $\pm$ 0.0633 & 0.0800 & 17.508 & 0.3664 \\
Gaussian PSF + x4 & 1 & 0.0391 & 9 & 0.1352 $\pm$ 0.0740 & 0.1300 & 14.042 & 0.4022 \\
Lanczos-3 LP + x4 & 1 & 0.0365 & 10 & 0.0918 $\pm$ 0.0531 & 0.0815 & 17.224 & 0.3723 \\
Training-calibrated fixed MTF & 1 & 0.0231 & 10 & 0.1006 $\pm$ 0.0812 & 0.0746 & 17.019 & 0.5060 \\
Bicubic/tricubic resize & 1 & 0.0402 & 10 & 0.1171 $\pm$ 0.0385 & 0.1187 & 14.743 & 0.2451 \\
\bottomrule
\end{tabular}
\end{minipage}\par\vspace{8pt}
\endgroup

\begingroup\fontsize{9}{10.8}\selectfont\setlength{\tabcolsep}{4pt}\renewcommand{\arraystretch}{1.12}\renewcommand{\SITableCaption}{\textbf{Frozen learned-proxy measurements by morphology profile.} The profile labels group the ten test units and retain their counts. Filtered LR L1 uses the same comparison band; raw LR image measurements are auxiliary.}\par\noindent\begin{minipage}{\textwidth}
\centering\captionsetup{hypcap=false}
\captionof{table}{\SITableCaption}\label{tab:si-compact-proxy-profiles}
\begin{tabular}{@{}>{\raggedright\arraybackslash}p{2.150in}>{\centering\arraybackslash}p{0.500in}>{\centering\arraybackslash}p{1.000in}>{\centering\arraybackslash}p{1.000in}>{\centering\arraybackslash}p{1.000in}@{}}
\toprule
Profile & $n$ & \shortstack{Filtered\\LR L1$\downarrow$} & \shortstack{Raw LR\\PSNR$\uparrow$} & \shortstack{Raw LR\\SSIM$\uparrow$} \\
\midrule
Compact/porous bone & 1 & 0.0198 & 23.305 & 0.3766 \\
Trabecular bone & 2 & 0.0329 & 27.086 & 0.4078 \\
Layered fibrous & 1 & 0.0203 & 23.995 & 0.7878 \\
Porous fibrous & 1 & 0.0207 & 24.298 & 0.7047 \\
Porous tubular & 5 & 0.0363 & 22.541 & 0.6734 \\
\bottomrule
\end{tabular}
\end{minipage}\par\vspace{8pt}
\endgroup

\subsection{A0--A3 filtered LR residual}
\label{sec:lr-residual}
This diagnostic compares existing A0 and A3 predictions with acquired LR using one frozen learned projection operator, affine intensity calibration, and the fixed $\sigma=1$ LR-voxel Gaussian filter. Unlike training's crop-scoped $A_b$, the diagnostic recomputes one $A_v$ per full registered evaluation volume by fitting $D_\phi(H_r)$ to acquired $L_r$. For a candidate HR reconstruction $H$ on that volume's grid, it shares that fit across the two predictions and measures
\begin{equation}
R_v(H)=\frac{1}{N_v}\left\|G_\sigma*A_v(D_\phi(H))-G_\sigma*L_r\right\|_1.
\label{eq:si-volume-residual}
\end{equation}
The fit uses finite valid intensities in $[0.05,0.98]$; the primary residual averages all voxels. Proxy forward passes use HR tiles of side 128 with halo 48. Intensities are normalized by dividing uint16 values by 65,535. The shared Gaussian has $\sigma=1$ LR voxel and zero padding. This is a full-volume LR-domain check, distinct from the crop-scoped training loss, registered-HR image similarity, and surface metrics.

On the ten paired registered test units, the mean filtered residual is 0.0237955 for A0 and 0.0511444 for A3. The paired A3-minus-A0 difference has mean 0.0273489, median 0.0252769, and sample SD 0.0225867. A3 has the larger residual on nine of ten units. The per-unit table includes this exception. Together with the component results, this offline check supports the LR anchor's observation-compatibility interpretation as an association measured on existing predictions.

\begingroup\fontsize{9}{10.8}\selectfont\setlength{\tabcolsep}{4pt}\renewcommand{\arraystretch}{1.12}\renewcommand{\SITableCaption}{\textbf{A0--A3 filtered LR residual on all ten test units.} Both outputs use the same frozen proxy, volume-scoped affine fit, and Gaussian filter. Values are normalized-intensity mean absolute discrepancies; the difference is A3 minus A0 and the ratio is A3/A0. Per-unit values are rounded for display; aggregate statistics above are calculated before rounding.}\par\noindent\begin{minipage}{\textwidth}
\centering\captionsetup{hypcap=false}
\captionof{table}{\SITableCaption}\label{tab:si-compact-lr-residual-by-unit}
\begin{tabular}{@{}>{\raggedright\arraybackslash}p{2.650in}>{\centering\arraybackslash}p{0.820in}>{\centering\arraybackslash}p{0.820in}>{\centering\arraybackslash}p{0.900in}>{\centering\arraybackslash}p{0.650in}@{}}
\toprule
Evaluation unit & A0$\downarrow$ & A3$\downarrow$ & A3 $-$ A0 & Ratio \\
\midrule
Bamboo\_A\_bin1x1\_ome\_0 & 0.0126 & 0.0630 & 0.0504 & 4.988 \\
Cardboard\_A\_bin1x1\_ome\_0 & 0.0070 & 0.0412 & 0.0342 & 5.920 \\
Cypress\_A\_bin1x1\_ome\_0 & 0.0218 & 0.0272 & 0.0054 & 1.247 \\
Elm\_A\_bin1x1\_ome\_0 & 0.0796 & 0.1082 & 0.0286 & 1.359 \\
Femur\_01\_80kV\_ome & 0.0221 & 0.0236 & 0.0015 & 1.069 \\
Larch\_B\_bin1x1\_ome\_0 & 0.0086 & 0.0545 & 0.0459 & 6.322 \\
MDF\_A\_bin1x1\_ome\_0 & 0.0112 & 0.0332 & 0.0220 & 2.958 \\
Oak\_A\_bin1x1\_ome\_0 & 0.0139 & 0.0322 & 0.0183 & 2.313 \\
Ox\_bone\_A\_bin1x1\_ome\_0 & 0.0076 & 0.0754 & 0.0678 & 9.916 \\
Vertebrae\_D\_80kV\_ome & 0.0535 & 0.0530 & -0.0005 & 0.991 \\
\bottomrule
\end{tabular}
\end{minipage}\par\vspace{8pt}
\endgroup

\subsection{Frozen-output perturbation diagnostics}
\label{sec:diagnostics}
The perturbation analysis evaluates frozen outputs under controlled perturbations without updating model weights or injecting a correction into the saved reconstructions. For each of the ten test objects it uses eight frozen $128^3$ HR crops, then equally averages the crop measurements within object and the object summaries across the test set. Its scores describe diagnostic component response in this crop-scoped setting. The PIN--A0 endpoint comparison remains the reconstruction evidence for adding the optional bounded correction to the core.

A single Python \texttt{random.Random(0)} is advanced across test objects in lexicographic name order. For each object, integer origins for $32^3$ LR crops are drawn independently along each axis, from zero through the corresponding LR dimension minus 32. Each HR origin is four times its LR origin; duplicate origins and candidates whose $128^3$ HR crop exceeds the volume boundary are rejected. The valid-intensity fraction is the fraction of all LR crop voxels with finite intensity strictly between 0.05 and 0.98. Candidates require this fraction to be at least 0.01 and the crop mean to exceed 0.05. Candidate collection stops at 96 eligible unique origins or 2,048 total draws, whichever occurs first; duplicate and rejected draws count toward the draw limit. They are sorted by decreasing valid fraction, with ties resolved by increasing LR-origin coordinates in $z,y,x$ order, and the first eight are frozen per object. The same catalog is reused for A0, PIN, and all perturbations. Gaussian blur uses a normalized separable Gaussian with $\sigma_{\mathrm{HR}}=1.7$ HR voxels, radius $\lceil3\sigma_{\mathrm{HR}}\rceil=6$ (13 HR-voxel samples per axis), and zero padding. Gaussian noise is generated by adding independent zero-mean Gaussian samples of SD 0.05. The grid adds 0.065 times the mean of three axis-aligned sinusoids with period 4, 8, or 16 HR voxels. Blur, noise, and grid outputs are clipped to $[0,1]$. Structural displacement is a circular shift by four HR voxels along $z$. Mismatched LR uses the cyclic next object in sorted test-object order at the same crop index, with the affine fit recomputed for the presented pair.

Frozen-output perturbations elicit distinct responses from the diagnostic components. Gaussian noise produces a strong structural response, while the period-eight grid has a substantially weaker response; mismatched LR increases the observation-domain discrepancy without changing the stored HR prediction. All three full PIN runs reach epoch 60 without copy-alarm termination. The A2 component-removal control exhibits an abnormal-copy warning under monitor-only operation. The assembled correction satisfies the prescribed 10\% output-space norm cap.

\begingroup\fontsize{9}{10.8}\selectfont\setlength{\tabcolsep}{4pt}\renewcommand{\arraystretch}{1.12}\renewcommand{\SITableCaption}{\textbf{Frozen-output diagnostic responses.} Every row uses ten objects with eight frozen crops per object. Component scores and active fraction are evaluated without an SR parameter update or a correction to the stored reconstructions. $\Delta r_{\mathrm{LR}}$ is the LR residual change; component scores are diagnostic quantities. Active fraction averages the four binary component-activity indicators per crop, then averages over the eight crops within each object and equally over the ten objects.}\begin{longtable}{@{}>{\raggedright\arraybackslash}p{0.450in}>{\centering\arraybackslash}p{1.770in}>{\centering\arraybackslash}p{0.590in}>{\centering\arraybackslash}p{0.590in}>{\centering\arraybackslash}p{0.590in}>{\centering\arraybackslash}p{0.670in}>{\centering\arraybackslash}p{0.590in}>{\centering\arraybackslash}p{0.590in}@{}}
\caption{\SITableCaption}\label{tab:si-compact-perturbation-response}\\
\toprule
Source & Perturbation & $\Delta r_{\mathrm{LR}}$ & Geometry & Copy & Structural & Energy & \shortstack{Active\\fraction} \\
\midrule
\endfirsthead
\multicolumn{8}{l}{\textit{Table \thetable{} continued}}\\
\toprule
Source & Perturbation & $\Delta r_{\mathrm{LR}}$ & Geometry & Copy & Structural & Energy & \shortstack{Active\\fraction} \\
\midrule
\endhead
\midrule\multicolumn{8}{r}{Continued on next page}\\
\endfoot
\bottomrule
\endlastfoot
A0 & Healthy control & 0.0000 & 0.0000 & 0.0085 & 1.9026 & 0.0000 & 0.2000 \\
A0 & Gaussian blur & 0.0279 & 0.0003 & 0.0361 & 1.8519 & 0.0068 & 0.4500 \\
A0 & Gaussian noise & 0.0029 & 0.0000 & 0.0078 & 16.3889 & 0.0000 & 0.2781 \\
A0 & Grid, period 4 & 0.0046 & 0.0000 & 0.0095 & 3.4227 & 0.0000 & 0.2281 \\
A0 & Grid, period 8 & -0.0000 & 0.0000 & 0.0086 & 1.6210 & 0.0000 & 0.1938 \\
A0 & Grid, period 16 & 0.0008 & 0.0000 & 0.0087 & 2.2992 & 0.0000 & 0.2375 \\
A0 & Local structural displacement & 0.0111 & 0.0134 & 0.0454 & 1.1059 & 0.0769 & 0.4719 \\
A0 & Mismatched LR & 0.1180 & 0.0598 & 0.0842 & 0.1978 & 0.0000 & 0.3406 \\
PIN & Healthy control & 0.0000 & 0.0000 & 0.0077 & 0.9902 & 0.0000 & 0.1812 \\
PIN & Gaussian blur & 0.0279 & 0.0004 & 0.0342 & 1.1319 & 0.0097 & 0.4281 \\
PIN & Gaussian noise & 0.0037 & 0.0000 & 0.0073 & 16.1821 & 0.0000 & 0.2719 \\
PIN & Grid, period 4 & 0.0052 & 0.0000 & 0.0084 & 2.5062 & 0.0000 & 0.2219 \\
PIN & Grid, period 8 & -0.0000 & 0.0000 & 0.0080 & 0.7663 & 0.0000 & 0.1750 \\
PIN & Grid, period 16 & 0.0007 & 0.0000 & 0.0078 & 1.4668 & 0.0000 & 0.2156 \\
PIN & Local structural displacement & 0.0108 & 0.0139 & 0.0463 & 0.4548 & 0.0765 & 0.4281 \\
PIN & Mismatched LR & 0.1169 & 0.0586 & 0.0813 & 0.1172 & 0.0000 & 0.3312 \\
\end{longtable}
\endgroup

\begingroup\fontsize{9}{10.8}\selectfont\setlength{\tabcolsep}{4pt}\renewcommand{\arraystretch}{1.12}\renewcommand{\SITableCaption}{\textbf{Frozen-output cotangent bounds.} The norm ratio and cap-hit fraction describe the assembled correction at the detached output leaf; it is not applied to saved predictions. Cap-hit fraction averages the per-crop indicator $\mathbf{1}[\|w\|_2\geq4]$ within object and then equally across objects, with $w$ defined in Eq.~\eqref{eq:si-cotangent-sum}.}\begin{longtable}{@{}>{\raggedright\arraybackslash}p{0.450in}>{\centering\arraybackslash}p{1.770in}>{\centering\arraybackslash}p{0.590in}>{\centering\arraybackslash}p{0.590in}@{}}
\caption{\SITableCaption}\label{tab:si-compact-perturbation-bounds}\\
\toprule
Source & Perturbation & \shortstack{$\|u\|/$\\$\|v_{\mathrm{LR}}\|$} & \shortstack{Cap-hit\\fraction} \\
\midrule
\endfirsthead
\multicolumn{4}{l}{\textit{Table \thetable{} continued}}\\
\toprule
Source & Perturbation & \shortstack{$\|u\|/$\\$\|v_{\mathrm{LR}}\|$} & \shortstack{Cap-hit\\fraction} \\
\midrule
\endhead
\midrule\multicolumn{4}{r}{Continued on next page}\\
\endfoot
\bottomrule
\endlastfoot
A0 & Healthy control & 0.0476 & 0.0375 \\
A0 & Gaussian blur & 0.0457 & 0.0750 \\
A0 & Gaussian noise & 0.1000 & 0.2500 \\
A0 & Grid, period 4 & 0.0721 & 0.2875 \\
A0 & Grid, period 8 & 0.0412 & 0.0125 \\
A0 & Grid, period 16 & 0.0559 & 0.1250 \\
A0 & Local structural displacement & 0.0303 & 0.0000 \\
A0 & Mismatched LR & 0.0001 & 0.0000 \\
PIN & Healthy control & 0.0257 & 0.0000 \\
PIN & Gaussian blur & 0.0285 & 0.0625 \\
PIN & Gaussian noise & 0.1000 & 0.2750 \\
PIN & Grid, period 4 & 0.0605 & 0.1375 \\
PIN & Grid, period 8 & 0.0201 & 0.0000 \\
PIN & Grid, period 16 & 0.0374 & 0.0250 \\
PIN & Local structural displacement & 0.0140 & 0.0000 \\
PIN & Mismatched LR & 0.0001 & 0.0000 \\
\end{longtable}
\endgroup

\FloatBarrier
\clearpage
\section{Qualitative A2--A0--PIN comparison}
\label{sec:a2-gallery}

\FloatBarrier
{\small
\setbiblabelwidth{2}
\bibliographystylesupp{ieeenat_fullname}
\bibliographysupp{references_supp}
}

\end{document}